\pdfoutput=1

\documentclass[11pt]{article}

\usepackage{acl}

\usepackage{times}
\usepackage{latexsym}
\usepackage{booktabs}
\usepackage{tabularx}
\usepackage{wrapfig}
\usepackage{algorithm}
\usepackage{algpseudocode}

\algrenewcommand\algorithmicindent{1.2em}

\usepackage{balance}
\usepackage[T1]{fontenc}

\usepackage[utf8]{inputenc}

\usepackage{microtype}

\usepackage{inconsolata}

\usepackage{kotex}
\usepackage{graphicx}
\usepackage{tcolorbox}
\usepackage{epsfig}
\usepackage{amsmath}
\usepackage{amssymb}
\usepackage{booktabs}
\usepackage{arydshln}
\usepackage{multirow}
\usepackage{wrapfig}
\usepackage{caption}
\usepackage{mathtools}
\usepackage{float}
\usepackage{placeins}
\newcommand{\J}{\mathcal{J}}
\newcommand{\clip}{\operatorname{clip}}
\newcommand{\E}{\mathbb{E}}

\title{SCoPE VLM: Selective Context Processing for Efficient Document Navigation in Vision-Language Models}

\author{
Gyubeum Lim$^{1}$, Yemo Koo$^{2}$, Vijay Krishna Madisetti$^{1}$ \\
$^{1}$Georgia Institute of Technology \quad $^{2}$Konkuk University \\
\texttt{\{glim36,vkm\}@gatech.edu} \quad \texttt{yemoyemo010831@konkuk.ac.kr}
}

\begin{document}
\maketitle
\begin{abstract}
Understanding long-context visual information remains a fundamental challenge for vision-language models, particularly in agentic tasks such as GUI control and web navigation. While web pages and GUI environments are inherently structured documents, current VLMs typically neglect decision-oriented document understanding in their training objectives. Existing approaches primarily extend visual embeddings to process long, high-resolution inputs, but these methods are memory-intensive and impractical for locally deployable solutions. To address these issues, we propose SCoPE VLM, a document navigation expert that leverages a novel Chain of Scroll mechanism to selectively and recursively navigate documents, focusing exclusively on relevant segments. We introduce a dedicated data generation pipeline to construct informative Chain of Scroll trajectories and Episodic Group Relative Policy Optimization, a tailored reinforcement learning method to bridge the gap between training and inference.  Our method substantially reduces memory usage and effectively models human-like reading behaviors. To the best of our knowledge, SCoPE VLM is the first framework to explicitly model agentic reading patterns in multi-page document question answering, advancing the capabilities of multimodal agents.
\end{abstract}

\section{Introduction}

Recent advancements in Vision-Language models (VLMs) ~\cite{OpenAI2023GPT4V, openai2024gpt4ocard, liu2023visualinstructiontuning, qwen, chen2023internvl} have demonstrated outstanding capabilities in visual understanding, achieving a tight alignment between text and images to answer complex queries. Despite these successes, handling long-context information remains a substantial challenge, especially for long documents and GUI environments. In fact, tasks involving GUI and web environments~\cite{xie2024osworldbenchmarkingmultimodalagents, koh2024visualwebarenaevaluatingmultimodalagents, sun2025osgenesisautomatingguiagent} inherently require both sophisticated multimodal document understanding and agentic capabilities. However, conventional approaches neglect action-based document understanding. Furthermore, state-of-the-art VLMs struggle in these environments due to their limited capability to explore and process extensive multimodal contexts. To address this, reinforcement learning offers a promising approach to teach such agentic behavior. While training directly on document-based environments offers a practical and accessible alternative to costly and unstable GUI simulators~\cite{bai2024digirltraininginthewilddevicecontrol}, this remains underexplored in the community.

\begin{figure}[!t]
\begin{center}
\includegraphics[width=1.0\columnwidth]{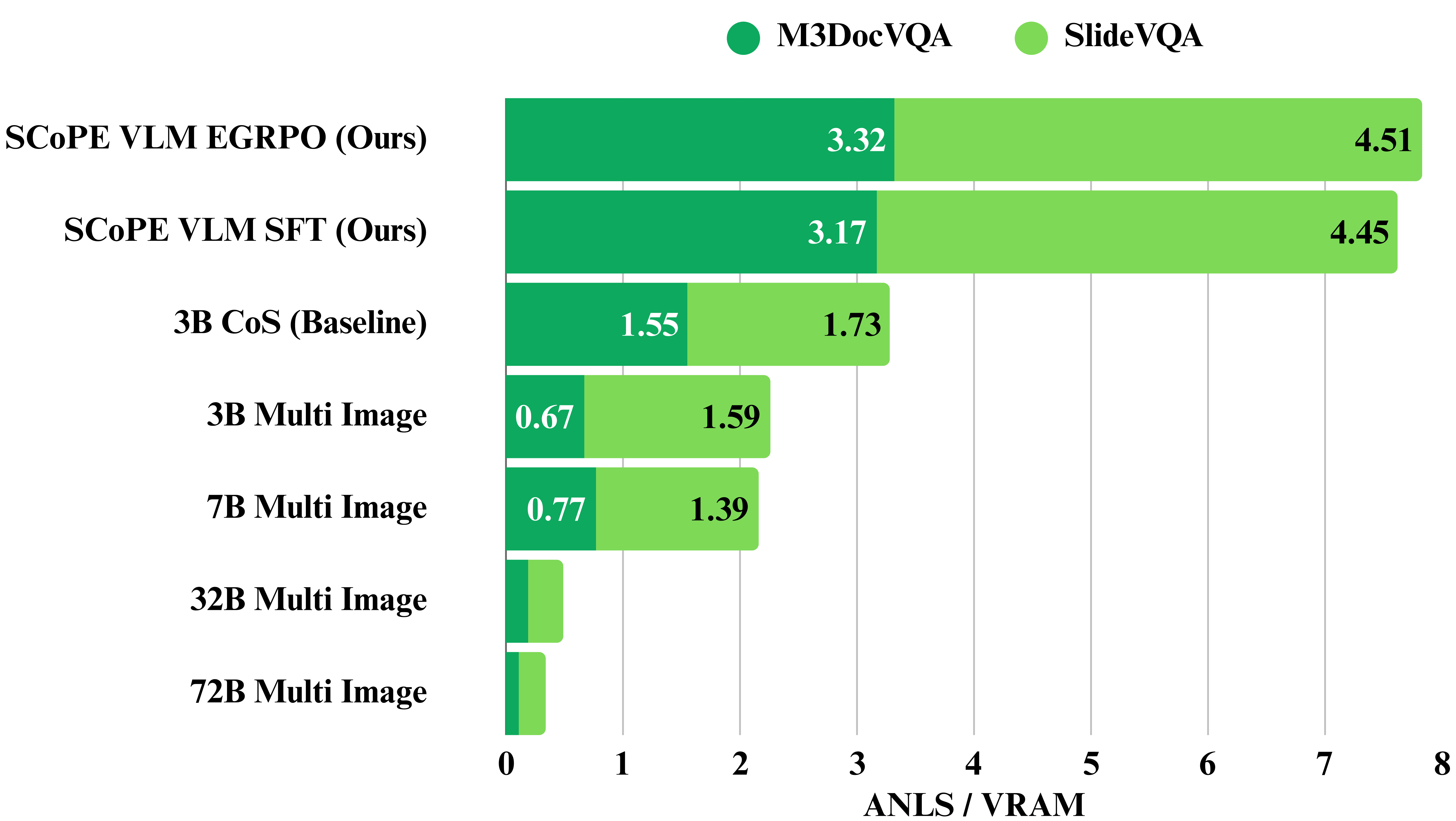}
  \captionof{figure}{Performance-efficiency comparison of SCoPE VLM variants against Qwen2.5 VL series on M3DocVQA and SlideVQA from Table~\ref{tab:vram}, Section~\ref{sec:Experiments}, achieving 2.38x higher efficiency than the baseline.}
  \label{fig:front}
\end{center}
\end{figure}

Despite these advantages in action-based document understanding, existing approaches merely extend visual embeddings to process high-resolution images in long-context scenarios \cite{qwen25vl, ye2024mplugowl3longimagesequenceunderstanding, li2024llavaonevisioneasyvisualtask}. These methods rely on processing the entire visual context in a single pass, which is inevitably memory-intensive. As a result, they are not only impractical for local deployment but also fundamentally unscalable to real-world scenarios such as 100-page documents or websites.

\begin{figure*}[!t]
\begin{center}
\resizebox{1.0\textwidth}{!}{%
\includegraphics[width=\textwidth]{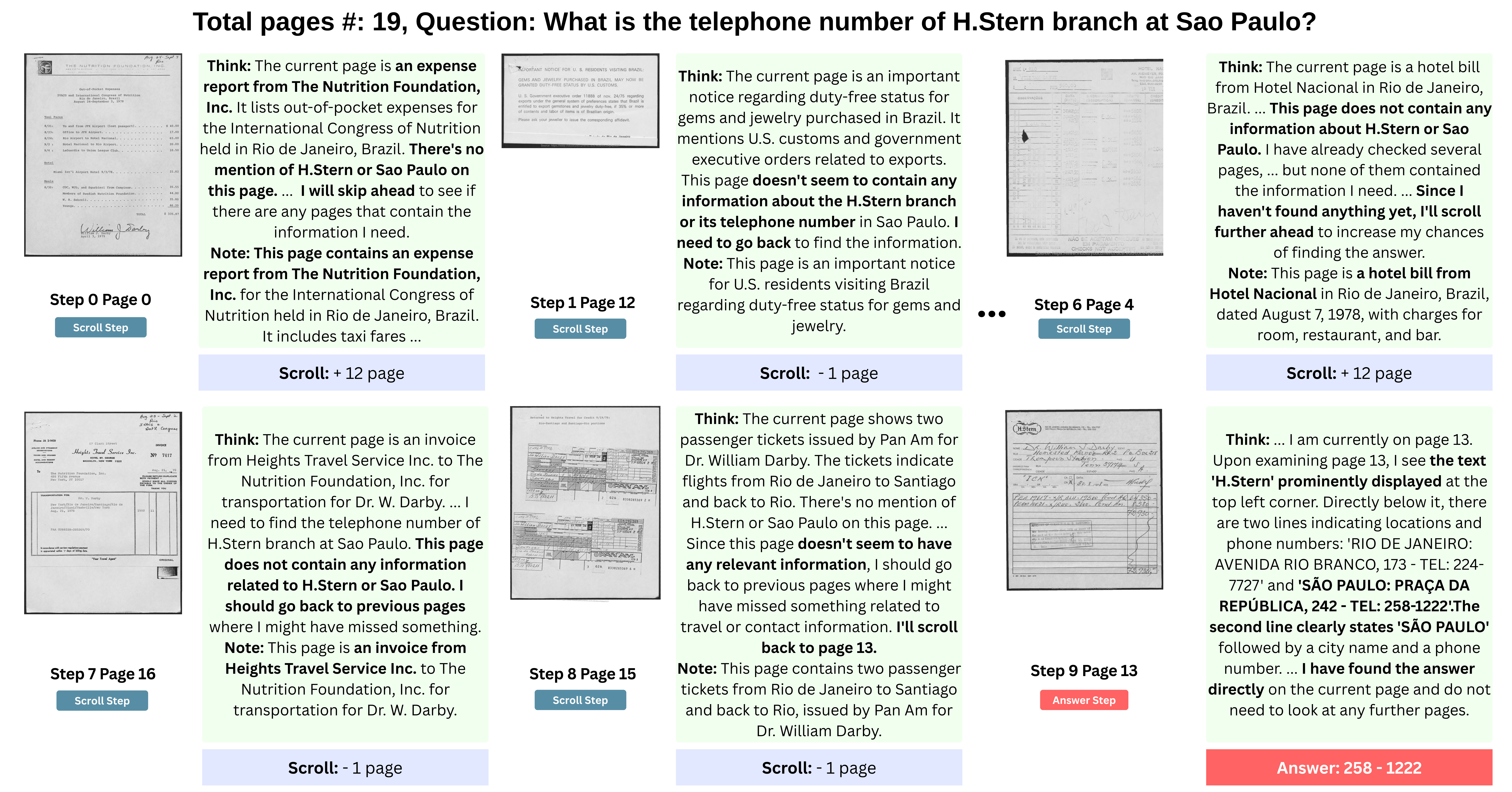}
}
\caption{Chain of Scroll navigation trajectories. SCoPE VLM with EGRPO efficiently locates telephone information in a 19-page document, visiting only 9 pages.}
\label{fig:egrpo_answer}
\end{center}
\end{figure*}

To mitigate these limitations, we propose SCoPE VLM, a document navigation expert that leverages the Chain of Scroll framework (CoS), an action-based inference time scaling strategy that allows the model to focus only on relevant document segments to answer the user query. The framework is executed through numerous recursive generations based on query-relevant context, which the model generated in the previous step.

To train such advanced decision-making capabilities in VLMs, we introduce the cold start dataset for Supervised-Fine-Tuning (SFT) CoS Framework. Although SFT is effective in teacher-forcing the output distribution quickly, a gap still presents between the inference and SFT training. Thus, we further enhance the model’s agentic capabilities by introducing Episodic Group Relative Policy Optimization (EGRPO) to further push the limit of SCoPE VLM for effective context-based action selection.

Empirical evaluations illustrate that SCoPE VLM maintains the accuracy of existing baselines while drastically improving memory efficiency. As illustrated in Figure~\ref{fig:front}, our model exhibits superior ANLS/VRAM performance, with the total scores reflecting the aggregated metrics across both evaluation datasets. Additionally, our method effectively generalizes its learned navigation capabilities to GUI control tasks, demonstrating enhanced adaptability compared to the baseline model. By explicitly modeling agentic behaviors in document-based question answering, SCoPE VLM represents a critical advancement toward more efficient, capable, and locally deployable multimodal agents.

\section{Related Work}
\subsection{Embedding High Resolution Images}
To incorporate high-resolution images, conventional VLMs leverage multi-window based vision token processing. LLaVA-Next~\cite{
liu2024llavanext}
pioneered this approach by splitting high-resolution images into up to five windows, using 5x more visual tokens than standard single window image processing~\cite{
liu2023visualinstructiontuning} to better capture fine details. Furthermore, InternVL 1.5 enhanced this paradigm by scaling the number of tiles up to x40 of the single window, allocating 10496 tokens per image through its tile-based processing system. Qwen2.5 VL~\cite{
qwen25vl} further extends by exploiting the Naive Dynamic Resolution~\cite{dehghani2023patchnpacknavit} which consumes up to 16384 image tokens. They demonstrate that the increased number of visual tokens enables better preservation of fine-grained visual understanding. However, the computational costs scale exponentially with token counts, which results in high memory requirements in both training and inference time. These trade-offs highlight that the recent advancements in high-resolution image processing are not scalable to long document understanding.

\begin{figure*}[!t]
\begin{center}
\resizebox{1.0\textwidth}{!}{%
\includegraphics[width=\textwidth]{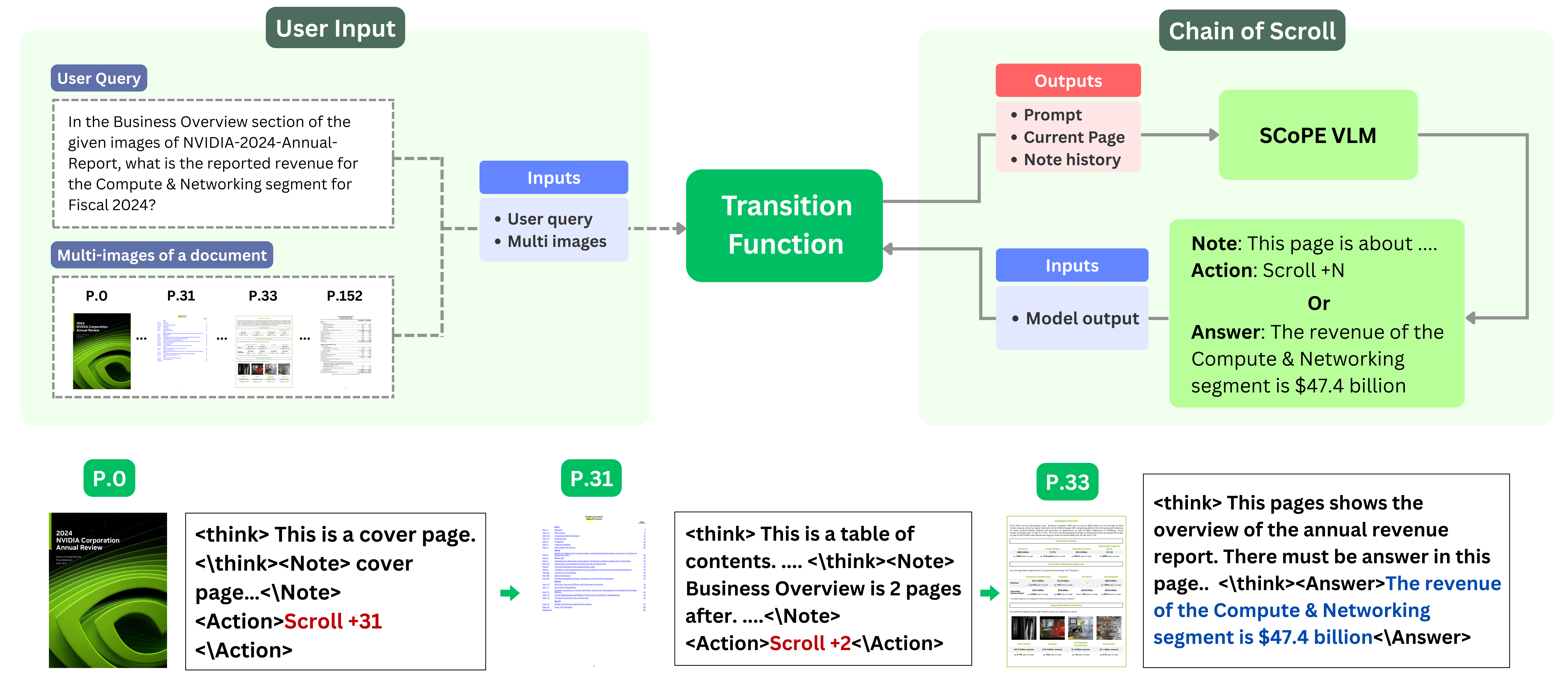}
}
\caption{Overview of the Chain of Scroll framework. The model iteratively decides whether to scroll to new pages or generate an answer based on accumulated context and relevance signals, emulating human-like selective document navigation.}
\label{fig:CoS}
\end{center}
\end{figure*}

\subsection{Vision Token Compression}
A prominent strategy for mitigating extensive vision token processing is vision token compression. Honeybee~\cite{cha2023honeybee}  refines token compression by proposing locality-enhanced projectors, thus improving token relevance in dense visual data. In video understanding, where every frame contributes a dense set of visual tokens, visual token compression becomes more critical. PVC~\cite{yang2024pvcprogressivevisualtoken} and LongVU~\cite{shen2024longvuspatiotemporaladaptivecompression} both attempt to mitigate this issue by dynamically compressing tokens, preserving critical information while discarding less relevant details. However, these methods frequently struggle with performance degradation when tasks demand high resolution.

\subsection{Multimodal Retrieval-Augmented Generation (RAG)}

Multimodal Retrieval-Augmented Generation (RAG) has emerged as a promising framework for addressing the challenges of integrating extensive external multimodal contexts. These systems combine a retrieval mechanism to search relevant multimodal contexts with a generation model \cite{yasunaga2023retrievalaugmentedmultimodallanguagemodeling}. Despite its strengths, RAG systems struggle with processing on-demand user queries involving long-context documents. A primary constraint is the reliance on a pre-structured database, which must be prepared in advance \cite{lewis2021retrievalaugmentedgenerationknowledgeintensivenlp}. This requirement introduces challenges in dynamic or real-time scenarios where the user query may involve newly introduced data. 

\section{Proposed Methods}
\subsection{Chain of Scroll}

\begin{algorithm}[!h]
  \caption{Chain of Scroll (CoS)}
  \label{alg:cos_inference}
  \small
  \begin{algorithmic}[1]
    \Statex \textbf{Input:}
    \Statex \quad $q$ \hfill\Comment{user’s query}
    \Statex \quad $\mathit{imgs} = \{\mathit{Image}_0, \dots, \mathit{Image}_N\}$ \hfill\Comment{input images}
    \Statex \quad $\mathit{maxSteps}$ \hfill\Comment{max steps allowed to scroll}
    \Statex \textbf{Output:} $\mathit{answer}$
    \Statex \rule{\linewidth}{.2pt}

    \Statex \textbf{Initialization:}
    \State $\mathit{answer} \gets \varnothing$          \Comment{no answer yet}
    \State $\mathit{page} \gets 0$                      \Comment{start at first page}
    \State $\mathit{scroll} \gets 0$                    \Comment{number of pages to move}
    \State $\mathit{notes} \gets \varnothing$           \Comment{all collected notes}
    \State $\mathit{visited} \gets [\text{False}]^{|\mathit{images}|}$ \Comment{store visited pages}
    \Statex \rule{\linewidth}{.2pt}

    \Statex \textbf{Run Chain of Scroll:}
    \For{$\mathit{step} \gets 1$ \textbf{to} $\min(\mathit{maxSteps}, |\mathit{images}|)$}
        \State $\mathit{c}\gets(q,\mathit{imgs},\mathit{page},\mathit{scroll},\mathit{notes},\mathit{visited})$
        \State $\mathit{prompt},\mathit{cur\_img},page,\mathit{visited}
               \gets \textsc{Tran\_fn}(\mathit{c})$        
        \State $\mathit{response} \gets \pi_{\theta}(\mathit{\mathit{cur\_img}, prompt})$
        \State $(\mathit{cur\_note}, \mathit{scroll}, \mathit{answer})
               \gets \textsc{Parse}(\mathit{response})$
        \State $\mathit{notes} \gets \mathit{notes} \cup \{\mathit{cur\_note}\}$
        \If{$\mathit{answer} \neq \varnothing$}
            \State \textbf{break}    \Comment{answer found, early exit}
        \EndIf
    \EndFor
    \State \Return $\mathit{answer}$
  \end{algorithmic}
\end{algorithm}

In real-world document question answering, queries are often localized to specific sections of a document rather than requiring holistic processing. While each page carries its own content, its relevance to the query varies; some pages provide critical information, while others are entirely irrelevant. Humans naturally adapt to this structure by skipping or answering based on contextual signals. This selective nature leads to efficient context processing with a minimal number of pages visited. To emulate this adaptive behavior, we propose the Chain of Scroll, an action-based document navigation strategy to mimic human-like reading behavior by making informed decisions at each step.

The Chain of Scroll framework allows the model to make recursive decisions—whether to scroll or answer—based on contextual cues accumulated throughout the episode. By modeling the semantic coherence between the query and document segments, the model learns to predict which regions are likely to contain relevant information, how far to skip, and when to stop. This selective navigation not only reduces unnecessary computation but also mirrors human-like reading behavior, where not all content is equally attended to. The overall flow of the proposed CoS framework is illustrated in Figure~\ref {fig:CoS} and formally outlined in Algorithm~\ref{alg:cos_inference}. In addition, Figure~\ref{fig:egrpo_answer} demonstrates the human-like document navigation to answer the user query, while processing half of the document. See Section~\ref{sec:full_main} for the full trajectory. For the formal representation of the CoS framework, including the action space and transition function, see Section~\ref{sec:cos_formal} in the appendix.

\noindent \textbf{Action space:}
In the CoS framework, the model operates within a discrete action space consisting of two steps: Scroll and Answer. At the beginning of each CoS trajectory, the model is given the first page of the document along with the user query. After performing Chain of Thought (CoT) reasoning ~\cite{wei2022chain, wang2023selfconsistency, mu2023embodiedgpt, lu2022learn, guo2024mammothvlelicitingmultimodalreasoning}, the model may choose to take notes and return a scroll value relative to the current page number. These generated notes are accumulated and used as context for subsequent steps via the defined transition function. Once the model has gathered sufficient information to address the query, it selects the Answer action following another CoT step. Thus, the action space comprises (1) a scroll value range bounded by the total number of pages in both positive and negative directions and all possible notes with query-relevant information, and (2) all possible answers. Both notes and answers are represented within the model’s token vocabulary.

\noindent \textbf{Transition function:} To emulate action-based context processing, it is essential to design a function that maps a previous state to the next, enabling sequential decision-making without a pre-defined environment. RL4VLM \cite{zhai2024finetuninglargevisionlanguagemodels} introduced a post-processing function to integrate VLMs into the decision-making loop; however, it still depended on fixed environments. In contrast, we propose a transition function that enables step-by-step progression for targeted document navigation as a Markov Decision Process (MDP). The major role of the transition function is to transform user queries into a structured CoS prompt (see an example in Figure~\ref{fig:cos-prompt} in the appendix), bring the chosen page by the action from the previous step, validate scroll actions, manage visited page histories, and accumulate evolving notes that summarize key observations. At each step, it processes the model's output, maps to the next step, and contextualizes based on the previous responses. As shown in Algorithm~\ref{alg:cos_inference}, the transition function plays the role of an environment by directing states and propagating contextual information. This function is used during both training and inference in our framework. It also serves as an exception handler, returning the context based on a random page when given an illegal scroll action.

\noindent \textbf{Input prompt:}
As shown in Figure~\ref{fig:cos-prompt} in the appendix, the input prompt dynamically updates at each step with accumulated notes that preserve query-related context. This enables the model to maintain a continuous memory of its exploration based on globally informed decisions about whether to scroll further or generate an answer. Figure~\ref{fig:egrpo_answer} illustrates how context from previous steps can effectively persist across exploration. In contrast to traditional multi-turn inference, our framework functions through multiple iterations of single-turn inference, with previous context preserved in accumulated notes. This design considerably improves memory efficiency while also allowing analysis of multi-page documents in high-resolution, with memory requirements comparable to those of processing a single image. See Section~\ref{sec:cos_prompt} in the appendix for the CoS prompt and an example.

\subsection{SCoPE Dataset}

\begin{table}[!h]
  \centering
  \renewcommand{\arraystretch}{1.5}
  {\fontsize{34pt}{36pt}\selectfont
    \resizebox{0.5\textwidth}{!}{%
      \begin{tabular}{lcccc}
        \hline
        \textbf{Name}            & \textbf{M3DocVQA} & \textbf{MP-DocVQA} &
        \textbf{SlideVQA}        & \textbf{MMLongBench-Doc} \\
        \hline
        \#Pages / question        & 8.95              & 18.65             & 20.00                   & 47.67                   \\
        \#Questions               & 8K                & 8K                & 4K                      & 1K                      \\
        Evidence page             & Multi         & Single            & Multi                   & Multi                   \\
        \hline
      \end{tabular}
    }
  }
  \caption{Training datasets for supervised fine-tuning and reinforcement learning stages.}
  \label{tab:dataset}
\end{table}

\begin{figure*}[!t]
\begin{center}
\resizebox{1.0\textwidth}{!}{%
\includegraphics[width=\textwidth]{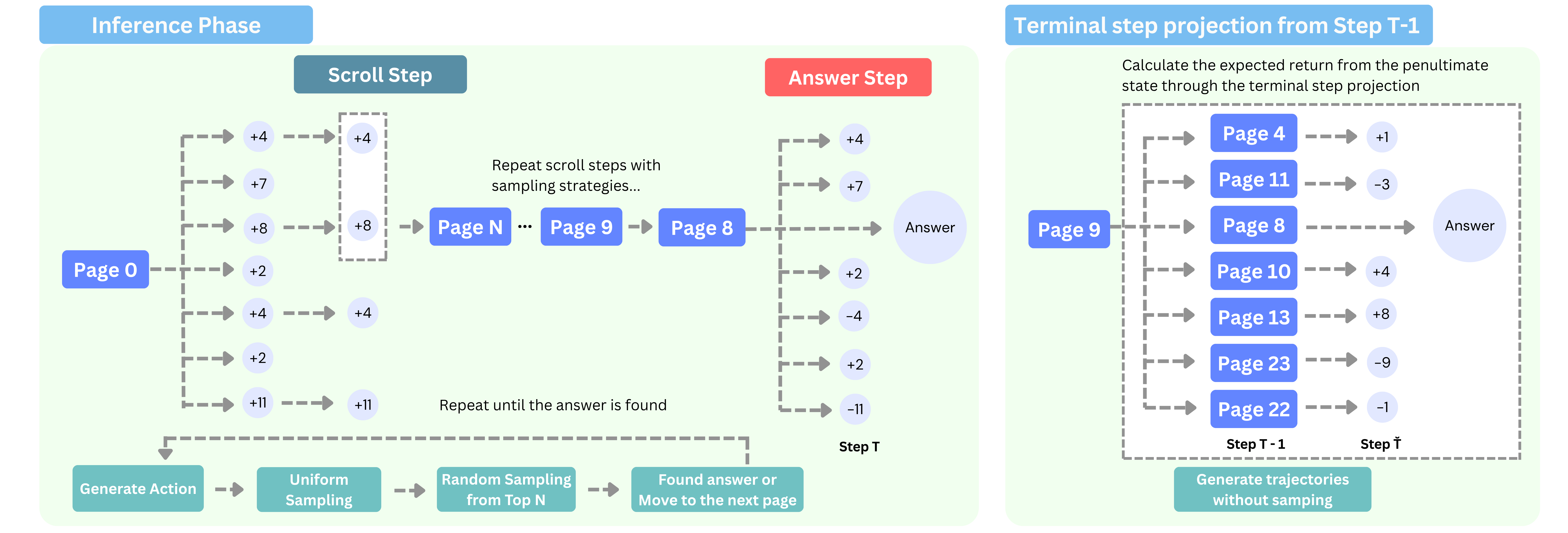}
}
\caption{Overview of Episodic Group Relative Policy Optimization. The tailored framework for CoS employs group-based sampling with top-N selection during document navigation. Terminal step projection from the penultimate state enables return estimation that guides the model toward correct answers.}
\label{fig:egrpo_main}
\end{center}
\end{figure*}

The SFT dataset for CoS is designed to capture the reasoning and decision-making capabilities of large-scale VLMs as a baseline. To generate high-quality annotations similar to Figure~\ref{fig:egrpo_answer} at every sampled steps, we employ a pseudo-labeling ensemble comprising Gemini 1.5 Pro, Gemini Flash 2.0, and Gemini Flash 2.0 Thinking~\cite{geminiteam2024gemini15unlockingmultimodal}, optimizing for generation time, cost, and quality. This ensemble is applied to a comprehensive corpus of 21K multi-page document question-answering samples, as shown in Table~\ref{tab:dataset}~\cite{cho2024m3docragmultimodalretrievalneed, tito2023hierarchicalmultimodaltransformersmultipage, tanaka2023slidevqadatasetdocumentvisual, ma2024mmlongbenchdocbenchmarkinglongcontextdocument}.

Since CoS is a novel task, even state-of-the-art proprietary VLMs lack the capabilities to generate high-quality CoS episodes without guidance. To address this limitation, we leverage datasets with explicit annotations of the evidence pages that must be visited. For M3DocVQA, we utilize the annotation model to pre-select query-relevant pages for each question.

Before generating CoS step responses, we sample page trajectories that models choose to read. Each trajectory designates page 0 as the initial state and selects one target page as the terminal state. To introduce natural variance, we randomly sample trajectory lengths, while incorporating both target pages and unrelated pages as intermediate steps to simulate realistic reading patterns.

During annotation, pseudo-labeling models receive three key inputs: the current page index, the question, and a defined scroll action transitioning to the next step along the predefined trajectory. This approach enables generations of appropriate scroll step responses. By explicitly providing the next page in advance, we induce high-quality reasoning that semantically aligns with the annotated action. For terminal steps, the annotation model formulates the final answer based on the current image, accumulated notes from previous steps, and the answer. This guided approach ensures that reasoning, memorization behavior, and decision-making are tightly aligned throughout the trajectory. Also, we annotated samples without answers to return “The answer cannot be found.” The response of our model to a query with no answer is presented in Section~\ref{sec:no_answer}.

\subsection{Reinforcement Learning}
With the SCoPE dataset, we perform Supervised Fine-Tuning (SFT) on Qwen2.5 VL 3B Instruct due to its strong performance in high-resolution document understanding. While SFT effectively establishes basic decision-making capabilities, several challenges persist. First, the SFT model occasionally selects invalid actions that exceed the allowed image index range. Second, it often continues scrolling even after all document pages have been processed. See Section~\ref{sec:full_main} for the full Chain of Scroll trajectory of SCoPE VLM SFT which fails to generate a valid answer after visiting all pages in the given document. These issues highlight a fundamental challenge: determining whether to continue scrolling or to provide an answer based on potentially incomplete information. Such decision-making under uncertainty cannot be adequately addressed by SFT alone, as it lacks mechanisms to encourage exploration beyond ground-truth trajectories. To overcome these limitations, we incorporate reinforcement learning (RL) to optimize CoS capabilities of SCoPE VLM.

\noindent \textbf{Episodic Group Relative Policy Optimization:}
To adapt GRPO~\cite{grpo} to the episodic structure of CoS, we introduce Episodic Group Relative Policy Optimization (EGRPO), a group-based approach tailored for multi-step document navigation tasks. As illustrated in Figure~\ref{fig:egrpo_main}, EGRPO generates $\tilde{G}$ candidate responses at each step and employs a two-stage sampling strategy: first, we uniformly sample $G$ candidates from the original $\tilde{G}$ to preserve reward diversity, then randomly select one response from the top-$N$ candidates ranked by reward scores. This hybrid approach effectively balances exploration and exploitation—the uniform sampling maintains diversity across the reward spectrum while the top-$N$ selection ensures high-quality action selection, allowing the model to efficiently explore diverse trajectories while systematically pursuing paths that maximize returns.

EGRPO extends GRPO by optimizing both the terminal step $T$ (the answer step) and the penultimate step $T-1$ (the final scroll step). To estimate future returns without exhaustive trajectory generation, we introduce a terminal step projection mechanism. After episode termination, we perform one additional generation to the projected terminal step ($\check{T}$) from each penultimate state without sampling. This allows us to evaluate the quality of penultimate actions by observing where they lead, effectively propagating reward signals backward through the trajectory. The terminal state projection is under the assumption that the penultimate scroll step represents a state from which critical information needed to answer the user's question is accessible. This assumption suggests that the probability of providing a correct answer given the penultimate state approximates the product of reaching the answer page and correctly extracting the answer once there. Also, we assume that following the dynamic programming principle, the training process optimizes not only the final transitions but the entire trajectories across diverse training sequences of varying lengths.

The EGRPO objective is provided below:

{\footnotesize
\setlength{\abovedisplayskip}{4pt}
\setlength{\belowdisplayskip}{4pt}
\begin{align}
\J_{\text{EGRPO}}(\theta)
  &= \gamma\,\J^{(T)}(\theta)+\J^{(T-1)}(\theta)             \label{eq:Jegrpo}\\[2pt]
\J^{(t)}(\theta)
  &= \E_{(q,a)\sim\mathcal{D}}
     \E_{\{o_i^{(t)}\}_{i=1}^{G}}\!
     \Biggl[
       \frac{1}{G}\sum_{i=1}^{G}\frac{1}{|o_i^{(t)}|}              \\[-3pt]
&      
       \sum_{k=1}^{|o_i^{(t)}|}
       \min\!\Bigl(
         \rho_{i,k}^{(t)}\hat A_{i}^{(t)},\;
         \clip\!\bigl(\rho_{i,k}^{(t)},1-\varepsilon,1+\varepsilon\bigr)\hat A_{i}^{(t)}
       \Bigr)
     \Biggr]                                             \label{eq:Jt}\\[-3pt]
&\text{where } t \in \{T-1, T\},\; \{\tilde o_j^{(t)}\}_{j=1}^{\tilde G}\!\sim\!\pi_{\theta_{\text{old}}}(\cdot\!\mid\!q),\ \notag\\
&\quad\{o_i^{(t)}\}_{i=1}^{G}\!\subseteq\!\{\tilde o_j^{(t)}\}_{j=1}^{\tilde G} \notag\\[2pt]
\rho_{i,k}^{(t)}(\theta)
  &= \frac{\pi_\theta\!\bigl(o_{i,k}^{(t)}\mid q,\,o_{i,<k}^{(t)}\bigr)}
          {\pi_{\theta_{\text{ref}}}\!\bigl(o_{i,k}^{(t)}\mid q,\,o_{i,<k}^{(t)}\bigr)},
  \quad
  \hat A_{i}^{(t)} = \dfrac{\hat r_{i}^{(t)}-\mu^{(t)}}{\sigma^{(t)}},
      \label{eq:ratio}\\
\hat r_{i}^{(T)}
  &= r_i^{(T)}, \quad \hat r_{i}^{(T-1)} = r_i^{(T-1)}+r_i^{(\check{T})}.
\end{align}
}

\noindent where $\pi_{\theta_{\text{old}}}$ generates the initial candidates and $\{o_j^{(t)}\}_{i=1}^{G}$ represents the uniformly sampled subset.
$k$ indexes the token position.
$t$ refers to the time step.
The advantage estimation incorporates both immediate rewards and projected future returns.
$\gamma$ balances the relative importance of terminal and penultimate objectives.
By focusing on these critical transition steps and maintaining computational efficiency through selective sampling, EGRPO enables effective learning in the episodic CoS framework.
See Section~\ref{egrpo_derivation} in the appendix for formal derivations, assumptions, and the algorithm.

\begin{table*}[ht!]
  \centering
  \resizebox{\textwidth}{!}{%
    \begin{tabular}{@{}llcccccc@{}}
      \toprule
      \textbf{Models} & \textbf{Type}
        & \textbf{DocVQA}
        & \textbf{MP-DocVQA}
        & \textbf{DUDE}
        & \textbf{M3DocVQA (single doc.)}
        & \textbf{SlideVQA}
        & \textbf{MMLongBench-Doc} \\
      \midrule
      Average \#Images per Question        & --     &  1  & 5.57   & 5.61     & 13.18  & 20.00   & 51.72    \\
      \midrule
      \multicolumn{8}{l}{\textbf{Open Source General VLMs}} \\
      \midrule
      LLAVA One Vision 7B            & MI & 87.05& 59.29     & 35.60    & 35.42     & 48.73     & 9.53    \\
      Qwen 2.5 VL 3B                 & MI &  92.54  & 76.28   &  46.14      &  43.72  &  51.46     &  14.02    \\
      Qwen 2.5 VL 7B                 & MI &  94.61  & 80.99   & 49.16      &  52.54     &   57.22    &   15.21   \\
      \midrule
      \multicolumn{8}{l}{\textbf{Chain of Scroll (Ours)}} \\
      \midrule
      Qwen 2.5 VL 3B        & CoS   &  66.65  &  47.78   &   31.73    & 19.54    &  23.23     &  9.08    \\
      Qwen 2.5 VL 72B        & CoS     &  90.60  & 80.83   & 48.42      & 35.83     &  66.73    & –    \\
      SCoPE VLM 3B SFT (Ours)        & CoS  &  85.41  & 74.49   & 42.82    & 46.13     & 59.88     &  16.89    \\
      SCoPE VLM 3B EGRPO (Ours)     & CoS & 85.39 & 73.07 & 42.29 & 48.27 & 57.31 &  17.90     \\
      \bottomrule
    \end{tabular}%
  }
  \caption{Performance comparison of VLMs on multi-page document benchmarks. SCoPE VLMs and Qwen series used the identical maximum image tokens per generation step for comparable memory usage.} 
  \label{tab:main}
\end{table*}

\section{Experiments}
\label{sec:Experiments}

\subsection{Training setup for SFT and RL}
During the SFT stage, we fine-tune Qwen 2.5 VL 3B on the full SCoPE dataset with a maximum of 1,003,520 pixels. For the RL stage, we deploy EGRPO to optimize both overall performance and navigation efficiency for Chain of Scroll. SCoPE VLM EGRPO is trained with LoRA~\cite{hu2021loralowrankadaptationlarge} for 2,500 steps using the same maximum pixel setting as the SFT stage. Further details can be found in Section~\ref{sec:hyperparameters} of the appendix.

\subsection{Evaluation}

The experimental evaluation of CoS focuses on assessing its effectiveness and efficiency on six distinct multi-page document question answering benchmarks with ANLS score. We conducted these evaluations under consistent constraints of the maximum number of visual tokens processed per step (Table~\ref{tab:main}, and ~\ref{tab:inference_strategy_comparison}, and \ref{tab:training_method_comparison}) and per image (Table~\ref{tab:vram}, and ~\ref{tab:AitZ} ) with CoS and Multi Image (MI) inference. For evaluation, we have extended LMMs-Eval~\cite{zhang2024lmmsevalrealitycheckevaluation} to support DUDE~\cite{vanlandeghem2023documentunderstandingdatasetevaluation}, M3DocVQA, SlideVQA, and MMLongBench-Doc. Additionally, the LoRA weights for SCoPE VLM EGRPO are merged prior to evaluation to ensure accurate VRAM usage measurement. See Section~\ref{sec:evaluation_setup} from the appendix for further details on the experiment setup.

\subsection{Enhancing decision-making capabilities under Chain of Scroll framework}

\begin{table*}[!t]
\centering
\resizebox{1.0\textwidth}{!}{
\begin{tabular}{@{}l l c c c c | c c c c@{}}
\toprule
\multicolumn{2}{c}{} &
\multicolumn{4}{c|}{\textbf{M3DocVQA (single doc.)}} &
\multicolumn{4}{c}{\textbf{SlideVQA}}\\
\textbf{Model} & \textbf{Method} &
\textbf{Visit Ratio} & \textbf{ANLS} & \textbf{Max VRAM (GB)} & \textbf{ANLS / VRAM} &
\textbf{Visit Ratio} & \textbf{ANLS} & \textbf{Max VRAM (GB)} & \textbf{ANLS / VRAM} \\
\midrule
Qwen 2.5 VL 3B          & MI  & 1.0 & 53.43 & 78.98 & 0.67 & 1.0 & 64.95 & 40.76 & 1.59 \\
Qwen 2.5 VL 7B          & MI  & 1.0 & 61.32 & 79.41 & 0.77 & 1.0 & 71.82 & 51.35 & 1.39 \\
Qwen 2.5 VL 32B         & MI  & 1.0 & 53.67 & 278.40 & 0.19 & 1.0 & 69.50 & 232.84 & 0.30 \\
Qwen 2.5 VL 72B         & MI  & 1.0 & 63.69 & 582.17 & 0.11 & 1.0 & 75.45 & 323.52 & 0.23 \\
\midrule
Qwen 2.5 VL 3B          & CoS          & 0.52 & 20.66 & 13.35 &  1.55 & 0.30 & 23.23 & 13.41 & 1.73 \\
Qwen 2.5 VL 72B         & CoS          & 0.59 &40.96 & 160.48 & 0.26 & 0.50 & 66.73 & 170.63 & 0.39 \\
SCoPE VLM 3B SFT (Ours)       & CoS          & 0.54 & 46.26 & 14.59 &  3.17 & 0.56 & 59.84 & 13.42 & 4.45 \\
SCoPE VLM 3B EGRPO (Ours)     & CoS          & 0.45 & 45.58 & 13.72 & 3.32 & 0.52 & 59.13 & 13.12 & 4.51 \\
\bottomrule
\end{tabular}}
\caption{ANLS scores, memory usage, and average visited‑page ratios per question for SCoPE VLM and Qwen 2.5 VL, evaluated on M3DocVQA and SlideVQA. All models are constrained to use the same maximum number of visual tokens per image.
}
\label{tab:vram}
\end{table*}

\begin{table*}[!t]
  \centering
  \resizebox{1.0\textwidth}{!}{%
  \begin{tabular}{@{}l|ccccc|ccccc|ccccc@{}}
    \toprule
    \multicolumn{1}{c|}{} &
    \multicolumn{5}{c|}{\textbf{General}} &
    \multicolumn{5}{c|}{\textbf{Web Shopping}} &
    \multicolumn{5}{c}{\textbf{Overall}}\\
    \textbf{Model} &
    \textbf{Click} & \textbf{Scroll} & \textbf{Stop} & \textbf{Total} & \textbf{Goal} &
    \textbf{Click} & \textbf{Scroll} & \textbf{Stop} & \textbf{Total} & \textbf{Goal} &
    \textbf{Click} & \textbf{Scroll} & \textbf{Stop} & \textbf{Total} & \textbf{Goal} \\
    \midrule
    Qwen 2.5‑VL‑3B            & 30.21 & 6.57 & 66.67 & 33.53 & 35.46 & 37.76 & 39.01 & 83.57 & 41.49 & 41.17 & 33.99 & 22.79 & 75.12 & 37.51 & 38.32 \\
    SCoPE VLM 3B SFT (Ours)   & 31.00 & 8.76 & 73.08 & 34.28 & 36.10 & 37.95 & 36.77 & 87.14 & 40.83 & 41.56 & 34.48 & 22.77 & 80.11 & 37.56 & 38.83 \\
    SCoPE VLM 3B EGRPO (Ours) & 33.55 & 8.03 & 76.28 & 35.69 & 37.51 & 38.31 & 45.29 & 84.29 & 42.33 & 42.32 & 35.93 & 26.66 & 80.29 & 39.01 & 39.92 \\
    \bottomrule
  \end{tabular}}
  \caption{Performance results on the AitZ~\cite{zhang2024android} across General and Web Shopping test splits. The table reports exact-match accuracies for Click, Scroll, Stop and their average (Total). Goal progress (Goal) indicates average task completion rate per step. All models are fine-tuned on the benchmark's training set prior to evaluation.}
  \label{tab:AitZ}
\end{table*}

\noindent \textbf{Challenges of CoS:} In Table~\ref{tab:main}, the baseline 3B model shows a drastic performance drop in CoS from the conventional method. Even the 72B model scores lower than the 3B model with multi-image inference on average. This shows that even the largest scale model in Qwen2.5 VL series fails to generalize decision-making performance, although it shows moderate capabilities in GUI control and document question answering. This highlights the difficulties in question answering under agentic exploration. Furthermore, CoS consistently underperforms on DocVQA, a single-image document question answering benchmark~\cite{mathew2021docvqadatasetvqadocument}. For single-page document question answering, models do not require complex agentic navigation but rather rely solely on visual understanding. This confirms the observation from MPO~\cite{wang2025enhancingreasoningabilitymultimodal} that although Chain of Thought effectively enhances logical reasoning, the distribution shift between SFT and inference degrades multimodal understanding.

\noindent \textbf{Effectiveness of SCoPE dataset:} Despite the great difficulties, our models successfully leverage the inference time scaling strategy and consistently outperform the equivalent baseline both in CoS and multi-image inference in benchmarks with the longer pages under similar memory resource limitations. In Table~\ref{tab:main}, the gains are well demonstrated on M3DocVQA and SlideVQA, where SCoPE VLM 3B SFT scores more than two times higher in ANLS (19.54 → 46.13) than the Qwen 2.5VL 3B with CoS and 5\% higher than the multi-image inference. This clearly supports the effectiveness of the proposed SCoPE dataset, showing substantial performance gains throughout the experiments. This further highlights the supremacy of the CoS framework as it performs better than the multi-image inference under similar restrictions of VRAM allocation. It is also notable that both SCoPE VLMs achieve performance comparable to the 72B model on longer documents. This shows that our models have comparable exploration abilities to the largest model in the Qwen2.5 VL series.

\begin{table*}[!t]
  \centering
  \large
  \resizebox{1.0\textwidth}{!}{%
  \begin{tabular}{@{}l cc|cc|cc cc|cc|cc cc@{}}
    \toprule
    \multicolumn{1}{c}{} &
    \multicolumn{6}{c}{\textbf{MP-DocVQA}} &
    \multicolumn{6}{c}{\textbf{M3DocVQA}} &
    \multicolumn{2}{c}{\textbf{Average}}\\
    \cmidrule(lr){2-7} \cmidrule(lr){8-13} \cmidrule(lr){14-15}
    \multicolumn{1}{c}{} &
    \multicolumn{2}{c|}{\textbf{Serial}} &
    \multicolumn{2}{c|}{\textbf{Random}} &
    \multicolumn{2}{c}{\textbf{CoS}} &
    \multicolumn{2}{c|}{\textbf{Serial}} &
    \multicolumn{2}{c|}{\textbf{Random}} &
    \multicolumn{2}{c}{\textbf{CoS}} &
    \multicolumn{2}{c}{}\\
    \textbf{Model} &
    \textbf{Visit Ratio} & \textbf{ANLS} &
    \textbf{Visit Ratio} & \textbf{ANLS} &
    \textbf{Visit Ratio} & \textbf{ANLS} &
    \textbf{Visit Ratio} & \textbf{ANLS} &
    \textbf{Visit Ratio} & \textbf{ANLS} &
    \textbf{Visit Ratio} & \textbf{ANLS} &
    \textbf{Visit Ratio} & \textbf{ANLS} \\
    \midrule
    SFT                 & 22.02 & 40.25 & 21.65 & 46.80 & 75.59 & 74.49 & 44.65 & 41.22 & 48.82 & 39.21 & 53.40 & 46.13 & 44.36 & 48.02 \\
    GRPO (Last, 2) & 20.05 & 39.83 & 20.36 & 45.20 & 71.40 & 73.14 & 31.69 & 38.51 & 35.31 & 37.30 & 46.21 & 46.71 & 37.50 & 46.78 \\
    EGRPO               & 18.85 & 39.93 & 18.50 & 44.76 & 69.97 & 73.90 & 37.14 & 39.74 & 37.80 & 41.32 & 47.20 & 47.67 & 38.24 & 47.89 \\
    \bottomrule
  \end{tabular}}
  \caption{Comparison of Visit Ratio and ANLS across inference strategies on MP-DocVQA and M3DocVQA. All models achieve higher ANLS under CoS, demonstrating effective learned search strategies. EGRPO further achieves superior navigation efficiency and task performance, validating its effectiveness in optimizing both objectives.}
  \label{tab:inference_strategy_comparison}
\end{table*}

\begin{table*}[!t]
  \centering
  \resizebox{1.0\textwidth}{!}{%
  \begin{tabular}{@{}l cccccc|cccccc@{}}
    \toprule
    & \multicolumn{6}{c|}{\textbf{ANLS (\%)}} & \multicolumn{6}{c}{\textbf{Visit Ratio (\%)}} \\
    \cmidrule(lr){2-7} \cmidrule(lr){8-13}
    \textbf{Model} &
    \textbf{MP-DocVQA} & \textbf{SlideVQA} & \textbf{M3DocVQA} & \textbf{DUDE} & \textbf{MMLong} & \textbf{Avg.} &
    \textbf{MP-DocVQA} & \textbf{SlideVQA} & \textbf{M3DocVQA} & \textbf{DUDE} & \textbf{MMLong} & \textbf{Avg.} \\
    \midrule
    SFT             & 74.49 & 59.88 & 46.13 & 42.82 & 16.89 & 48.04 & 75.59 & 55.79 & 53.40 & 82.40 & 122.68 & 77.97 \\
    GRPO (Last)     & 72.40 & 56.53 & 46.71 & 42.40 & 17.99 & 47.20 & 70.31 & 49.86 & 45.94 & 74.39 & 101.47 & 68.39 \\
    GRPO (Last, 2)  & 73.14 & 56.78 & 46.71 & 42.05 & 17.72 & 47.28 & 71.40 & 53.45 & 46.21 & 75.02 & 108.84 & 70.98 \\
    EGRPO           & 73.90 & 58.78 & 47.67 & 42.87 & 17.53 & 48.15 & 69.97 & 54.20 & 47.20 & 76.84 & 113.04 & 72.25 \\
    \midrule
    & \multicolumn{6}{c|}{\textbf{Action Success Ratio  (\%)}} & \multicolumn{6}{c}{\textbf{No Answer Ratio (\%)}} \\
    \cmidrule(lr){2-7} \cmidrule(lr){8-13}
    \textbf{Model} &
    \textbf{MP-DocVQA} & \textbf{SlideVQA} & \textbf{M3DocVQA} & \textbf{DUDE} & \textbf{MMLong} & \textbf{Avg.} &
    \textbf{MP-DocVQA} & \textbf{SlideVQA} & \textbf{M3DocVQA} & \textbf{DUDE} & \textbf{MMLong} & \textbf{Avg.} \\
    \midrule
    SFT             & 94.42 & 78.49 & 87.39 & 90.90 & 62.52 & 82.74 & 1.33 & 0.79 & 3.50 & 3.71 & 31.44 & 8.15 \\
    GRPO (Last)     & 95.78 & 80.10 & 91.02 & 92.69 & 65.55 & 85.02 & 0.46 & 0.42 & 1.40 & 2.95 & 18.06 & 4.66 \\
    GRPO (Last, 2)  & 96.02 & 79.97 & 91.33 & 93.11 & 63.98 & 84.88 & 0.64 & 0.30 & 1.40 & 3.47 & 24.47 & 6.06 \\
    EGRPO           & 95.44 & 79.33 & 89.69 & 92.77 & 65.84 & 84.61 & 0.62 & 0.48 & 2.33 & 5.73 & 27.04 & 7.24 \\
    \bottomrule
  \end{tabular}}
  \caption{Performance-efficiency comparison of SFT, GRPO, and EGRPO on the Chain of Scroll framework across five document understanding benchmarks. EGRPO maintains task performance comparable to SFT while improving efficiency metrics, whereas GRPO methods exhibit performance-efficiency trade-offs with reduced ANLS scores.}
  \label{tab:training_method_comparison}
\end{table*}

\noindent \textbf{Memory Efficiency of SCoPE VLM:} CoS ultimately transforms the multi-page document question answering problem into a single image. These characteristics of the CoS framework bring a significant increase in the performance per memory usage. Table~\ref{tab:vram} highlights the memory efficiency of SCoPE VLMs, which shows more than two times higher ANLS / VRAM on both benchmarks. In addition, SCoPE VLMs show significantly lower VRAM usage to complete the benchmarks, which is about three to five times less than the conventional method requires. Compared to the baseline, SCoPE VLM SFT exhibits an increase in visit ratio due to the enhanced exploration capabilities, which results in an increase in overall performance. Moreover, the additional EGRPO stage successfully mitigates this issue by reducing the visit ratio. This is demonstrated on M3DocVQA, where EGRPO achieves approximately 16.67\% improvement over SFT. Despite fewer page visits, SCoPE VLM EGRPO maintains performance in ANLS, indicating its improved reasoning and decision-making capabilities.

\noindent \textbf{Effectiveness of Document Navigation Capabilities on GUI Control:}
To evaluate whether the multimodal agentic reasoning and decision-making capabilities learned from Chain of Scroll generalize to other downstream tasks, we conducted experiments on the AitZ dataset, a mobile GUI control dataset. We focused on two key actions that serve as direct proxies for core competencies: Scroll actions for assessing navigation proficiency and Stop actions for measuring task completion accuracy. These actions provide ideal benchmarks for validating the transferability of our model's navigation capabilities to new domains. As shown in Table~\ref{tab:AitZ}, the two SCoPE VLM variants demonstrate strong adaptation performance. Especially, the SCoPE VLM EGRPO exhibits better adaptation than the SFT variant, achieving an overall average exact-match accuracy of 39.01\%. This performance gain stems from substantial improvements over Qwen2.5 VL 3B in both targeted actions: Stop accuracy increased from 75.12\% to 80.29\%, while Scroll accuracy improved from 22.79\% to 26.66\%. These results provide compelling evidence that our document navigation capabilities can be effectively generalized to GUI control across different contexts. Although the performance gains in Scroll and Stop accuracy between SFT and EGRPO are mixed, EGRPO  demonstrates better adaptation to ground truth actions overall and achieves notably higher goal progress rates. This further confirms the effectiveness of learning the transition from searching steps to the terminal step in multimodal agents. Detailed experimental setup and comprehensive results are provided in Sections~\ref{sec:evaluation_setup} and~\ref{GUI-result-appendix}, respectively.

\subsection{Ablation study}
To evaluate the effectiveness of the Chain of Scroll (CoS) framework and EGRPO, we conducted an ablation study as presented in Table~\ref{tab:inference_strategy_comparison} and Table~\ref{tab:training_method_comparison}. In Table~\ref{tab:inference_strategy_comparison}, we compare CoS with two alternative strategies: serial and random processing. We implement random navigation by replacing the model's scroll action output with a random scroll to assess whether our models learned effective search trajectories or simply visited pages without meaningful selection. In Table~\ref{tab:training_method_comparison}, we demonstrate the effectiveness of EGRPO by comparing it with SFT and GRPO variants. EGRPO and GRPO variant models in both tables are trained for a reduced step count of 1,000 with the same effective group size for gradient updates. Further details are provided in Section~\ref{sec:evaluation_setup} in the appendix.

\noindent \textbf{Inference Strategy Comparison:}
The evaluation results in Table~\ref{tab:inference_strategy_comparison} demonstrate that CoS consistently outperforms both Serial and Random approaches across the MP-DocVQA and M3DocVQA benchmarks. For MP-DocVQA, all compared models using CoS achieve significantly higher ANLS scores than the serial and random methods. For instance, the SFT model improves from 40.25 (Serial) and 46.80 (Random) to 74.49 (CoS). This substantial improvement demonstrates that the models have learned effective search strategies, and this trend is consistently observed regardless of the post-training method employed. A similar trend is observed on M3DocVQA, where the SFT model improves from 41.22 (Serial) to 46.13 (CoS), with GRPO and EGRPO showing a similar tendency. This clearly demonstrates that all of the SCoPE VLM variants successfully learned effective context-aware, action-based reasoning for document navigation. Furthermore, EGRPO consistently shows lower page visit ratios compared to GRPO while maintaining high performance across all three inference strategies. This validates its effectiveness in optimizing both navigation efficiency and task performance.

\noindent \textbf{Training Method Comparison:}
The extended analysis in Table~\ref{tab:training_method_comparison} further confirms EGRPO's effectiveness in addressing the limitations of SFT and vanilla GRPO variants. To validate the contributions of terminal step projection from EGRPO, we compared EGRPO with GRPO trained using only the terminal step (Last) and both the terminal and penultimate steps (Last, 2) as noted in the table.

In Table~\ref{tab:training_method_comparison}, the EGRPO model performs on par with SFT in ANLS scores (48.15 vs. 48.04 average) while significantly improving navigation efficiency. EGRPO reduces the average visit ratio from 77.97 to 72.25. The Action Success Ratio results show that EGRPO achieves consistent improvements in legal action selection. GRPO variants show improvements in the objectives but experience a clear performance drop. This is evidenced by MP-DocVQA and SlideVQA, where the GRPO (Last) model drops from 74.49 to 72.40 in MP-DocVQA and from 59.88 to 56.53 in SlideVQA.

Although the vanilla GRPO variants significantly sacrifice performance, they show better optimization in average page visits per question and answer return rate. Regarding the No Answer Ratio, vanilla GRPO variants optimize this metric more aggressively than EGRPO. GRPO (Last) achieved a No Answer Ratio of 4.66, whereas EGRPO achieved 7.24. However, this aggressive optimization comes at the cost of reduced ANLS performance. This performance gap suggests that GRPO variants are more prone to reward hacking and overfitting, leading to premature or speculative answers rather than thorough document exploration. In contrast, EGRPO delays returning an answer until it finds definitive evidence, resulting in more accurate responses overall. While the improvement in No Answer Ratio is more modest for EGRPO compared to GRPO variants, it still notably reduces this ratio compared to the SFT baseline, from 8.15 to 7.24. This tendency is also observed in SCoPE VLM SFT and EGRPO. The performance-efficiency comparison of SCoPE VLMs and baseline models in Table~\ref{tab:main} is detailed in Section~\ref{sec:main-performance-efficiency-appendix} in the appendix.

These results validate that EGRPO effectively addresses the fundamental challenge of balancing exploration and exploitation in agentic document navigation. While vanilla GRPO variants achieve aggressive optimization of efficiency metrics, they do so at the expense of task performance, exhibiting patterns that prioritize quick termination over thorough evidence gathering. In contrast, EGRPO's terminal step projection mechanism enables efficient backward propagation of reward signals through trajectories, delivering focused optimization toward answer-critical decisions while preserving the sample-efficient grouped updates of GRPO. This design enables EGRPO to achieve balanced optimization essential in real-world long document understanding scenarios.

\section{Conclusion}

In conclusion, this work addresses a critical gap in VLMs by introducing Chain of Scroll, the first framework to model agentic behaviors in multimodal document understanding. Through SCoPE dataset, we substantially improve the basic decision-making capabilities. With the Episodic Group Relative Policy Optimization, SCoPE VLM learns better Chain of Scroll trajectories, enabling effective long document question answering.

Through comprehensive evaluations, SCoPE VLM achieves high memory efficiency and competitive accuracy compared to state-of-the-art VLMs. Despite using fewer parameters and substantially less memory, our model demonstrates promising performance across benchmarks, underscoring the value of selective document exploration over exhaustive context processing. Furthermore, the strong adaptation to GUI control tasks validates the importance of document-based navigation pretraining for developing transferable agentic capabilities. Our ablation study confirms that our models learned effective reasoning and decision-making capabilities, while the additional EGRPO stage further optimizes navigation behavior by balancing efficiency with performance. By introducing an action-driven approach to document understanding, our work represents a significant step toward more capable and efficient multimodal agents.

\section{Limitations}
The major limitations of this work can be summarized in three key aspects: the scale of the training data, the variance in document lengths across the training and benchmark datasets, and the constrained demonstration of EGRPO's potential due to the use of LoRA. First, the training data is notably limited, especially for final answer pages, which are restricted by the number of available questions (approximately 19K). When distributed across intermediate scroll steps, the number of answer-step samples becomes insufficient to fully train a 3B model from the SFT stage. In addition, due to the limited number of training document samples, there is a risk of amplifying the bias present in the training data. Using Gemini series alone for the annotation model may also have introduced bias in the generated reasoning. Second, both the training and benchmark datasets primarily consist of documents with fewer than 100 pages. This calls for further research to explore behavioral differences between short documents and longer, more contextually extensive documents. Furthermore, while LoRA helps reduce the training burden, it inherently restricts the potential of Episodic Group Relative Policy Optimization.

Another limitation concerns the trade-off between domain-specific adaptation and general capability. Recent works~\cite{hu2024mplugdocowl2highresolutioncompressingocrfree, zhai2024finetuninglargevisionlanguagemodels} explore parameter updates primarily for document understanding or agentic tasks, often at the sacrifice of general ability for domain-specific adaptation. While SCoPE VLM may compromise generality just as those works, this can be mitigated through two strategies: (1) employing larger-scale base models that can generalize better even without additional training steps, as they already possess agentic decision-making capabilities, or (2) using blended SFT and RL that combine domain-specific and general-purpose corpora. Despite our method remaining suitable for local deployment due to its efficient parameter footprint, the potential trade-off between specialization and generalization in our domain-adapted sLLMs through the proposed Chain of Scroll framework requires further investigation.

Additionally, our current scope is limited to single-document QA. While SCoPE VLM supports multi-hop reasoning within a single document by actively performing memory-based aggregation during recursive scrolling, we do not address cross-document reasoning. Cross-document reasoning represents an important capability gap and constitutes an exciting extension for future work.

\bibliography{custom}

@misc{OpenAI2023GPT4V,
  title={GPT-4V},
  author={OpenAI},
  year={2023},
  url={https://openai.com/research/gpt-4v-system-card}
}

@misc{openai2024gpt4ocard,
      title={GPT-4o System Card}, 
      author={OpenAI and : and Aaron Hurst and Adam Lerer and Adam P. Goucher and Adam Perelman and Aditya Ramesh and Aidan Clark and AJ Ostrow and Akila Welihinda and Alan Hayes and Alec Radford and Aleksander Mądry and Alex Baker-Whitcomb and Alex Beutel and Alex Borzunov and Alex Carney and Alex Chow and Alex Kirillov and Alex Nichol and Alex Paino and Alex Renzin and Alex Tachard Passos and Alexander Kirillov and Alexi Christakis and Alexis Conneau and Ali Kamali and Allan Jabri and Allison Moyer and Allison Tam and Amadou Crookes and Amin Tootoochian and Amin Tootoonchian and Ananya Kumar and Andrea Vallone and Andrej Karpathy and Andrew Braunstein and Andrew Cann and Andrew Codispoti and Andrew Galu and Andrew Kondrich and Andrew Tulloch and Andrey Mishchenko and Angela Baek and Angela Jiang and Antoine Pelisse and Antonia Woodford and Anuj Gosalia and Arka Dhar and Ashley Pantuliano and Avi Nayak and Avital Oliver and Barret Zoph and Behrooz Ghorbani and Ben Leimberger and Ben Rossen and Ben Sokolowsky and Ben Wang and Benjamin Zweig and Beth Hoover and Blake Samic and Bob McGrew and Bobby Spero and Bogo Giertler and Bowen Cheng and Brad Lightcap and Brandon Walkin and Brendan Quinn and Brian Guarraci and Brian Hsu and Bright Kellogg and Brydon Eastman and Camillo Lugaresi and Carroll Wainwright and Cary Bassin and Cary Hudson and Casey Chu and Chad Nelson and Chak Li and Chan Jun Shern and Channing Conger and Charlotte Barette and Chelsea Voss and Chen Ding and Cheng Lu and Chong Zhang and Chris Beaumont and Chris Hallacy and Chris Koch and Christian Gibson and Christina Kim and Christine Choi and Christine McLeavey and Christopher Hesse and Claudia Fischer and Clemens Winter and Coley Czarnecki and Colin Jarvis and Colin Wei and Constantin Koumouzelis and Dane Sherburn and Daniel Kappler and Daniel Levin and Daniel Levy and David Carr and David Farhi and David Mely and David Robinson and David Sasaki and Denny Jin and Dev Valladares and Dimitris Tsipras and Doug Li and Duc Phong Nguyen and Duncan Findlay and Edede Oiwoh and Edmund Wong and Ehsan Asdar and Elizabeth Proehl and Elizabeth Yang and Eric Antonow and Eric Kramer and Eric Peterson and Eric Sigler and Eric Wallace and Eugene Brevdo and Evan Mays and Farzad Khorasani and Felipe Petroski Such and Filippo Raso and Francis Zhang and Fred von Lohmann and Freddie Sulit and Gabriel Goh and Gene Oden and Geoff Salmon and Giulio Starace and Greg Brockman and Hadi Salman and Haiming Bao and Haitang Hu and Hannah Wong and Haoyu Wang and Heather Schmidt and Heather Whitney and Heewoo Jun and Hendrik Kirchner and Henrique Ponde de Oliveira Pinto and Hongyu Ren and Huiwen Chang and Hyung Won Chung and Ian Kivlichan and Ian O'Connell and Ian O'Connell and Ian Osband and Ian Silber and Ian Sohl and Ibrahim Okuyucu and Ikai Lan and Ilya Kostrikov and Ilya Sutskever and Ingmar Kanitscheider and Ishaan Gulrajani and Jacob Coxon and Jacob Menick and Jakub Pachocki and James Aung and James Betker and James Crooks and James Lennon and Jamie Kiros and Jan Leike and Jane Park and Jason Kwon and Jason Phang and Jason Teplitz and Jason Wei and Jason Wolfe and Jay Chen and Jeff Harris and Jenia Varavva and Jessica Gan Lee and Jessica Shieh and Ji Lin and Jiahui Yu and Jiayi Weng and Jie Tang and Jieqi Yu and Joanne Jang and Joaquin Quinonero Candela and Joe Beutler and Joe Landers and Joel Parish and Johannes Heidecke and John Schulman and Jonathan Lachman and Jonathan McKay and Jonathan Uesato and Jonathan Ward and Jong Wook Kim and Joost Huizinga and Jordan Sitkin and Jos Kraaijeveld and Josh Gross and Josh Kaplan and Josh Snyder and Joshua Achiam and Joy Jiao and Joyce Lee and Juntang Zhuang and Justyn Harriman and Kai Fricke and Kai Hayashi and Karan Singhal and Katy Shi and Kavin Karthik and Kayla Wood and Kendra Rimbach and Kenny Hsu and Kenny Nguyen and Keren Gu-Lemberg and Kevin Button and Kevin Liu and Kiel Howe and Krithika Muthukumar and Kyle Luther and Lama Ahmad and Larry Kai and Lauren Itow and Lauren Workman and Leher Pathak and Leo Chen and Li Jing and Lia Guy and Liam Fedus and Liang Zhou and Lien Mamitsuka and Lilian Weng and Lindsay McCallum and Lindsey Held and Long Ouyang and Louis Feuvrier and Lu Zhang and Lukas Kondraciuk and Lukasz Kaiser and Luke Hewitt and Luke Metz and Lyric Doshi and Mada Aflak and Maddie Simens and Madelaine Boyd and Madeleine Thompson and Marat Dukhan and Mark Chen and Mark Gray and Mark Hudnall and Marvin Zhang and Marwan Aljubeh and Mateusz Litwin and Matthew Zeng and Max Johnson and Maya Shetty and Mayank Gupta and Meghan Shah and Mehmet Yatbaz and Meng Jia Yang and Mengchao Zhong and Mia Glaese and Mianna Chen and Michael Janner and Michael Lampe and Michael Petrov and Michael Wu and Michele Wang and Michelle Fradin and Michelle Pokrass and Miguel Castro and Miguel Oom Temudo de Castro and Mikhail Pavlov and Miles Brundage and Miles Wang and Minal Khan and Mira Murati and Mo Bavarian and Molly Lin and Murat Yesildal and Nacho Soto and Natalia Gimelshein and Natalie Cone and Natalie Staudacher and Natalie Summers and Natan LaFontaine and Neil Chowdhury and Nick Ryder and Nick Stathas and Nick Turley and Nik Tezak and Niko Felix and Nithanth Kudige and Nitish Keskar and Noah Deutsch and Noel Bundick and Nora Puckett and Ofir Nachum and Ola Okelola and Oleg Boiko and Oleg Murk and Oliver Jaffe and Olivia Watkins and Olivier Godement and Owen Campbell-Moore and Patrick Chao and Paul McMillan and Pavel Belov and Peng Su and Peter Bak and Peter Bakkum and Peter Deng and Peter Dolan and Peter Hoeschele and Peter Welinder and Phil Tillet and Philip Pronin and Philippe Tillet and Prafulla Dhariwal and Qiming Yuan and Rachel Dias and Rachel Lim and Rahul Arora and Rajan Troll and Randall Lin and Rapha Gontijo Lopes and Raul Puri and Reah Miyara and Reimar Leike and Renaud Gaubert and Reza Zamani and Ricky Wang and Rob Donnelly and Rob Honsby and Rocky Smith and Rohan Sahai and Rohit Ramchandani and Romain Huet and Rory Carmichael and Rowan Zellers and Roy Chen and Ruby Chen and Ruslan Nigmatullin and Ryan Cheu and Saachi Jain and Sam Altman and Sam Schoenholz and Sam Toizer and Samuel Miserendino and Sandhini Agarwal and Sara Culver and Scott Ethersmith and Scott Gray and Sean Grove and Sean Metzger and Shamez Hermani and Shantanu Jain and Shengjia Zhao and Sherwin Wu and Shino Jomoto and Shirong Wu and Shuaiqi and Xia and Sonia Phene and Spencer Papay and Srinivas Narayanan and Steve Coffey and Steve Lee and Stewart Hall and Suchir Balaji and Tal Broda and Tal Stramer and Tao Xu and Tarun Gogineni and Taya Christianson and Ted Sanders and Tejal Patwardhan and Thomas Cunninghman and Thomas Degry and Thomas Dimson and Thomas Raoux and Thomas Shadwell and Tianhao Zheng and Todd Underwood and Todor Markov and Toki Sherbakov and Tom Rubin and Tom Stasi and Tomer Kaftan and Tristan Heywood and Troy Peterson and Tyce Walters and Tyna Eloundou and Valerie Qi and Veit Moeller and Vinnie Monaco and Vishal Kuo and Vlad Fomenko and Wayne Chang and Weiyi Zheng and Wenda Zhou and Wesam Manassra and Will Sheu and Wojciech Zaremba and Yash Patil and Yilei Qian and Yongjik Kim and Youlong Cheng and Yu Zhang and Yuchen He and Yuchen Zhang and Yujia Jin and Yunxing Dai and Yury Malkov},
      year={2024},
      eprint={2410.21276},
      archivePrefix={arXiv},
      primaryClass={cs.CL},
      url={https://arxiv.org/abs/2410.21276}, 
}

@misc{geminiteam2024gemini15unlockingmultimodal,
      title={Gemini 1.5: Unlocking multimodal understanding across millions of tokens of context}, 
      author={Gemini Team and Petko Georgiev and Ving Ian Lei and Ryan Burnell and Libin Bai and Anmol Gulati and Garrett Tanzer and Damien Vincent and Zhufeng Pan and Shibo Wang and Soroosh Mariooryad and Yifan Ding and Xinyang Geng and Fred Alcober and Roy Frostig and Mark Omernick and Lexi Walker and Cosmin Paduraru and Christina Sorokin and Andrea Tacchetti and Colin Gaffney and Samira Daruki and Olcan Sercinoglu and Zach Gleicher and Juliette Love and Paul Voigtlaender and Rohan Jain and Gabriela Surita and Kareem Mohamed and Rory Blevins and Junwhan Ahn and Tao Zhu and Kornraphop Kawintiranon and Orhan Firat and Yiming Gu and Yujing Zhang and Matthew Rahtz and Manaal Faruqui and Natalie Clay and Justin Gilmer and JD Co-Reyes and Ivo Penchev and Rui Zhu and Nobuyuki Morioka and Kevin Hui and Krishna Haridasan and Victor Campos and Mahdis Mahdieh and Mandy Guo and Samer Hassan and Kevin Kilgour and Arpi Vezer and Heng-Tze Cheng and Raoul de Liedekerke and Siddharth Goyal and Paul Barham and DJ Strouse and Seb Noury and Jonas Adler and Mukund Sundararajan and Sharad Vikram and Dmitry Lepikhin and Michela Paganini and Xavier Garcia and Fan Yang and Dasha Valter and Maja Trebacz and Kiran Vodrahalli and Chulayuth Asawaroengchai and Roman Ring and Norbert Kalb and Livio Baldini Soares and Siddhartha Brahma and David Steiner and Tianhe Yu and Fabian Mentzer and Antoine He and Lucas Gonzalez and Bibo Xu and Raphael Lopez Kaufman and Laurent El Shafey and Junhyuk Oh and Tom Hennigan and George van den Driessche and Seth Odoom and Mario Lucic and Becca Roelofs and Sid Lall and Amit Marathe and Betty Chan and Santiago Ontanon and Luheng He and Denis Teplyashin and Jonathan Lai and Phil Crone and Bogdan Damoc and Lewis Ho and Sebastian Riedel and Karel Lenc and Chih-Kuan Yeh and Aakanksha Chowdhery and Yang Xu and Mehran Kazemi and Ehsan Amid and Anastasia Petrushkina and Kevin Swersky and Ali Khodaei and Gowoon Chen and Chris Larkin and Mario Pinto and Geng Yan and Adria Puigdomenech Badia and Piyush Patil and Steven Hansen and Dave Orr and Sebastien M. R. Arnold and Jordan Grimstad and Andrew Dai and Sholto Douglas and Rishika Sinha and Vikas Yadav and Xi Chen and Elena Gribovskaya and Jacob Austin and Jeffrey Zhao and Kaushal Patel and Paul Komarek and Sophia Austin and Sebastian Borgeaud and Linda Friso and Abhimanyu Goyal and Ben Caine and Kris Cao and Da-Woon Chung and Matthew Lamm and Gabe Barth-Maron and Thais Kagohara and Kate Olszewska and Mia Chen and Kaushik Shivakumar and Rishabh Agarwal and Harshal Godhia and Ravi Rajwar and Javier Snaider and Xerxes Dotiwalla and Yuan Liu and Aditya Barua and Victor Ungureanu and Yuan Zhang and Bat-Orgil Batsaikhan and Mateo Wirth and James Qin and Ivo Danihelka and Tulsee Doshi and Martin Chadwick and Jilin Chen and Sanil Jain and Quoc Le and Arjun Kar and Madhu Gurumurthy and Cheng Li and Ruoxin Sang and Fangyu Liu and Lampros Lamprou and Rich Munoz and Nathan Lintz and Harsh Mehta and Heidi Howard and Malcolm Reynolds and Lora Aroyo and Quan Wang and Lorenzo Blanco and Albin Cassirer and Jordan Griffith and Dipanjan Das and Stephan Lee and Jakub Sygnowski and Zach Fisher and James Besley and Richard Powell and Zafarali Ahmed and Dominik Paulus and David Reitter and Zalan Borsos and Rishabh Joshi and Aedan Pope and Steven Hand and Vittorio Selo and Vihan Jain and Nikhil Sethi and Megha Goel and Takaki Makino and Rhys May and Zhen Yang and Johan Schalkwyk and Christina Butterfield and Anja Hauth and Alex Goldin and Will Hawkins and Evan Senter and Sergey Brin and Oliver Woodman and Marvin Ritter and Eric Noland and Minh Giang and Vijay Bolina and Lisa Lee and Tim Blyth and Ian Mackinnon and Machel Reid and Obaid Sarvana and David Silver and Alexander Chen and Lily Wang and Loren Maggiore and Oscar Chang and Nithya Attaluri and Gregory Thornton and Chung-Cheng Chiu and Oskar Bunyan and Nir Levine and Timothy Chung and Evgenii Eltyshev and Xiance Si and Timothy Lillicrap and Demetra Brady and Vaibhav Aggarwal and Boxi Wu and Yuanzhong Xu and Ross McIlroy and Kartikeya Badola and Paramjit Sandhu and Erica Moreira and Wojciech Stokowiec and Ross Hemsley and Dong Li and Alex Tudor and Pranav Shyam and Elahe Rahimtoroghi and Salem Haykal and Pablo Sprechmann and Xiang Zhou and Diana Mincu and Yujia Li and Ravi Addanki and Kalpesh Krishna and Xiao Wu and Alexandre Frechette and Matan Eyal and Allan Dafoe and Dave Lacey and Jay Whang and Thi Avrahami and Ye Zhang and Emanuel Taropa and Hanzhao Lin and Daniel Toyama and Eliza Rutherford and Motoki Sano and HyunJeong Choe and Alex Tomala and Chalence Safranek-Shrader and Nora Kassner and Mantas Pajarskas and Matt Harvey and Sean Sechrist and Meire Fortunato and Christina Lyu and Gamaleldin Elsayed and Chenkai Kuang and James Lottes and Eric Chu and Chao Jia and Chih-Wei Chen and Peter Humphreys and Kate Baumli and Connie Tao and Rajkumar Samuel and Cicero Nogueira dos Santos and Anders Andreassen and Nemanja Rakićević and Dominik Grewe and Aviral Kumar and Stephanie Winkler and Jonathan Caton and Andrew Brock and Sid Dalmia and Hannah Sheahan and Iain Barr and Yingjie Miao and Paul Natsev and Jacob Devlin and Feryal Behbahani and Flavien Prost and Yanhua Sun and Artiom Myaskovsky and Thanumalayan Sankaranarayana Pillai and Dan Hurt and Angeliki Lazaridou and Xi Xiong and Ce Zheng and Fabio Pardo and Xiaowei Li and Dan Horgan and Joe Stanton and Moran Ambar and Fei Xia and Alejandro Lince and Mingqiu Wang and Basil Mustafa and Albert Webson and Hyo Lee and Rohan Anil and Martin Wicke and Timothy Dozat and Abhishek Sinha and Enrique Piqueras and Elahe Dabir and Shyam Upadhyay and Anudhyan Boral and Lisa Anne Hendricks and Corey Fry and Josip Djolonga and Yi Su and Jake Walker and Jane Labanowski and Ronny Huang and Vedant Misra and Jeremy Chen and RJ Skerry-Ryan and Avi Singh and Shruti Rijhwani and Dian Yu and Alex Castro-Ros and Beer Changpinyo and Romina Datta and Sumit Bagri and Arnar Mar Hrafnkelsson and Marcello Maggioni and Daniel Zheng and Yury Sulsky and Shaobo Hou and Tom Le Paine and Antoine Yang and Jason Riesa and Dominika Rogozinska and Dror Marcus and Dalia El Badawy and Qiao Zhang and Luyu Wang and Helen Miller and Jeremy Greer and Lars Lowe Sjos and Azade Nova and Heiga Zen and Rahma Chaabouni and Mihaela Rosca and Jiepu Jiang and Charlie Chen and Ruibo Liu and Tara Sainath and Maxim Krikun and Alex Polozov and Jean-Baptiste Lespiau and Josh Newlan and Zeyncep Cankara and Soo Kwak and Yunhan Xu and Phil Chen and Andy Coenen and Clemens Meyer and Katerina Tsihlas and Ada Ma and Juraj Gottweis and Jinwei Xing and Chenjie Gu and Jin Miao and Christian Frank and Zeynep Cankara and Sanjay Ganapathy and Ishita Dasgupta and Steph Hughes-Fitt and Heng Chen and David Reid and Keran Rong and Hongmin Fan and Joost van Amersfoort and Vincent Zhuang and Aaron Cohen and Shixiang Shane Gu and Anhad Mohananey and Anastasija Ilic and Taylor Tobin and John Wieting and Anna Bortsova and Phoebe Thacker and Emma Wang and Emily Caveness and Justin Chiu and Eren Sezener and Alex Kaskasoli and Steven Baker and Katie Millican and Mohamed Elhawaty and Kostas Aisopos and Carl Lebsack and Nathan Byrd and Hanjun Dai and Wenhao Jia and Matthew Wiethoff and Elnaz Davoodi and Albert Weston and Lakshman Yagati and Arun Ahuja and Isabel Gao and Golan Pundak and Susan Zhang and Michael Azzam and Khe Chai Sim and Sergi Caelles and James Keeling and Abhanshu Sharma and Andy Swing and YaGuang Li and Chenxi Liu and Carrie Grimes Bostock and Yamini Bansal and Zachary Nado and Ankesh Anand and Josh Lipschultz and Abhijit Karmarkar and Lev Proleev and Abe Ittycheriah and Soheil Hassas Yeganeh and George Polovets and Aleksandra Faust and Jiao Sun and Alban Rrustemi and Pen Li and Rakesh Shivanna and Jeremiah Liu and Chris Welty and Federico Lebron and Anirudh Baddepudi and Sebastian Krause and Emilio Parisotto and Radu Soricut and Zheng Xu and Dawn Bloxwich and Melvin Johnson and Behnam Neyshabur and Justin Mao-Jones and Renshen Wang and Vinay Ramasesh and Zaheer Abbas and Arthur Guez and Constant Segal and Duc Dung Nguyen and James Svensson and Le Hou and Sarah York and Kieran Milan and Sophie Bridgers and Wiktor Gworek and Marco Tagliasacchi and James Lee-Thorp and Michael Chang and Alexey Guseynov and Ale Jakse Hartman and Michael Kwong and Ruizhe Zhao and Sheleem Kashem and Elizabeth Cole and Antoine Miech and Richard Tanburn and Mary Phuong and Filip Pavetic and Sebastien Cevey and Ramona Comanescu and Richard Ives and Sherry Yang and Cosmo Du and Bo Li and Zizhao Zhang and Mariko Iinuma and Clara Huiyi Hu and Aurko Roy and Shaan Bijwadia and Zhenkai Zhu and Danilo Martins and Rachel Saputro and Anita Gergely and Steven Zheng and Dawei Jia and Ioannis Antonoglou and Adam Sadovsky and Shane Gu and Yingying Bi and Alek Andreev and Sina Samangooei and Mina Khan and Tomas Kocisky and Angelos Filos and Chintu Kumar and Colton Bishop and Adams Yu and Sarah Hodkinson and Sid Mittal and Premal Shah and Alexandre Moufarek and Yong Cheng and Adam Bloniarz and Jaehoon Lee and Pedram Pejman and Paul Michel and Stephen Spencer and Vladimir Feinberg and Xuehan Xiong and Nikolay Savinov and Charlotte Smith and Siamak Shakeri and Dustin Tran and Mary Chesus and Bernd Bohnet and George Tucker and Tamara von Glehn and Carrie Muir and Yiran Mao and Hideto Kazawa and Ambrose Slone and Kedar Soparkar and Disha Shrivastava and James Cobon-Kerr and Michael Sharman and Jay Pavagadhi and Carlos Araya and Karolis Misiunas and Nimesh Ghelani and Michael Laskin and David Barker and Qiujia Li and Anton Briukhov and Neil Houlsby and Mia Glaese and Balaji Lakshminarayanan and Nathan Schucher and Yunhao Tang and Eli Collins and Hyeontaek Lim and Fangxiaoyu Feng and Adria Recasens and Guangda Lai and Alberto Magni and Nicola De Cao and Aditya Siddhant and Zoe Ashwood and Jordi Orbay and Mostafa Dehghani and Jenny Brennan and Yifan He and Kelvin Xu and Yang Gao and Carl Saroufim and James Molloy and Xinyi Wu and Seb Arnold and Solomon Chang and Julian Schrittwieser and Elena Buchatskaya and Soroush Radpour and Martin Polacek and Skye Giordano and Ankur Bapna and Simon Tokumine and Vincent Hellendoorn and Thibault Sottiaux and Sarah Cogan and Aliaksei Severyn and Mohammad Saleh and Shantanu Thakoor and Laurent Shefey and Siyuan Qiao and Meenu Gaba and Shuo-yiin Chang and Craig Swanson and Biao Zhang and Benjamin Lee and Paul Kishan Rubenstein and Gan Song and Tom Kwiatkowski and Anna Koop and Ajay Kannan and David Kao and Parker Schuh and Axel Stjerngren and Golnaz Ghiasi and Gena Gibson and Luke Vilnis and Ye Yuan and Felipe Tiengo Ferreira and Aishwarya Kamath and Ted Klimenko and Ken Franko and Kefan Xiao and Indro Bhattacharya and Miteyan Patel and Rui Wang and Alex Morris and Robin Strudel and Vivek Sharma and Peter Choy and Sayed Hadi Hashemi and Jessica Landon and Mara Finkelstein and Priya Jhakra and Justin Frye and Megan Barnes and Matthew Mauger and Dennis Daun and Khuslen Baatarsukh and Matthew Tung and Wael Farhan and Henryk Michalewski and Fabio Viola and Felix de Chaumont Quitry and Charline Le Lan and Tom Hudson and Qingze Wang and Felix Fischer and Ivy Zheng and Elspeth White and Anca Dragan and Jean-baptiste Alayrac and Eric Ni and Alexander Pritzel and Adam Iwanicki and Michael Isard and Anna Bulanova and Lukas Zilka and Ethan Dyer and Devendra Sachan and Srivatsan Srinivasan and Hannah Muckenhirn and Honglong Cai and Amol Mandhane and Mukarram Tariq and Jack W. Rae and Gary Wang and Kareem Ayoub and Nicholas FitzGerald and Yao Zhao and Woohyun Han and Chris Alberti and Dan Garrette and Kashyap Krishnakumar and Mai Gimenez and Anselm Levskaya and Daniel Sohn and Josip Matak and Inaki Iturrate and Michael B. Chang and Jackie Xiang and Yuan Cao and Nishant Ranka and Geoff Brown and Adrian Hutter and Vahab Mirrokni and Nanxin Chen and Kaisheng Yao and Zoltan Egyed and Francois Galilee and Tyler Liechty and Praveen Kallakuri and Evan Palmer and Sanjay Ghemawat and Jasmine Liu and David Tao and Chloe Thornton and Tim Green and Mimi Jasarevic and Sharon Lin and Victor Cotruta and Yi-Xuan Tan and Noah Fiedel and Hongkun Yu and Ed Chi and Alexander Neitz and Jens Heitkaemper and Anu Sinha and Denny Zhou and Yi Sun and Charbel Kaed and Brice Hulse and Swaroop Mishra and Maria Georgaki and Sneha Kudugunta and Clement Farabet and Izhak Shafran and Daniel Vlasic and Anton Tsitsulin and Rajagopal Ananthanarayanan and Alen Carin and Guolong Su and Pei Sun and Shashank V and Gabriel Carvajal and Josef Broder and Iulia Comsa and Alena Repina and William Wong and Warren Weilun Chen and Peter Hawkins and Egor Filonov and Lucia Loher and Christoph Hirnschall and Weiyi Wang and Jingchen Ye and Andrea Burns and Hardie Cate and Diana Gage Wright and Federico Piccinini and Lei Zhang and Chu-Cheng Lin and Ionel Gog and Yana Kulizhskaya and Ashwin Sreevatsa and Shuang Song and Luis C. Cobo and Anand Iyer and Chetan Tekur and Guillermo Garrido and Zhuyun Xiao and Rupert Kemp and Huaixiu Steven Zheng and Hui Li and Ananth Agarwal and Christel Ngani and Kati Goshvadi and Rebeca Santamaria-Fernandez and Wojciech Fica and Xinyun Chen and Chris Gorgolewski and Sean Sun and Roopal Garg and Xinyu Ye and S. M. Ali Eslami and Nan Hua and Jon Simon and Pratik Joshi and Yelin Kim and Ian Tenney and Sahitya Potluri and Lam Nguyen Thiet and Quan Yuan and Florian Luisier and Alexandra Chronopoulou and Salvatore Scellato and Praveen Srinivasan and Minmin Chen and Vinod Koverkathu and Valentin Dalibard and Yaming Xu and Brennan Saeta and Keith Anderson and Thibault Sellam and Nick Fernando and Fantine Huot and Junehyuk Jung and Mani Varadarajan and Michael Quinn and Amit Raul and Maigo Le and Ruslan Habalov and Jon Clark and Komal Jalan and Kalesha Bullard and Achintya Singhal and Thang Luong and Boyu Wang and Sujeevan Rajayogam and Julian Eisenschlos and Johnson Jia and Daniel Finchelstein and Alex Yakubovich and Daniel Balle and Michael Fink and Sameer Agarwal and Jing Li and Dj Dvijotham and Shalini Pal and Kai Kang and Jaclyn Konzelmann and Jennifer Beattie and Olivier Dousse and Diane Wu and Remi Crocker and Chen Elkind and Siddhartha Reddy Jonnalagadda and Jong Lee and Dan Holtmann-Rice and Krystal Kallarackal and Rosanne Liu and Denis Vnukov and Neera Vats and Luca Invernizzi and Mohsen Jafari and Huanjie Zhou and Lilly Taylor and Jennifer Prendki and Marcus Wu and Tom Eccles and Tianqi Liu and Kavya Kopparapu and Francoise Beaufays and Christof Angermueller and Andreea Marzoca and Shourya Sarcar and Hilal Dib and Jeff Stanway and Frank Perbet and Nejc Trdin and Rachel Sterneck and Andrey Khorlin and Dinghua Li and Xihui Wu and Sonam Goenka and David Madras and Sasha Goldshtein and Willi Gierke and Tong Zhou and Yaxin Liu and Yannie Liang and Anais White and Yunjie Li and Shreya Singh and Sanaz Bahargam and Mark Epstein and Sujoy Basu and Li Lao and Adnan Ozturel and Carl Crous and Alex Zhai and Han Lu and Zora Tung and Neeraj Gaur and Alanna Walton and Lucas Dixon and Ming Zhang and Amir Globerson and Grant Uy and Andrew Bolt and Olivia Wiles and Milad Nasr and Ilia Shumailov and Marco Selvi and Francesco Piccinno and Ricardo Aguilar and Sara McCarthy and Misha Khalman and Mrinal Shukla and Vlado Galic and John Carpenter and Kevin Villela and Haibin Zhang and Harry Richardson and James Martens and Matko Bosnjak and Shreyas Rammohan Belle and Jeff Seibert and Mahmoud Alnahlawi and Brian McWilliams and Sankalp Singh and Annie Louis and Wen Ding and Dan Popovici and Lenin Simicich and Laura Knight and Pulkit Mehta and Nishesh Gupta and Chongyang Shi and Saaber Fatehi and Jovana Mitrovic and Alex Grills and Joseph Pagadora and Tsendsuren Munkhdalai and Dessie Petrova and Danielle Eisenbud and Zhishuai Zhang and Damion Yates and Bhavishya Mittal and Nilesh Tripuraneni and Yannis Assael and Thomas Brovelli and Prateek Jain and Mihajlo Velimirovic and Canfer Akbulut and Jiaqi Mu and Wolfgang Macherey and Ravin Kumar and Jun Xu and Haroon Qureshi and Gheorghe Comanici and Jeremy Wiesner and Zhitao Gong and Anton Ruddock and Matthias Bauer and Nick Felt and Anirudh GP and Anurag Arnab and Dustin Zelle and Jonas Rothfuss and Bill Rosgen and Ashish Shenoy and Bryan Seybold and Xinjian Li and Jayaram Mudigonda and Goker Erdogan and Jiawei Xia and Jiri Simsa and Andrea Michi and Yi Yao and Christopher Yew and Steven Kan and Isaac Caswell and Carey Radebaugh and Andre Elisseeff and Pedro Valenzuela and Kay McKinney and Kim Paterson and Albert Cui and Eri Latorre-Chimoto and Solomon Kim and William Zeng and Ken Durden and Priya Ponnapalli and Tiberiu Sosea and Christopher A. Choquette-Choo and James Manyika and Brona Robenek and Harsha Vashisht and Sebastien Pereira and Hoi Lam and Marko Velic and Denese Owusu-Afriyie and Katherine Lee and Tolga Bolukbasi and Alicia Parrish and Shawn Lu and Jane Park and Balaji Venkatraman and Alice Talbert and Lambert Rosique and Yuchung Cheng and Andrei Sozanschi and Adam Paszke and Praveen Kumar and Jessica Austin and Lu Li and Khalid Salama and Bartek Perz and Wooyeol Kim and Nandita Dukkipati and Anthony Baryshnikov and Christos Kaplanis and XiangHai Sheng and Yuri Chervonyi and Caglar Unlu and Diego de Las Casas and Harry Askham and Kathryn Tunyasuvunakool and Felix Gimeno and Siim Poder and Chester Kwak and Matt Miecnikowski and Vahab Mirrokni and Alek Dimitriev and Aaron Parisi and Dangyi Liu and Tomy Tsai and Toby Shevlane and Christina Kouridi and Drew Garmon and Adrian Goedeckemeyer and Adam R. Brown and Anitha Vijayakumar and Ali Elqursh and Sadegh Jazayeri and Jin Huang and Sara Mc Carthy and Jay Hoover and Lucy Kim and Sandeep Kumar and Wei Chen and Courtney Biles and Garrett Bingham and Evan Rosen and Lisa Wang and Qijun Tan and David Engel and Francesco Pongetti and Dario de Cesare and Dongseong Hwang and Lily Yu and Jennifer Pullman and Srini Narayanan and Kyle Levin and Siddharth Gopal and Megan Li and Asaf Aharoni and Trieu Trinh and Jessica Lo and Norman Casagrande and Roopali Vij and Loic Matthey and Bramandia Ramadhana and Austin Matthews and CJ Carey and Matthew Johnson and Kremena Goranova and Rohin Shah and Shereen Ashraf and Kingshuk Dasgupta and Rasmus Larsen and Yicheng Wang and Manish Reddy Vuyyuru and Chong Jiang and Joana Ijazi and Kazuki Osawa and Celine Smith and Ramya Sree Boppana and Taylan Bilal and Yuma Koizumi and Ying Xu and Yasemin Altun and Nir Shabat and Ben Bariach and Alex Korchemniy and Kiam Choo and Olaf Ronneberger and Chimezie Iwuanyanwu and Shubin Zhao and David Soergel and Cho-Jui Hsieh and Irene Cai and Shariq Iqbal and Martin Sundermeyer and Zhe Chen and Elie Bursztein and Chaitanya Malaviya and Fadi Biadsy and Prakash Shroff and Inderjit Dhillon and Tejasi Latkar and Chris Dyer and Hannah Forbes and Massimo Nicosia and Vitaly Nikolaev and Somer Greene and Marin Georgiev and Pidong Wang and Nina Martin and Hanie Sedghi and John Zhang and Praseem Banzal and Doug Fritz and Vikram Rao and Xuezhi Wang and Jiageng Zhang and Viorica Patraucean and Dayou Du and Igor Mordatch and Ivan Jurin and Lewis Liu and Ayush Dubey and Abhi Mohan and Janek Nowakowski and Vlad-Doru Ion and Nan Wei and Reiko Tojo and Maria Abi Raad and Drew A. Hudson and Vaishakh Keshava and Shubham Agrawal and Kevin Ramirez and Zhichun Wu and Hoang Nguyen and Ji Liu and Madhavi Sewak and Bryce Petrini and DongHyun Choi and Ivan Philips and Ziyue Wang and Ioana Bica and Ankush Garg and Jarek Wilkiewicz and Priyanka Agrawal and Xiaowei Li and Danhao Guo and Emily Xue and Naseer Shaik and Andrew Leach and Sadh MNM Khan and Julia Wiesinger and Sammy Jerome and Abhishek Chakladar and Alek Wenjiao Wang and Tina Ornduff and Folake Abu and Alireza Ghaffarkhah and Marcus Wainwright and Mario Cortes and Frederick Liu and Joshua Maynez and Andreas Terzis and Pouya Samangouei and Riham Mansour and Tomasz Kępa and François-Xavier Aubet and Anton Algymr and Dan Banica and Agoston Weisz and Andras Orban and Alexandre Senges and Ewa Andrejczuk and Mark Geller and Niccolo Dal Santo and Valentin Anklin and Majd Al Merey and Martin Baeuml and Trevor Strohman and Junwen Bai and Slav Petrov and Yonghui Wu and Demis Hassabis and Koray Kavukcuoglu and Jeff Dean and Oriol Vinyals},
      year={2024},
      eprint={2403.05530},
      archivePrefix={arXiv},
      primaryClass={cs.CL},
      url={https://arxiv.org/abs/2403.05530}, 
}

@misc{liu2023visualinstructiontuning,
      title={Visual Instruction Tuning}, 
      author={Haotian Liu and Chunyuan Li and Qingyang Wu and Yong Jae Lee},
      year={2023},
      eprint={2304.08485},
      archivePrefix={arXiv},
      primaryClass={cs.CV},
      url={https://arxiv.org/abs/2304.08485}, 
}

@article{qwen,
  title={Qwen-vl: A frontier large vision-language model with versatile abilities},
  author={Bai, Jinze and Bai, Shuai and Yang, Shusheng and Wang, Shijie and Tan, Sinan and Wang, Peng and Lin, Junyang and Zhou, Chang and Zhou, Jingren},
  journal={arXiv preprint arXiv:2308.12966},
  year={2023}
}

@article{chen2023internvl,
  title={InternVL: Scaling up Vision Foundation Models and Aligning for Generic Visual-Linguistic Tasks},
  author={Chen, Zhe and Wu, Jiannan and Wang, Wenhai and Su, Weijie and Chen, Guo and Xing, Sen and Zhong, Muyan and Zhang, Qinglong and Zhu, Xizhou and Lu, Lewei and Li, Bin and Luo, Ping and Lu, Tong and Qiao, Yu and Dai, Jifeng},
  journal={arXiv preprint arXiv:2312.14238},
  year={2023}
}

@misc{qwen25vl,
      title={Qwen2.5-VL Technical Report}, 
      author={Shuai Bai and Keqin Chen and Xuejing Liu and Jialin Wang and Wenbin Ge and Sibo Song and Kai Dang and Peng Wang and Shijie Wang and Jun Tang and Humen Zhong and Yuanzhi Zhu and Mingkun Yang and Zhaohai Li and Jianqiang Wan and Pengfei Wang and Wei Ding and Zheren Fu and Yiheng Xu and Jiabo Ye and Xi Zhang and Tianbao Xie and Zesen Cheng and Hang Zhang and Zhibo Yang and Haiyang Xu and Junyang Lin},
      year={2025},
      eprint={2502.13923},
      archivePrefix={arXiv},
      primaryClass={cs.CV},
      url={https://arxiv.org/abs/2502.13923}, 
}

@misc{dehghani2023patchnpacknavit,
      title={Patch n' Pack: NaViT, a Vision Transformer for any Aspect Ratio and Resolution}, 
      author={Mostafa Dehghani and Basil Mustafa and Josip Djolonga and Jonathan Heek and Matthias Minderer and Mathilde Caron and Andreas Steiner and Joan Puigcerver and Robert Geirhos and Ibrahim Alabdulmohsin and Avital Oliver and Piotr Padlewski and Alexey Gritsenko and Mario Lučić and Neil Houlsby},
      year={2023},
      eprint={2307.06304},
      archivePrefix={arXiv},
      primaryClass={cs.CV},
      url={https://arxiv.org/abs/2307.06304}, 
}

@misc{ye2024mplugowl3longimagesequenceunderstanding,
      title={mPLUG-Owl3: Towards Long Image-Sequence Understanding in Multi-Modal Large Language Models}, 
      author={Jiabo Ye and Haiyang Xu and Haowei Liu and Anwen Hu and Ming Yan and Qi Qian and Ji Zhang and Fei Huang and Jingren Zhou},
      year={2024},
      eprint={2408.04840},
      archivePrefix={arXiv},
      primaryClass={cs.CV},
      url={https://arxiv.org/abs/2408.04840}, 
}

@misc{liu2024llavanext,
    title={LLaVA-NeXT: Improved reasoning, OCR, and world knowledge},
    url={https://llava-vl.github.io/blog/2024-01-30-llava-next/},
    author={Liu, Haotian and Li, Chunyuan and Li, Yuheng and Li, Bo and Zhang, Yuanhan and Shen, Sheng and Lee, Yong Jae},
    month={January},
    year={2024}
}

@misc{li2024llavaonevisioneasyvisualtask,
      title={LLaVA-OneVision: Easy Visual Task Transfer}, 
      author={Bo Li and Yuanhan Zhang and Dong Guo and Renrui Zhang and Feng Li and Hao Zhang and Kaichen Zhang and Peiyuan Zhang and Yanwei Li and Ziwei Liu and Chunyuan Li},
      year={2024},
      eprint={2408.03326},
      archivePrefix={arXiv},
      primaryClass={cs.CV},
      url={https://arxiv.org/abs/2408.03326}, 
}

@article{cha2023honeybee,
  title={Honeybee: Locality-enhanced Projector for Multimodal LLM},
  author={Junbum Cha and Wooyoung Kang and Jonghwan Mun and Byungseok Roh},
  journal={arXiv preprint arXiv:2312.06742},
  year={2023}
}

@misc{shen2024longvuspatiotemporaladaptivecompression,
      title={LongVU: Spatiotemporal Adaptive Compression for Long Video-Language Understanding}, 
      author={Xiaoqian Shen and Yunyang Xiong and Changsheng Zhao and Lemeng Wu and Jun Chen and Chenchen Zhu and Zechun Liu and Fanyi Xiao and Balakrishnan Varadarajan and Florian Bordes and Zhuang Liu and Hu Xu and Hyunwoo J. Kim and Bilge Soran and Raghuraman Krishnamoorthi and Mohamed Elhoseiny and Vikas Chandra},
      year={2024},
      eprint={2410.17434},
      archivePrefix={arXiv},
      primaryClass={cs.CV},
      url={https://arxiv.org/abs/2410.17434}, 
}

@misc{yang2024pvcprogressivevisualtoken,
      title={PVC: Progressive Visual Token Compression for Unified Image and Video Processing in Large Vision-Language Models}, 
      author={Chenyu Yang and Xuan Dong and Xizhou Zhu and Weijie Su and Jiahao Wang and Hao Tian and Zhe Chen and Wenhai Wang and Lewei Lu and Jifeng Dai},
      year={2024},
      eprint={2412.09613},
      archivePrefix={arXiv},
      primaryClass={cs.CV},
      url={https://arxiv.org/abs/2412.09613}, 
}

@article{wei2022chain,
  title={Chain-of-thought prompting elicits reasoning in large language models},
  author={Wei, Jason and Wang, Xuezhi and Schuurmans, Dale and Bosma, Maarten and Xia, Fei and Chi, Ed and Le, Quoc V and Zhou, Denny and others},
  journal={Advances in Neural Information Processing Systems},
  volume={35},
  pages={24824--24837},
  year={2022}
}

@inproceedings{
wang2023selfconsistency,
title={Self-Consistency Improves Chain of Thought Reasoning in Language Models},
author={Xuezhi Wang and Jason Wei and Dale Schuurmans and Quoc V Le and Ed H. Chi and Sharan Narang and Aakanksha Chowdhery and Denny Zhou},
booktitle={The Eleventh International Conference on Learning Representations },
year={2023},
url={https://openreview.net/forum?id=1PL1NIMMrw}
}

@article{mu2023embodiedgpt,
  title={Embodiedgpt: Vision-language pre-training via embodied chain of thought},
  author={Mu, Yao and Zhang, Qinglong and Hu, Mengkang and Wang, Wenhai and Ding, Mingyu and Jin, Jun and Wang, Bin and Dai, Jifeng and Qiao, Yu and Luo, Ping},
  journal={arXiv preprint arXiv:2305.15021},
  year={2023}
}

@article{lu2022learn,
  title={Learn to explain: Multimodal reasoning via thought chains for science question answering},
  author={Lu, Pan and Mishra, Swaroop and Xia, Tanglin and Qiu, Liang and Chang, Kai-Wei and Zhu, Song-Chun and Tafjord, Oyvind and Clark, Peter and Kalyan, Ashwin},
  journal={Advances in Neural Information Processing Systems},
  volume={35},
  pages={2507--2521},
  year={2022}
}

@misc{guo2024mammothvlelicitingmultimodalreasoning,
      title={MAmmoTH-VL: Eliciting Multimodal Reasoning with Instruction Tuning at Scale}, 
      author={Jarvis Guo and Tuney Zheng and Yuelin Bai and Bo Li and Yubo Wang and King Zhu and Yizhi Li and Graham Neubig and Wenhu Chen and Xiang Yue},
      year={2024},
      eprint={2412.05237},
      archivePrefix={arXiv},
      primaryClass={cs.CL},
      url={https://arxiv.org/abs/2412.05237}, 
}

@misc{lewis2021retrievalaugmentedgenerationknowledgeintensivenlp,
      title={Retrieval-Augmented Generation for Knowledge-Intensive NLP Tasks}, 
      author={Patrick Lewis and Ethan Perez and Aleksandra Piktus and Fabio Petroni and Vladimir Karpukhin and Naman Goyal and Heinrich Küttler and Mike Lewis and Wen-tau Yih and Tim Rocktäschel and Sebastian Riedel and Douwe Kiela},
      year={2021},
      eprint={2005.11401},
      archivePrefix={arXiv},
      primaryClass={cs.CL},
      url={https://arxiv.org/abs/2005.11401}, 
}

@misc{cho2024m3docragmultimodalretrievalneed,
      title={M3DocRAG: Multi-modal Retrieval is What You Need for Multi-page Multi-document Understanding}, 
      author={Jaemin Cho and Debanjan Mahata and Ozan Irsoy and Yujie He and Mohit Bansal},
      year={2024},
      eprint={2411.04952},
      archivePrefix={arXiv},
      primaryClass={cs.CV},
      url={https://arxiv.org/abs/2411.04952}, 
}

@misc{yasunaga2023retrievalaugmentedmultimodallanguagemodeling,
      title={Retrieval-Augmented Multimodal Language Modeling}, 
      author={Michihiro Yasunaga and Armen Aghajanyan and Weijia Shi and Rich James and Jure Leskovec and Percy Liang and Mike Lewis and Luke Zettlemoyer and Wen-tau Yih},
      year={2023},
      eprint={2211.12561},
      archivePrefix={arXiv},
      primaryClass={cs.CV},
      url={https://arxiv.org/abs/2211.12561}, 
}

@misc{hu2024mplugdocowl2highresolutioncompressingocrfree,
      title={mPLUG-DocOwl2: High-resolution Compressing for OCR-free Multi-page Document Understanding}, 
      author={Anwen Hu and Haiyang Xu and Liang Zhang and Jiabo Ye and Ming Yan and Ji Zhang and Qin Jin and Fei Huang and Jingren Zhou},
      year={2024},
      eprint={2409.03420},
      archivePrefix={arXiv},
      primaryClass={cs.CV},
      url={https://arxiv.org/abs/2409.03420}, 
}

@misc{yu2025dapoopensourcellmreinforcement,
      title={DAPO: An Open-Source LLM Reinforcement Learning System at Scale}, 
      author={Qiying Yu and Zheng Zhang and Ruofei Zhu and Yufeng Yuan and Xiaochen Zuo and Yu Yue and Tiantian Fan and Gaohong Liu and Lingjun Liu and Xin Liu and Haibin Lin and Zhiqi Lin and Bole Ma and Guangming Sheng and Yuxuan Tong and Chi Zhang and Mofan Zhang and Wang Zhang and Hang Zhu and Jinhua Zhu and Jiaze Chen and Jiangjie Chen and Chengyi Wang and Hongli Yu and Weinan Dai and Yuxuan Song and Xiangpeng Wei and Hao Zhou and Jingjing Liu and Wei-Ying Ma and Ya-Qin Zhang and Lin Yan and Mu Qiao and Yonghui Wu and Mingxuan Wang},
      year={2025},
      eprint={2503.14476},
      archivePrefix={arXiv},
      primaryClass={cs.LG},
      url={https://arxiv.org/abs/2503.14476}, 
}

@misc{deepseekai2025deepseekr1incentivizingreasoningcapability,
      title={DeepSeek-R1: Incentivizing Reasoning Capability in LLMs via Reinforcement Learning}, 
      author={DeepSeek-AI and Daya Guo and Dejian Yang and Haowei Zhang and Junxiao Song and Ruoyu Zhang and Runxin Xu and Qihao Zhu and Shirong Ma and Peiyi Wang and Xiao Bi and Xiaokang Zhang and Xingkai Yu and Yu Wu and Z. F. Wu and Zhibin Gou and Zhihong Shao and Zhuoshu Li and Ziyi Gao and Aixin Liu and Bing Xue and Bingxuan Wang and Bochao Wu and Bei Feng and Chengda Lu and Chenggang Zhao and Chengqi Deng and Chenyu Zhang and Chong Ruan and Damai Dai and Deli Chen and Dongjie Ji and Erhang Li and Fangyun Lin and Fucong Dai and Fuli Luo and Guangbo Hao and Guanting Chen and Guowei Li and H. Zhang and Han Bao and Hanwei Xu and Haocheng Wang and Honghui Ding and Huajian Xin and Huazuo Gao and Hui Qu and Hui Li and Jianzhong Guo and Jiashi Li and Jiawei Wang and Jingchang Chen and Jingyang Yuan and Junjie Qiu and Junlong Li and J. L. Cai and Jiaqi Ni and Jian Liang and Jin Chen and Kai Dong and Kai Hu and Kaige Gao and Kang Guan and Kexin Huang and Kuai Yu and Lean Wang and Lecong Zhang and Liang Zhao and Litong Wang and Liyue Zhang and Lei Xu and Leyi Xia and Mingchuan Zhang and Minghua Zhang and Minghui Tang and Meng Li and Miaojun Wang and Mingming Li and Ning Tian and Panpan Huang and Peng Zhang and Qiancheng Wang and Qinyu Chen and Qiushi Du and Ruiqi Ge and Ruisong Zhang and Ruizhe Pan and Runji Wang and R. J. Chen and R. L. Jin and Ruyi Chen and Shanghao Lu and Shangyan Zhou and Shanhuang Chen and Shengfeng Ye and Shiyu Wang and Shuiping Yu and Shunfeng Zhou and Shuting Pan and S. S. Li and Shuang Zhou and Shaoqing Wu and Shengfeng Ye and Tao Yun and Tian Pei and Tianyu Sun and T. Wang and Wangding Zeng and Wanjia Zhao and Wen Liu and Wenfeng Liang and Wenjun Gao and Wenqin Yu and Wentao Zhang and W. L. Xiao and Wei An and Xiaodong Liu and Xiaohan Wang and Xiaokang Chen and Xiaotao Nie and Xin Cheng and Xin Liu and Xin Xie and Xingchao Liu and Xinyu Yang and Xinyuan Li and Xuecheng Su and Xuheng Lin and X. Q. Li and Xiangyue Jin and Xiaojin Shen and Xiaosha Chen and Xiaowen Sun and Xiaoxiang Wang and Xinnan Song and Xinyi Zhou and Xianzu Wang and Xinxia Shan and Y. K. Li and Y. Q. Wang and Y. X. Wei and Yang Zhang and Yanhong Xu and Yao Li and Yao Zhao and Yaofeng Sun and Yaohui Wang and Yi Yu and Yichao Zhang and Yifan Shi and Yiliang Xiong and Ying He and Yishi Piao and Yisong Wang and Yixuan Tan and Yiyang Ma and Yiyuan Liu and Yongqiang Guo and Yuan Ou and Yuduan Wang and Yue Gong and Yuheng Zou and Yujia He and Yunfan Xiong and Yuxiang Luo and Yuxiang You and Yuxuan Liu and Yuyang Zhou and Y. X. Zhu and Yanhong Xu and Yanping Huang and Yaohui Li and Yi Zheng and Yuchen Zhu and Yunxian Ma and Ying Tang and Yukun Zha and Yuting Yan and Z. Z. Ren and Zehui Ren and Zhangli Sha and Zhe Fu and Zhean Xu and Zhenda Xie and Zhengyan Zhang and Zhewen Hao and Zhicheng Ma and Zhigang Yan and Zhiyu Wu and Zihui Gu and Zijia Zhu and Zijun Liu and Zilin Li and Ziwei Xie and Ziyang Song and Zizheng Pan and Zhen Huang and Zhipeng Xu and Zhongyu Zhang and Zhen Zhang},
      year={2025},
      eprint={2501.12948},
      archivePrefix={arXiv},
      primaryClass={cs.CL},
      url={https://arxiv.org/abs/2501.12948}, 
}

@misc{grpo,
      title={DeepSeekMath: Pushing the Limits of Mathematical Reasoning in Open Language Models}, 
      author={Zhihong Shao and Peiyi Wang and Qihao Zhu and Runxin Xu and Junxiao Song and Xiao Bi and Haowei Zhang and Mingchuan Zhang and Y. K. Li and Y. Wu and Daya Guo},
      year={2024},
      eprint={2402.03300},
      archivePrefix={arXiv},
      primaryClass={cs.CL},
      url={https://arxiv.org/abs/2402.03300}, 
}

@misc{bai2024digirltraininginthewilddevicecontrol,
      title={DigiRL: Training In-The-Wild Device-Control Agents with Autonomous Reinforcement Learning}, 
      author={Hao Bai and Yifei Zhou and Mert Cemri and Jiayi Pan and Alane Suhr and Sergey Levine and Aviral Kumar},
      year={2024},
      eprint={2406.11896},
      archivePrefix={arXiv},
      primaryClass={cs.LG},
      url={https://arxiv.org/abs/2406.11896}, 
}

@misc{zhai2024finetuninglargevisionlanguagemodels,
      title={Fine-Tuning Large Vision-Language Models as Decision-Making Agents via Reinforcement Learning}, 
      author={Yuexiang Zhai and Hao Bai and Zipeng Lin and Jiayi Pan and Shengbang Tong and Yifei Zhou and Alane Suhr and Saining Xie and Yann LeCun and Yi Ma and Sergey Levine},
      year={2024},
      eprint={2405.10292},
      archivePrefix={arXiv},
      primaryClass={cs.AI},
      url={https://arxiv.org/abs/2405.10292}, 
}

@misc{zheng2023judgingllmasajudgemtbenchchatbot,
      title={Judging LLM-as-a-Judge with MT-Bench and Chatbot Arena}, 
      author={Lianmin Zheng and Wei-Lin Chiang and Ying Sheng and Siyuan Zhuang and Zhanghao Wu and Yonghao Zhuang and Zi Lin and Zhuohan Li and Dacheng Li and Eric P. Xing and Hao Zhang and Joseph E. Gonzalez and Ion Stoica},
      year={2023},
      eprint={2306.05685},
      archivePrefix={arXiv},
      primaryClass={cs.CL},
      url={https://arxiv.org/abs/2306.05685}, 
}

@misc{tanaka2023slidevqadatasetdocumentvisual,
      title={SlideVQA: A Dataset for Document Visual Question Answering on Multiple Images}, 
      author={Ryota Tanaka and Kyosuke Nishida and Kosuke Nishida and Taku Hasegawa and Itsumi Saito and Kuniko Saito},
      year={2023},
      eprint={2301.04883},
      archivePrefix={arXiv},
      primaryClass={cs.CL},
      url={https://arxiv.org/abs/2301.04883}, 
}

@misc{vanlandeghem2023documentunderstandingdatasetevaluation,
      title={Document Understanding Dataset and Evaluation (DUDE)}, 
      author={Jordy Van Landeghem and Rubén Tito and Łukasz Borchmann and Michał Pietruszka and Paweł Józiak and Rafał Powalski and Dawid Jurkiewicz and Mickaël Coustaty and Bertrand Ackaert and Ernest Valveny and Matthew Blaschko and Sien Moens and Tomasz Stanisławek},
      year={2023},
      eprint={2305.08455},
      archivePrefix={arXiv},
      primaryClass={cs.CV},
      url={https://arxiv.org/abs/2305.08455}, 
}

@misc{ma2024mmlongbenchdocbenchmarkinglongcontextdocument,
      title={MMLongBench-Doc: Benchmarking Long-context Document Understanding with Visualizations}, 
      author={Yubo Ma and Yuhang Zang and Liangyu Chen and Meiqi Chen and Yizhu Jiao and Xinze Li and Xinyuan Lu and Ziyu Liu and Yan Ma and Xiaoyi Dong and Pan Zhang and Liangming Pan and Yu-Gang Jiang and Jiaqi Wang and Yixin Cao and Aixin Sun},
      year={2024},
      eprint={2407.01523},
      archivePrefix={arXiv},
      primaryClass={cs.CV},
      url={https://arxiv.org/abs/2407.01523}, 
}

@misc{tito2023hierarchicalmultimodaltransformersmultipage,
      title={Hierarchical multimodal transformers for Multi-Page DocVQA}, 
      author={Rubèn Tito and Dimosthenis Karatzas and Ernest Valveny},
      year={2023},
      eprint={2212.05935},
      archivePrefix={arXiv},
      primaryClass={cs.CV},
      url={https://arxiv.org/abs/2212.05935}, 
}

@misc{xie2024osworldbenchmarkingmultimodalagents,
      title={OSWorld: Benchmarking Multimodal Agents for Open-Ended Tasks in Real Computer Environments}, 
      author={Tianbao Xie and Danyang Zhang and Jixuan Chen and Xiaochuan Li and Siheng Zhao and Ruisheng Cao and Toh Jing Hua and Zhoujun Cheng and Dongchan Shin and Fangyu Lei and Yitao Liu and Yiheng Xu and Shuyan Zhou and Silvio Savarese and Caiming Xiong and Victor Zhong and Tao Yu},
      year={2024},
      eprint={2404.07972},
      archivePrefix={arXiv},
      primaryClass={cs.AI},
      url={https://arxiv.org/abs/2404.07972}, 
}

@misc{koh2024visualwebarenaevaluatingmultimodalagents,
      title={VisualWebArena: Evaluating Multimodal Agents on Realistic Visual Web Tasks}, 
      author={Jing Yu Koh and Robert Lo and Lawrence Jang and Vikram Duvvur and Ming Chong Lim and Po-Yu Huang and Graham Neubig and Shuyan Zhou and Ruslan Salakhutdinov and Daniel Fried},
      year={2024},
      eprint={2401.13649},
      archivePrefix={arXiv},
      primaryClass={cs.LG},
      url={https://arxiv.org/abs/2401.13649}, 
}

@misc{sun2025osgenesisautomatingguiagent,
      title={OS-Genesis: Automating GUI Agent Trajectory Construction via Reverse Task Synthesis}, 
      author={Qiushi Sun and Kanzhi Cheng and Zichen Ding and Chuanyang Jin and Yian Wang and Fangzhi Xu and Zhenyu Wu and Chengyou Jia and Liheng Chen and Zhoumianze Liu and Ben Kao and Guohao Li and Junxian He and Yu Qiao and Zhiyong Wu},
      year={2025},
      eprint={2412.19723},
      archivePrefix={arXiv},
      primaryClass={cs.AI},
      url={https://arxiv.org/abs/2412.19723}, 
}

@misc{hu2021loralowrankadaptationlarge,
      title={LoRA: Low-Rank Adaptation of Large Language Models}, 
      author={Edward J. Hu and Yelong Shen and Phillip Wallis and Zeyuan Allen-Zhu and Yuanzhi Li and Shean Wang and Lu Wang and Weizhu Chen},
      year={2021},
      eprint={2106.09685},
      archivePrefix={arXiv},
      primaryClass={cs.CL},
      url={https://arxiv.org/abs/2106.09685}, 
}

@misc{biten2019scenetextvisualquestion,
      title={Scene Text Visual Question Answering}, 
      author={Ali Furkan Biten and Ruben Tito and Andres Mafla and Lluis Gomez and Marçal Rusiñol and Ernest Valveny and C. V. Jawahar and Dimosthenis Karatzas},
      year={2019},
      eprint={1905.13648},
      archivePrefix={arXiv},
      primaryClass={cs.CV},
      url={https://arxiv.org/abs/1905.13648}, 
}

@misc{wang2025enhancingreasoningabilitymultimodal,
      title={Enhancing the Reasoning Ability of Multimodal Large Language Models via Mixed Preference Optimization}, 
      author={Weiyun Wang and Zhe Chen and Wenhai Wang and Yue Cao and Yangzhou Liu and Zhangwei Gao and Jinguo Zhu and Xizhou Zhu and Lewei Lu and Yu Qiao and Jifeng Dai},
      year={2025},
      eprint={2411.10442},
      archivePrefix={arXiv},
      primaryClass={cs.CL},
      url={https://arxiv.org/abs/2411.10442}, 
}

@misc{zhang2024lmmsevalrealitycheckevaluation,
      title={LMMs-Eval: Reality Check on the Evaluation of Large Multimodal Models}, 
      author={Kaichen Zhang and Bo Li and Peiyuan Zhang and Fanyi Pu and Joshua Adrian Cahyono and Kairui Hu and Shuai Liu and Yuanhan Zhang and Jingkang Yang and Chunyuan Li and Ziwei Liu},
      year={2024},
      eprint={2407.12772},
      archivePrefix={arXiv},
      primaryClass={cs.CL},
      url={https://arxiv.org/abs/2407.12772}, 
}

@misc{mathew2021docvqadatasetvqadocument,
      title={DocVQA: A Dataset for VQA on Document Images}, 
      author={Minesh Mathew and Dimosthenis Karatzas and C. V. Jawahar},
      year={2021},
      eprint={2007.00398},
      archivePrefix={arXiv},
      primaryClass={cs.CV},
      url={https://arxiv.org/abs/2007.00398}, 
}

@article{meng2025mm,
  title={MM-Eureka: Exploring Visual Aha Moment with Rule-based Large-scale Reinforcement Learning},
  author={Meng, Fanqing and Du, Lingxiao and Liu, Zongkai and Zhou, Zhixiang and Lu, Quanfeng and Fu, Daocheng and Shi, Botian and Wang, Wenhai and He, Junjun and Zhang, Kaipeng and others},
  journal={arXiv preprint arXiv:2503.07365},
  year={2025}
}

@misc{zhang2024android,
      title={Android in the Zoo: Chain-of-Action-Thought for GUI Agents}, 
      author={Jiwen Zhang and Jihao Wu and Yihua Teng and Minghui Liao and Nuo Xu and Xiao Xiao and Zhongyu Wei and Duyu Tang},
      year={2024},
      eprint={2403.02713},
      archivePrefix={arXiv},
      primaryClass={cs.CL}
}

\appendix
\onecolumn 
\clearpage
\section{Prompt for Chain of Scroll}\label{sec:cos_prompt}

\begin{figure*}[!h]
  \centering
  \resizebox{1.0\textwidth}{!}{%
    \begin{minipage}{\textwidth}
      \footnotesize
      \begin{tcolorbox}[
        colback=white,
        colframe=black,
        boxsep=3pt,
        arc=0pt,
        outer arc=0pt
      ]
        \textbf{Chain of Scroll Prompt}\\[3pt]
        At each step, you will receive this prompt repeatedly, enabling you to scroll through the document page by page to gather information and ultimately answer the question. Imagine yourself as a human analyzing the document, making observations, and reasoning about whether to continue scrolling or answer the question. The first page of the document is page number 0.\\[4pt]
        \textbf{Question:} \textcolor{blue}{[Question]}\\
        \textbf{Previous\_Note:} \textcolor{blue}{[Previous\_Note]}\\
        \textbf{Current\_page\_num:} \textcolor{blue}{[Current\_page\_num]}\\
        \textbf{Total\_page\_num:} \textcolor{blue}{[Total\_page\_num]}\\[4pt]
        Using the given information above, you can choose to scroll the document to explore other pages or answer the question. If you choose to scroll, return your thoughts, notes to pass question‐relevant information to the next step, and scroll values to scroll forward or backward. Return the thinking process in \textbf{<think>…</think>}, the notes in \textbf{<note>…</note>}, and the scroll value \textbf{(+n or -n)} in \textbf{<scroll>…</scroll>} tags. If you choose to answer, return your thoughts and final answer to the given question. Return the thinking process in \textbf{<think>…</think>} and the answer in \textbf{<answer>…</answer>} tags.
      \end{tcolorbox}
    \end{minipage}%
  }
  \caption{Input prompt of CoS framework. At every step, the \textcolor{blue}{blue} part is replaced with the input query, accumulated previous notes, current page number, and total page number until the model chooses to return the answer.}
  \label{fig:cos-prompt}
\end{figure*}

\begin{figure}[!h]
  \centering
  \resizebox{1.0\textwidth}{!}{%
    \begin{minipage}{\textwidth}
      \footnotesize
      \begin{tcolorbox}[
        colback=white,
        colframe=black,
        boxsep=3pt,
        arc=0pt,
        outer arc=0pt
      ]
        \textbf{Chain of Scroll Prompt}\\[3pt]
        At each step, you will receive this prompt repeatedly, enabling you to scroll through the document page by page to gather information and ultimately answer the question. Imagine yourself as a human analyzing the document, making observations, and reasoning about whether to continue scrolling or answer to the question. The first page of the document is page number 0.\\[4pt]
        \textbf{Question:} \textcolor{blue}{Which Cluster ranks highest in Importance?}\\
        \textbf{Previous\_Note:} \textcolor{blue}{Page 0: The title of the presentation is Structuring Mobile and Contextual Learning. The presentation is by Dr. Christian Glahn and Prof. Dr. Marcus Specht. Page 17: Page 17 describes Cluster 2: Contextual Learning and lists its characteristics. Page 6: Page 6 discusses Group Concept Mapping which is a method with Qualitative and Quantitative approaches. The quantitative approach includes sorting \& rating importance and feasibility, as well as hierarchical cluster analysis. Page 13: Page 13 presents a concept map illustrating core challenges, including Contextual Learning, Transitions between Contexts, Access to Learning, E-Inclusion, and Orchestrating Learning across Contexts.}\\
        \textbf{Current\_page\_num:} \textcolor{blue}{19}\\
        \textbf{Total\_page\_num:} \textcolor{blue}{20}\\[4pt]
        Using the given information above, you can choose to scroll the document to explore other pages or answer the question. If you choose to scroll, return your thoughts, notes to pass question-relevant information to the next step, and scroll values to scroll forward or backward. Return the thinking process in \textbf{<think>…</think>}, the notes in \textbf{<note>…</note>}, and the scroll value \textbf{(+n or -n)} in \textbf{<scroll>…</scroll>} tags. If you choose to answer, return your thoughts and final answer to the given question. Return the thinking process in \textbf{<think>…</think>} and the answer in \textbf{<answer>…</answer>} tags.
      \end{tcolorbox}
    \end{minipage}%
  }
  \caption{Example shows the input prompt of CoS framework for slidevqa in SCoPE dataset with the dataset id of "00mlearn2011glahn-111020092730-phpapp02\_95\_28\_4".}
  \label{fig:cos-prompt-slidevqa}
\end{figure}

Figure~\ref{fig:cos-prompt} and ~\ref{fig:cos-prompt-slidevqa} illustrates the input prompt template used in the Chain of Scroll (CoS) framework. At each decision step, the model is presented with a structured prompt that includes the user’s question, the current page number being viewed, the total number of pages in the document, and any previously accumulated notes. The model is asked to simulate human-like reasoning by either scrolling through additional pages or answering the question based on the information seen so far. It must explicitly output its reasoning process using `<think>...</think>`, relevant information to carry forward in `<note>...</note>`, and a scroll decision using `<scroll>...</scroll>`. If it opts to answer the question, it returns its final reasoning and conclusion using `<think>...</think>` and `<answer>...</answer>` tags. This recursive prompting mechanism allows the model to incrementally explore long documents and build up a chain of contextual understanding before answering. Since this prompt contains all query-related information from the conversation history, the CoS framework treats each step as a single turn. Previous conversation history is not provided at each step. This design allows CoS to retain global information while navigating the document, enabling the incorporation of holistic information in the final answer.

\clearpage

\section{SCoPE dataset generation}\label{sec:scope_dataset}

\subsection{Overall Annotation Process}

The annotation generation pipeline for SCoPE dataset creates synthetic human-like document navigation traces for multi-page document question answering. The system employs a four-stage approach: (1) first identifying evidence pages containing answer-relevant information when ground truth is unavailable, (2) randomly selecting intermediate pages to create realistic exploration trajectories, (3) annotating scroll steps with human-like reasoning for each page visited, and (4) generating the final answer step with reasoning. The pipeline leverages large language models to simulate human cognitive processes during document exploration, producing training data that captures both successful information-seeking strategies and natural exploration behaviors. The resulting annotations include page-level observations, accumulated notes across pages, and human-like reasoning chains that justify navigation decisions and answer derivation.

\subsection{Prompt for Evidence Page Identification}

\begin{figure*}[h!]
  \centering
  \resizebox{1.0\textwidth}{!}{%
    \begin{minipage}{\textwidth}
      \footnotesize
      \begin{tcolorbox}[
        colback=white,
        colframe=black,
        boxsep=3pt,
        arc=0pt,
        outer arc=0pt
      ]
        \textbf{Evidence Page Identification Prompt}\\[3pt]
        You are given multiple pages of a document, along with a question and an answer that were provided for a query about the document.\\[4pt]
        \textbf{Question:} \textcolor{blue}{[Question]}\\
        \textbf{Answer:} \textcolor{blue}{[Answer]}\\[4pt]
        Your task is to finding all pages of the document are necessary to answer the query accurately. You need to return the image number which starts from 0.
        Return your findings as a list of page numbers in the following format.
        If all of the images seem to be necessary, put all page indices in the list.\\[4pt]
        However, if there is no required pages in the given images return an empty list.\\
        Each image is a page, even if the image is a crop of a bigger image.\\
        Make sure to look back your choice and there must be the \textbf{given answer} in the selected pages.\\[4pt]
        You can only speak json and put all of your thoughts under the thoughts.\\[4pt]
        Use this JSON schema:\\
        Return: \{'thoughts': str, 'output': int\}
      \end{tcolorbox}
    \end{minipage}%
  }
  \caption{Target page identification prompt used to determine which pages contain information necessary to answer a given question. This prompt is only deployed when ground truth answer pages are not available in the dataset.}
  \label{fig:target-prompt}
\end{figure*}

Figure~\ref{fig:target-prompt} presents the prompt template for identifying evidence pages within a document. This prompt is employed only when the dataset lacks pre-annotated answer page information. Given a question-answer pair and multiple document pages, the model must analyze each page to determine which ones contain essential information for answering the question. The model outputs its reasoning process and a list of relevant page indices in JSON format, ensuring that the identified pages collectively contain the information needed to derive the given answer.
\clearpage

\subsection{Prompt for Scroll Step Generation}

\begin{figure*}[h!]
  \centering
  \resizebox{1.0\textwidth}{!}{%
    \begin{minipage}{\textwidth}
      \footnotesize
      \begin{tcolorbox}[
        colback=white,
        colframe=black,
        boxsep=3pt,
        arc=0pt,
        outer arc=0pt
      ]
        \textbf{Scroll Step Generation Prompt}\\[3pt]
        You are given:
        \begin{itemize}
          \item A single-page image of a document
          \item A question
          \item The number of pages to be skipped and scrolled in the next step
          \item Current page number
          \item The total number of pages in the document
          \item Notes containing relevant information from other pages
        \end{itemize}
        At each step, you receive this prompt repeatedly, enabling you to scroll through the document page by page to gather information and ultimately answer the question. Imagine yourself as a human analyzing the document, making observations, and reasoning about whether to continue scrolling. Your task is to generate realistic, human-like reasoning for decision-making. Think as if you have the choice to either answer or continue exploring based on your notes and findings, while also determining the appropriate scroll value—though you are not allowed to answer at this step. The first page of the document is page number 0.\\[4pt]
        \textbf{Question:} \textcolor{blue}{[Question]}\\
        \textbf{Previous Note:} \textcolor{blue}{[Previous\_note]}\\
        \textbf{Scroll\_value:} \textcolor{blue}{[Scroll\_value]}\\
        \textbf{Current\_page\_num:} \textcolor{blue}{[Current\_page\_num]}\\
        \textbf{Total\_page\_num:} \textcolor{blue}{[Total\_page\_num]}\\[4pt]
        Your job is to:
        \begin{enumerate}
          \item Identify any information on the \textbf{current page} that can be useful to answer the question. Do not repeat the previous note and its information. Only return the new information. Also, note a brief summary of the current page. Do not format the note at all. Put a string simply.
          \item Write out an in-depth thinking process about how you find this relevant information and reasoning to conclude to scroll by the scroll value (The thoughts should not reveal that it is instructed and the scroll value is provided.. Answer as if you are not given the scroll value. You still need to provide profound reasoning that you need to scroll [Scroll\_value]).
        \end{enumerate}
        Return your response in \textbf{JSON format}.
      \end{tcolorbox}
    \end{minipage}%
  }
  \caption{Scroll step generation prompt for creating intermediate navigation annotations. This prompt generates human-like reasoning for each page visited during document exploration.}
  \label{fig:scroll-prompt}
\end{figure*}

Figure~\ref{fig:scroll-prompt} shows the prompt template for generating intermediate navigation steps. For each page in the exploration trajectory except the final page, this prompt elicits reasoning about the current page's content and justification for the navigation decision. The model must identify new relevant information on the current page, maintain accumulated notes from previous pages, and generate plausible reasoning for why a human would choose to scroll to the next page. Crucially, the model must generate reasoning as if it independently decided to navigate, even though the scroll direction is predetermined, ensuring the annotations reflect natural exploration patterns.
\clearpage

\subsection{Prompt for Answer Step Generation}

\begin{figure*}[!h]
  \centering
  \resizebox{1.0\textwidth}{!}{%
    \begin{minipage}{\textwidth}
      \footnotesize
      \begin{tcolorbox}[
        colback=white,
        colframe=black,
        boxsep=3pt,
        arc=0pt,
        outer arc=0pt
      ]
        \textbf{Answer Step Generation Prompt}\\[3pt]
        You are given:
        \begin{itemize}
          \item A single-page image of a document
          \item A question
          \item Current page number
          \item The total number of pages in the document
          \item Notes containing relevant information from other pages
        \end{itemize}
        At each step, you receive this prompt repeatedly, enabling you to scroll through the document page by page to gather information and ultimately answer the question. Imagine yourself as a human analyzing the document, making observations, and reasoning about whether to continue scrolling. Your task is to generate realistic, human-like reasoning for decision-making. Think as if you have the choice to either answer or continue exploring based on your notes and findings without the answer given, though you must provide the final answer to the question in the end. The first page of the document is page number 0.\\
        Any pages including the first and last page may have the enough information to answer the question.\\[4pt]
        \textbf{Question:} \textcolor{blue}{[Question]}\\
        \textbf{Answer:} \textcolor{blue}{[Answer]}\\
        \textbf{Previous Note:} \textcolor{blue}{[Previous\_Note]}\\
        \textbf{Current\_page\_num:} \textcolor{blue}{[Current\_page\_num]}\\[4pt]
        Try to examine each step in depth and as if it is a realistic thinking process.\\

        Your task is to:
        \begin{enumerate}
          \item Identify any information on the \textbf{current page} that can be useful to answer the question.
          \item Write out an in-depth thinking process step by step to identify the answer to the question on the current page. There must be the answer in the current page.
          \item Explain why you now have enough information to provide the answer
          \item You MUST DERIVE THE FINAL ANSWER based on the previous note and the current page. \textit{You cannot say that the answer is given.}
        \end{enumerate}
        Return your response in the exact following \textbf{JSON format}.
      \end{tcolorbox}
    \end{minipage}%
  }
  \caption{Answer step generation prompt for creating answer derivation annotations at the terminal page of the exploration trajectory.}
  \label{fig:answer-prompt}
\end{figure*}

Figure~\ref{fig:answer-prompt} illustrates the prompt template for generating the answer step. Applied to the last page in the exploration trajectory, this prompt requires the model to synthesize information from all previously visited pages from accumulated notes with content from the current page to derive the answer. The model must demonstrate explicit reasoning about why sufficient information has been gathered, how the current page contributes to the answer, and provide a step-by-step derivation process. The prompt ensures that the model generates reasoning as if discovering the answer organically, rather than simply restating a provided answer, creating annotations that reflect genuine comprehension and synthesis processes.
\clearpage

\section{Formal representation of Chain of Scroll}\label{sec:cos_formal}
Let $\tau=(s_0,a_0,\dots,s_T,a_T)$ be a trajectory generated by policy $\pi_\theta$. In the Chain of Scroll framework, each state $s_t$ depends on the input query $q$, all input images $\mathit{imgs}$, current page index, and accumulated context (notes).\\

\noindent Let $\mathcal{V}$ be the vocabulary of tokens, and $\mathcal{V}^*$ denote the set of all finite sequences over $\mathcal{V}$.

\subsection{State Space}
We formalize the state at time $t$ as $s_t = (\mathit{page}_t, \mathit{notes}_t, \mathit{visited}_t)$ where:
\begin{itemize}
\item $\mathit{page}_t \in \{0, 1, \ldots, N\}$ is the current page index (0-based)
\item $\mathit{notes}_t \subseteq \mathcal{V}^*$ is the set of accumulated notes containing extracted information
\item $\mathit{visited}_t \in \{0, 1\}^{N+1}$ is a boolean array tracking which pages have been visited
\end{itemize}

\subsection{Context and Transition Function}
At each step, the context is assembled from $s_t$ and $a_t$ as:
\begin{equation}
\mathit{c}_t = (q, \mathit{imgs}, \mathit{page}_t, \mathit{scroll}_t, \mathit{notes}_t, \mathit{visited}_t)
\end{equation}
where $\mathit{imgs} = \{\mathit{Image}_0, \dots, \mathit{Image}_N\}$.\\

\noindent The transition function \textsc{Tran\_fn} processes this context:
\begin{equation}
\textsc{Tran\_fn}(\mathit{c}_t) \rightarrow (\mathit{prompt}_t, \mathit{cur\_img}_t, \mathit{page}_t, \mathit{visited}_t)
\end{equation}
where:
\begin{itemize}
\item $\mathit{prompt}_t \in \mathcal{V}^*$ is the generated prompt containing task instructions and accumulated context
\item $\mathit{cur\_img}_t = \mathit{imgs}[\mathit{page}_t]$ is the current page image
\end{itemize}

\subsection{Policy}
The policy $\pi_\theta$ takes the image and prompt to generate a response:
\begin{equation}
\pi_\theta(\mathit{cur\_img}_t, \mathit{prompt}_t) \rightarrow \mathit{response}_t \in \mathcal{V}^*
\end{equation}

\noindent The policy never directly receives the user query $q$, full image set $\textit{imgs}$, or raw state components. These are processed by $\textsc{Tran\_fn}$ into the prompt and \textit{prompt}$_t$.

\subsection{Parse Function}
The $\textsc{Parse}$ function extracts structured output from the model's response:

\begin{equation}
\textsc{Parse}(\mathit{response}_t) \rightarrow (\mathit{cur\_note}_t, \mathit{scroll}_t, \mathit{answer}_t)
\end{equation}
where $\mathit{answer}_t \in \mathcal{V}^*$.

\subsection{Action Space}
The action space is implicitly defined by the parse output:
\begin{itemize}
\item Scroll action: $(\mathit{cur\_note}_t, \mathit{scroll}_t, \varnothing)$ where $\mathit{scroll}_t$ must satisfy:
\begin{equation}
-\mathit{page}_t \leq \mathit{scroll}_t \leq N - \mathit{page}_t
\end{equation}
\item Answer action: $(\mathit{cur\_note}_t, \mathit{scroll}_t, \mathit{answer}_t)$ where $\mathit{answer}_t \neq \varnothing$
\end{itemize}

\subsection{State Updates and Termination}
\subsubsection{Initial State}
The initial state is defined as:
\begin{equation}
s_0 = (\mathit{page}_0 = 0, \mathit{notes}_0 = \varnothing, \mathit{visited}_0 = [0]^{N+1})
\end{equation}
with $\mathit{scroll}_0 = 0$.

\subsubsection{Notes Accumulation}
\begin{equation}
\mathit{notes}_{t+1} = \mathit{notes}_t \cup \{\mathit{cur\_note}_t\}
\end{equation}

\subsubsection{Page Update}
The page index is updated based on the scroll value:
\begin{equation}
\mathit{page}_{t+1} = \min(\max(0, \mathit{page}_t + \mathit{scroll}_t), N)
\end{equation}

\subsubsection{Visit History Update}
\begin{equation}
\mathit{visited}_{t+1}[i] = \begin{cases}
1 & \text{if } i = \mathit{page}_{t+1} \\
\mathit{visited}_t[i] & \text{otherwise}
\end{cases}
\end{equation}

\subsubsection{Episode Termination and Final Output}

An episode terminates when one of the following conditions is met:
\begin{itemize}
    \item An answer is found: $\mathit{answer}_t \neq \varnothing$
    \item Maximum steps reached: $t = T_{\max}$, where $T_{\max} = \min(\mathit{maxSteps}, |\mathit{imgs}|)$
\end{itemize}

\noindent The algorithm returns:
\begin{equation}
\mathit{answer} = \begin{cases}
    \mathit{answer}_t & \text{if } \mathit{answer}_t \neq \varnothing \text{ for some } t \leq T_{\max} \\
    \varnothing & \text{otherwise}
\end{cases}
\end{equation}

\noindent where the first non-empty answer encountered during the episode is returned.

\clearpage
\section{Episodic Group Relative Policy Optimization (EGRPO)\protect\footnotemark}
\label{egrpo_derivation}
\footnotetext{The training code is based on MM-Eureka~\cite{meng2025mm}.}

\begin{figure*}[!ht]
\begin{center}
\resizebox{1.0\textwidth}{!}{%
\includegraphics[width=\textwidth]{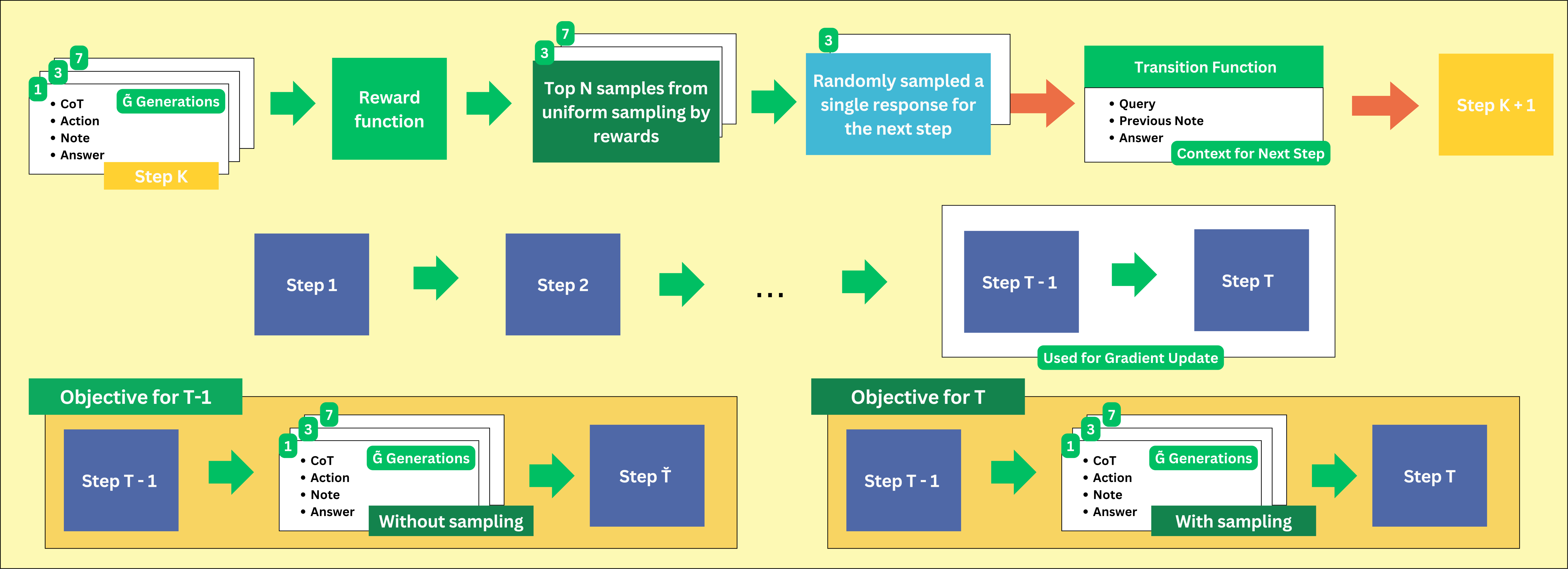}
}
\caption{Overview of Episodic Group Relative Policy Optimization (EGRPO). The algorithm generates $\tilde{G}$ candidate actions at each step, applies uniform sampling followed by random selection from top-$N$ to balance exploration-exploitation.}
\label{fig:egrpo_apdx}
\end{center}
\end{figure*}

While GRPO~\cite{deepseekai2025deepseekr1incentivizingreasoningcapability} has proven effective for single-step Chain-of-Thought reasoning, it does not account for multi-step frameworks like Chain of Scroll (CoS) where agents must navigate documents, accumulate information, and generate answers. Two critical problems emerge: (1) computing fine-grained rewards and values for every step is computationally expensive, especially with LLM-as-judge methods~\cite{zheng2023judgingllmasajudgemtbenchchatbot}, and (2) group-based value computation becomes intractable due to exponential growth in candidate trajectories. EGRPO addresses these by focusing gradient updates on penultimate and terminal steps, using projected states $\check{T}$ to estimate returns without exhaustive generation.

EGRPO bridges this gap by introducing a two-step objective that back-propagates rewards to the penultimate scroll step, exploiting that this state contains sufficient information to reach answer pages, coupling Random Sampling from Top-$N$ with Uniform Sampling for balanced exploration-exploitation, and removing KL penalties while using $\pi_{\theta_\text{ref}}$ and LoRA regularization. As illustrated in Figure~\ref{fig:egrpo_apdx}, the algorithm generates $\tilde{G}$ candidates per step, applies uniform sampling of $G$ candidates followed by random selection from top-$N$ (by rewards), ensuring efficient exploration. EGRPO employs a two-step objective that focuses gradient updates on the penultimate step ($T-1$) with future return estimation through projected terminal step $\check{T}$---which is generated without sampling strategies from $T-1$ to estimate the expected reward if the agent continues from that state---and the terminal step ($T$) with direct reward signals, where $\check{T}$ serves as a projection of what the terminal state would be if the agent took one more step from $T-1$ without exploration noise.

\subsection{Overview of EGRPO objectives}

The EGRPO objective function combines terminal and penultimate timestep objectives:
\begin{align}
\boxed{\mathcal{J}_{\text{EGRPO}}(\theta) = \gamma\,\mathcal{J}^{(T)}(\theta) + \mathcal{J}^{(T-1)}(\theta)} \label{eq:Jegrpo}
\end{align}

\noindent where $\theta$ represents the policy parameters, $\gamma$ is a weighting factor that balances the importance of the terminal step objective, $T$ denotes the terminal (final) timestep, and $T-1$ is the penultimate (second-to-last, the final scroll step) timestep.\\

\subsection{Group Sampling: Uniform Sampling and Random Sampling from Top-$N$ candidates}
\label{app:egrpo:groupsampling}

Following the inference phase illustrated in Figure~\ref{fig:egrpo_main} and ~\ref{fig:egrpo_apdx}, the model initially generates $\tilde{G}$ candidates and samples to ${G}$ candidates, evaluates them with the composite reward, and then proceeds in two stages:

\begin{enumerate}
\item \textbf{Uniform Sampling.}  $G$ candidates are drawn uniformly from $\tilde{G}$ to form a sorted mini-batch by rewards, preserving a wide range of diverse reward signals from generated samples.
\item \textbf{Random Sampling from Top-$N$ candidates.}  Among the $G$ candidates, we identify the $N$ highest-reward samples and randomly select the next environment action from this subset.  
\end{enumerate}

\noindent While a high initial number of generations is essential to increase exploration and discover diverse high-reward trajectories, this approach incurs substantial computational costs and often produces repetitive samples with limited unique solutions. Uniform sampling alone optimizes exploration by decreasing the size of the group while preserving the diversity; top-N selection maximizes exploitation but risks premature convergence to local optima. Our two-stage scheme inherits the best of both approaches while addressing the uniqueness problem: Top-$N$ selection guarantees that each sampled action is within high-reward trajectories and filters out repetitive low-quality samples, while random selection encourages exploration.

\subsection{Regularization Strategy}
\label{app:egrpo:reg}

To further optimize the computational cost to train present long horizon of the CoS, we optimize the regularization strategy used in GRPO. To address these challenges, we adopt a streamlined approach inspired by DAPO \cite{yu2025dapoopensourcellmreinforcement}. Specifically, we remove the explicit KL divergence penalty term from the objective function. Additionally, to promote exploration and prevent the policy from prematurely converging to the initial behavior, we modify the importance ratio $\rho_{i,k}^{(t)}(\theta)$ by replacing $\pi_{\theta_\text{old}}$ with the fixed reference policy $\pi_{\theta_\text{ref}}$.
This design choice is motivated by our use of LoRA \cite{hu2021loralowrankadaptationlarge} adaptation, which provides implicit regularization through its low-rank parameterization. By constraining updates to a low-dimensional subspace, LoRA naturally prevents excessive policy drift while significantly reducing memory requirements and computational burden compared to full model fine-tuning. This approach achieves effective regularization without the computational overhead of maintaining and updating a separate old policy network.

\subsection{Formal Derivation of the EGRPO Objective}
\label{app:egrpo:derivation}
\subsubsection{Underlying Assumptions in EGRPO}

\noindent In order to define the information content function $I: \mathcal{S} \rightarrow \mathcal{I}$ that maps states to their information content relevant for answer generation, we make the following key assumptions:\\

\noindent \textbf{Assumption 1: Representativeness of Penultimate State} The penultimate scroll step (Step T-1) is a representative state that contains or can access the critical information needed to reach the final page containing the answer. This assumption is valid for both single-hop retrieval and multi-hop reasoning where the last pieces of information needed to answer the user's question are accessible from the penultimate state:
\begin{equation}
P(\text{correct answer} | s_{T-1}) \approx P(\text{reaching answer page} | s_{T-1}) \cdot P(\text{correct answer} | \text{answer page reached})
\end{equation}
\noindent \textbf{Assumption 2: Context-Dependent Valid States.} For single-hop reasoning, there exist multiple independent valid penultimate states $s_{T-1} \in \mathcal{S}_{T-1}^*$ and answer states $s_T \in \mathcal{S}_T^*$. For multi-hop reasoning, the set of valid states may be constrained by information dependencies, where $\mathcal{S}_{T-1}^*(n_t) \subseteq \mathcal{S}_{T-1}^*$ depends on the accumulated information in the note buffer $n_t$. Formally:
\begin{equation}
\mathcal{S}_{T-1}^*(n_t) = \{s_{T-1} \in \mathcal{S} : I(n_t) \supseteq I_{\text{prereq}}\}
\end{equation}
where $I_{\text{prereq}}$ represents prerequisite information needed for the final reasoning step.\\

\noindent \textbf{Assumption 3: Dynamic Programming with Diverse Training.} By optimizing transitions from diverse penultimate to answer states across varied trajectory lengths during training, the model implicitly learns to increase probabilities of productive actions throughout entire trajectories—specifically navigation actions toward relevant pages, note-taking actions for critical information extraction, and appropriately-timed answer generation. This occurs through two mechanisms:
\begin{enumerate}
\item Value propagates backward through the trajectory via the Bellman equation:
\begin{equation}
V^{\pi}(s_t) = \mathbb{E}_{a_t \sim \pi} \left[ r_t + \gamma V^{\pi}(s_{t+1}) \right]
\end{equation}
Improving $V^{\pi}(s_{T-1})$ creates pressure to improve $V^{\pi}(s_{t})$ for all $t < T-1$, which increases $\pi(a|s_t)$ for actions $a$ that lead to high-value successor states.
\item During training, the same state can appear at different positions across trajectories of varying lengths. A state that serves as penultimate ($s_{T-1}$) in a short trajectory may appear much earlier ($s_t$ for $t \ll T$) in a longer trajectory. This teaches the model to recognize and execute critical transitions at any position.
\end{enumerate}

\subsubsection{Validity of Assumptions in Multi-Hop Reasoning}

The three assumptions naturally extend to multi-hop reasoning where agents must integrate information from multiple sources:\\

\noindent \textbf{Assumption 1:} The penultimate state $s_{T-1}$ in multi-hop reasoning contains all prerequisite information in the note buffer $n_{T-1}$, with only the final navigation remaining. The representative state property holds with the extended condition.

\noindent \textbf{Assumption 2:} Information dependencies constrain valid states based on accumulated facts. For a query requiring facts $\{f_1, f_2, ..., f_k\}$, valid penultimate states must satisfy:
\begin{equation}
\mathcal{S}_{T-1}^* = \{s : \{f_1, ..., f_k\} \subseteq n_t(s) \land \text{can\_reach\_answer}(s)\}
\end{equation}

\noindent \textbf{Assumption 3:} Multi-hop reasoning creates naturally diverse trajectory lengths. A state containing facts $\{f_1, f_2\}$ serves as penultimate in a 2-hop query but appears early in a 3-hop query requiring $\{f_1, f_2, f_3\}$. This diversity ensures the model learns to execute productive transitions from any position. Thus, learning to handle final-hop transitions improves navigation and note-taking decisions throughout the entire multi-hop chain.

\subsubsection{Terminal Step Projection}\label{sec:term_proj}
The traditional advantage function is:
\begin{equation}
A^\pi(s,a) = Q^\pi(s,a) - V^\pi(s)
\end{equation}
\noindent For the penultimate step in a trajectory, this becomes:
\begin{equation}
A^\pi(s_{T-1}, a_{T-1}) = Q^\pi(s_{T-1}, a_{T-1}) - V^\pi(s_{T-1})
\end{equation}
\noindent In EGRPO, we make the following estimations:
\noindent $Q^\pi(s_{T-1}, a_{T-1})$ is the expected return from taking action $a_{T-1}$ at state $s_{T-1}$:
\begin{equation}
Q^\pi(s_{T-1}, a_{T-1}) = \mathbb{E}_{\pi}\left[\sum_{t=T-1}^{T} r^{(t)} \mid s_{T-1}, a_{T-1}\right]
\end{equation}
For trajectory $i$, we estimate this as:
\begin{equation}
Q^\pi(s_{T-1}, a_{T-1}) \approx \hat{r}_i^{(T-1)} = \underbrace{r_i^{(T-1)}}_{\text{current reward}} + \underbrace{r_i^{(\check{T})}}_{\text{future return estimate}}
\end{equation}
where:
\begin{itemize}
    \item $r_i^{(T-1)}$ is the immediate reward at step $T-1$
    \item $r_i^{(\check{T})}$ is the estimated future return obtained through terminal step projection
\end{itemize}
\noindent The estimation $r_i^{(\check{T})}$ is critical for converting GRPO to the episodic setting:
\begin{enumerate}
    \item Unlike traditional RL that requires complete trajectories, EGRPO can compute returns and values using only the current group of samples by leveraging the group-based advantage estimation from GRPO. By projecting incomplete trajectories to terminal states, we obtain return estimates without waiting for full episode completion.
    
    \item Incomplete trajectories receive lower return estimates compared to successfully completed ones.
    This creates a natural reward gradient where:
    \begin{itemize}
        \item Successfully completed trajectories receive full terminal rewards $r_i^{(T)}$
        \item Incomplete trajectories receive projected rewards $r_i^{(\check{T})}$ that are typically lower
        \item This difference provides stronger learning signals for actions leading to completion
    \end{itemize}
\end{enumerate}

\subsubsection{The EGRPO Objective}
Upon the assumptions and derivations, the complete EGRPO objective extends GRPO and optimizes both terminal and penultimate steps jointly:
\begin{equation}
\J_{\text{EGRPO}}(\theta) = \gamma \J^{(T)}(\theta) + \J^{(T-1)}(\theta)
\end{equation}\\
Where each component $\J^{(t)}(\theta)$ for $t \in \{T-1, T\}$ is:
\begin{align}
\J^{(t)}(\theta) &= \mathbb{E}_{(q,a)\sim\mathcal{D}} \mathbb{E}_{\{o_i^{(t)}\}_{i=1}^{G}} \Biggl[ \frac{1}{G}\sum_{i=1}^{G}\frac{1}{|o_i^{(t)}|} \sum_{k=1}^{|o_i^{(t)}|} \min\Bigl( \rho_{i,k}^{(t)}\hat{A}_{i}^{(t)}, \text{clip}(\rho_{i,k}^{(t)}, 1-\varepsilon, 1+\varepsilon)\hat{A}_{i}^{(t)} \Bigr) \Biggr] \notag
\end{align}
\begin{align}
\text{with } \{\tilde{o}_j^{(t)}\}_{j=1}^{\tilde{G}} &\sim \pi_{\theta_{\text{old}}}(\cdot \mid q) \quad \text{(Total generated responses prior to sampling)} \notag\\
\text{and } \{o_i^{(t)}\}_{i=1}^{G} &\subseteq \{\tilde{o}_j^{(t)}\}_{j=1}^{\tilde{G}} \quad \text{(Sampled $G$ outputs by uniform sampling)}
\end{align}
The probability ratio is:
\begin{equation}
\rho_{i,k}^{(t)}(\theta) = \frac{\pi_\theta(o_{i,k}^{(t)}|q, o_{i,<k}^{(t)})}{\pi_{\theta_{\text{ref}}}(o_{i,k}^{(t)}|q, o_{i,<k}^{(t)})}
\end{equation}
\noindent Where k indexes the token position within each output sequence.\\
\noindent The normalized advantage is:
\begin{equation}
\hat{A}_{i}^{(t)} = \frac{\hat{r}_{i}^{(t)} - \mu^{(t)}}{\sigma^{(t)}}
\end{equation}
As derived in Section~\ref{sec:term_proj}, rewards are defined as:
\begin{equation}
\hat{r}_{i}^{(T)} = r_i^{(T)}, \quad \hat{r}_{i}^{(T-1)} = r_i^{(T-1)} + r_i^{(\check{T})}
\end{equation}
\noindent EGRPO introduces several key differences over standard GRPO to handle episodic tasks. The terminal step projection $r_i^{(\check{T})}$ enables efficient learning from variable-length trajectories and partial episodes, crucial for CoS where the SFT model suffers with not always reaching terminal states during exploration. By focusing optimization on penultimate-to-terminal transitions across diverse trajectory lengths, the model implicitly learns productive actions throughout entire trajectories, leveraging the dynamic programming principle stated in Assumption 3. The two-stage group sampling combines uniform sampling from $\tilde{G}$ to $G$ candidates to preserve diversity with random selection from top-N candidates to ensure high-quality actions while maintaining exploration. Together, these modifications extend GRPO's single-step optimization framework to episodic settings while maintaining computational efficiency through selective focus on critical state transitions.

\newpage

\subsubsection{Reward Function}
\begin{table}[h]
  \Large  
  \centering
  \resizebox{10cm}{!}{%
    \begin{tabular}{@{}lll c@{}}
      \toprule
      \textbf{Reward func.} & \textbf{Step} & \textbf{Condition} & \textbf{Return} \\
      \midrule
      \multirow{7}{*}{\parbox{1.5cm}{\centering \textit{Accuracy\\reward}\\[2pt]}}
        &        & Exception  & $-1$ \\
        \cmidrule(l){2-4}
        & \multirow{4}{*}{Scroll} 
            & Valid scroll                                           & $+2$ \\
        &   & Invalid scroll                                        & $-2$ \\
        &   & Valid scroll after $>\frac23$ pages read              & $2\times\frac{\text{pages\_read}}{\text{max\_page\_num}}$ \\
        &   & Scroll when all pages visited                         & $-4$ \\
        \cmidrule(l){2-4}
        & \multirow{2}{*}{Answer} 
            & Valid Answer                                          & $w\!\times\!\text{ANLS Score}$ \\
        &   & Answer $\ge 4\times$ GT length                        & $-1$ \\
      \midrule
      \multirow{5}{*}{\parbox{1.5cm}{\centering \textit{Format\\reward}\\[2pt]}}
        &      & Base score                                      & $1$ \\
        & Answer  & Valid <answer> tag                     & $+4$ \\
        &         & Valid <think> tag                      & $+2$ \\
        \cmidrule(l){2-4}
        & \multirow{2}{*}{Scroll} 
                   & Valid <scroll> tag                     & $+2$ \\
        &          & Valid <think> / <note> / Scroll Value & $+1/+1/+2$ \\
      \bottomrule
    \end{tabular}
  }
  \caption{Accuracy and format rewards for EGRPO. Maximum reward \(w\) is set to 7 for each.}
  \label{tab:reward}
\end{table}

For EGRPO, we design two reward functions to evaluate both scroll and answer step, as shown in Table~\ref{tab:reward}. For answer step, Accuracy reward utilizes the ANLS metric\cite{biten2019scenetextvisualquestion} to evaluate the correctness of the final answer provided at the episode's conclusion. To encourage exploration in the early steps and to penalize reading entire documents, we have set a decaying step accuracy and a stronger penalty for reading entire documents. Format reward monitors the adherence to structural conventions of every step.
\newpage

\subsection{Pseudocode of EGRPO}
\label{sec:egrpo:algo}

\begin{algorithm}[!h]
  \caption{Episodic Group Relative Policy Optimization (EGRPO)}
  \label{alg:egpro_alg}
  \small
  \begin{algorithmic}
    \Statex \textbf{Input:}
    \Statex \quad $q$ \hfill\Comment{user's query}
    \Statex \quad $\mathit{imgs} = \{\mathit{Image}_1, \dots, \mathit{Image}_P\}$ \hfill\Comment{ordered pages}
    \Statex \quad $\mathit{maxSteps}$ \hfill\Comment{maximum episode length}
    \Statex \quad $\mathcal{R}(\cdot)$ \hfill\Comment{reward function}
    \Statex \quad $\Call{TransitionFn}{\cdot}$ \hfill\Comment{transition function}
    \Statex \quad $\tilde{G}$ \hfill\Comment{original candidate groups}
    \Statex \quad $G$ \hfill\Comment{uniformly sampled subset size}
    \Statex \quad $N$ \hfill\Comment{top-N for ranking}
    \Statex \quad $\gamma$ \hfill\Comment{terminal step weight}
    \Statex \quad $\varepsilon$ \hfill\Comment{clipping parameter}
    \Statex\textbf{Output:} $\mathcal{L}_{\text{EGPRO}}$ \hfill\Comment{policy gradient loss}
    \Statex\rule{\linewidth}{.2pt}
    \Statex \textbf{Initialization:}
    \State $\mathit{step} \gets 0$ \Comment{Episode step counter}
    \State $\mathit{page} \gets [0]^{\times G}$ \Comment{Start at first page for all groups}
    \State $\mathit{scroll} \gets [0]^{\times G}$ \Comment{Initial scroll values}
    \State $\mathit{notes} \gets [\varnothing]^{\times G}$ \Comment{Empty note buffers}
    \State $\mathit{visited} \gets \text{False}^{G\times P}$ \Comment{Visitation matrix}
    \State $\mathit{rewards} \gets [\varnothing]^{\times G}$
    \State $\mathit{trajectory} \gets [\,]^{G}$ \Comment{Page trajectories}
    \State $\mathit{maxSteps} \gets \min\!\bigl(P,\,\textit{maxSteps}\bigr)$
    \Statex
    \State \textcolor{blue}{// \textit{Initialize with transition function (scroll to first page)}}
    \State $\mathit{c} \gets (q, \mathit{imgs}, \mathit{page}, \mathit{scroll}, \mathit{notes}, \mathit{visited})$
    \State $(\mathit{prompts},\ \mathit{imgInputs},\ \mathit{page},\ \mathit{visited},\ \mathit{trajectory},\ \mathit{done},\ \mathit{validScroll}) \gets$ \Call{TransitionFn}{$\mathit{c}$}
    \State $\mathit{buffer} \gets [\,]$ \Comment{Initialize empty buffer}
    \Statex \rule{\linewidth}{.2pt}
    \State \textcolor{blue}{// \textit{Generate CoS trajectories}}
    \While{$\mathit{step} < \mathit{maxSteps}$ \textbf{and} $\neg\mathit{done}$}
      \State $\{\tilde{o}_j\}_{j=1}^{\tilde{G}} \sim \pi_{\theta_{\text{old}}}(\cdot | \mathit{imgInputs}, \mathit{prompts})$ \Comment{Generate $\tilde{G}$ candidates}
      \State $\mathit{rewards}_{\tilde{G}} \gets \mathcal{R}(\{\tilde{o}_j\}_{j=1}^{\tilde{G}})$ 
      \State $\{o_i\}_{i=1}^{G} \gets \Call{OrderedUniformSample}{\{\tilde{o}_{\mathit{indices}[i]}\}_{i=1}^{G},\, \mathit{rewards}_{\tilde{G}}, G}$ \Comment{Ordered uniform sampling by rewards}
      \Statex
      \State \textcolor{blue}{// \textit{Parse all outputs in the uniformly sampled group, G}}
      \For{$i \in \{1, \ldots, G\}$}
        \State $(\mathit{cur\_note}_i, \mathit{scroll}_i, \mathit{answer}_i) \gets$ \Call{Parse}{$o_i$}
        \State $\mathit{notes}[i] \gets \mathit{notes}[i] \cup \{\mathit{cur\_note}_i\}$
        \State $\mathit{rewards}[i] \gets \mathit{rewards}_{\tilde{G}}[\mathit{indices}[i]]$
      \EndFor
      \State $\mathit{buffer} \gets \mathit{buffer} \cup \{(\{o_i\}_{i=1}^{G}, \mathit{rewards}, \mathit{prompts}, \mathit{imgInputs}, \mathit{page}, \mathit{scroll}, \mathit{notes}, \mathit{visited})\}$
      \Statex
      \State \textcolor{blue}{// \textit{Select best from top-N for next state}}
      \State $\mathit{top\_N} \gets$ \Call{TopN}{$\mathit{rewards}, N$}
      \State $i^* \gets$ \Call{RandomChoice}{$\mathit{top\_N}$} \Comment{Select the next state index}
    \Statex
    \State \textcolor{blue}{// \textit{Broadcast selected action to all groups}}
    \State $\mathit{page} \gets [\mathit{page}[i^*]]^{\times G}$ \Comment{All groups move to selected page}
    \State $\mathit{scroll} \gets [\mathit{scroll}[i^*]]^{\times G}$ \Comment{All use selected scroll}
    \State $\mathit{notes} \gets [\mathit{notes}[i^*]]^{\times G}$ \Comment{All use selected notes}
    \State $\mathit{visited} \gets [\mathit{visited}[i^*]]^{\times G}$ \Comment{Copy visitation state}
    \Statex
    \State \textcolor{blue}{// \textit{Transition function with batched input}}
    \State $\mathit{c} \gets (q, \mathit{imgs}, \mathit{page}, \mathit{scroll}, \mathit{notes}, \mathit{visited})$
    \State $(\mathit{prompts}, \mathit{imgInputs}, \mathit{page}, \mathit{visited}, \mathit{trajectory}, \mathit{done}, \mathit{validScroll}) \gets$ \Call{Transition\_Fn}{$\mathit{c}$}
    \State $\mathit{step} \gets \mathit{step} + 1$
    \EndWhile
    \algstore{egpro}
  \end{algorithmic}
\end{algorithm}

\newpage

\begin{algorithm}[!th]
  \ContinuedFloat
  \caption{Episodic Group Relative Policy Optimization (EGPRO) (continued)}
  \small
  \begin{algorithmic}
    \algrestore{egpro}
    \State $T \gets step$
    \Comment{Assign terminal step}
    \Statex
    \State \textcolor{blue}{// \textit{Terminal step loss}}
    \State $(\{o_i^{(T)}\}_{i=1}^{G}, \mathit{rewards}^{(T)}, \mathit{prompts}^{(T)}, \mathit{imgInputs}^{(T)}, \ldots) \gets \mathit{buffer}[T]$
    \State $\mu^{(T)} \gets \text{mean}(\mathit{rewards}^{(T)})$
    \State $\sigma^{(T)} \gets \text{std}(\mathit{rewards}^{(T)})$
    \For{$i \in \{1, \ldots, G\}$}
      \State $\hat{A}_i^{(T)} \gets \frac{\mathit{rewards}_i^{(T)} - \mu^{(T)}}{\sigma^{(T)} + \epsilon}$ \Comment{Normalize advantages}
      \State $\rho_i^{(T)} \gets \frac{\pi_\theta(o_i^{(T)}|\mathit{imgInputs}^{(T)}, \mathit{prompts}^{(T)})}
      {\pi_{\theta_{\text{ref}}}(o_i^{(T)}|\mathit{imgInputs}^{(T)}, \mathit{prompts}^{(T)})}$ \Comment{Probability ratio}
    \EndFor
    \State $\mathcal{L}^{(T)} \gets -\frac{1}{G}\sum_{i=1}^{G}\min\left(\rho_i^{(T)}\hat{A}_i^{(T)}, \text{clip}(\rho_i^{(T)}, 1\!-\!\varepsilon, 1\!+\!\varepsilon)\hat{A}_i^{(T)}\right)$
    \Statex

    \State \textcolor{blue}{// \textit{Terminal step projection from Penultimate step to estimate the future reward}}
    \State $(\{o_i^{(T-1)}\}_{i=1}^{G}, \mathit{rewards}^{(T-1)}, \mathit{prompts}^{(T-1)}, \mathit{imgInputs}^{(T-1)}, \ldots) \gets \mathit{buffer}[T-1]$ \Comment{Get unsampled penultimate step}
    \For{$i \in \{1, \ldots, G\}$}
      \State $o_i^{(\check{T})} \sim \pi_{\theta_{\text{old}}}(\cdot | \mathit{imgInputs}^{(T-1)}, \mathit{prompts}^{(T-1)})$ \Comment{Generate terminal state from each penultimate step}
      \State $\mathit{rewards}_i^{(\check{T})} \gets \mathcal{R}(o_i^{(\check{T})})$
      \State $\hat{\mathit{rewards}}_i^{(T-1)} \gets \mathit{rewards}_i^{(T-1)} + \mathit{rewards}_i^{(\check{T})}$ \Comment{Sum terminal state reward with the penultimate step to model return}
    \EndFor
    \State $\mu^{(T-1)} \gets \text{mean}(\hat{\mathit{rewards}}^{(T-1)})$
    \State $\sigma^{(T-1)} \gets \text{std}(\hat{\mathit{rewards}}^{(T-1)})$
    \For{$i \in \{1, \ldots, G\}$}
      \State $\hat{A}_i^{(T-1)} \gets \frac{\hat{\mathit{rewards}}_i^{(T-1)} - \mu^{(T-1)}}{\sigma^{(T-1)} + \epsilon}$
      \State $\rho_i^{(T-1)} \gets \frac{\pi_\theta(o_i^{(T-1)}|\mathit{imgInputs}^{(T-1)}, \mathit{prompts}^{(T-1)})}{\pi_{\theta_{\text{ref}}}(o_i^{(T-1)}|\mathit{imgInputs}^{(T-1)}, \mathit{prompts}^{(T-1)})}$
    \EndFor
    \State $\mathcal{L}^{(T-1)} \gets -\frac{1}{G}\sum_{i=1}^{G}\min\left(\rho_i^{(T-1)}\hat{A}_i^{(T-1)}, \text{clip}(\rho_i^{(T-1)}, 1\!-\!\varepsilon, 1\!+\!\varepsilon)\hat{A}_i^{(T-1)}\right)$
    \Statex
    \State $\mathcal{L}_{\text{EGPRO}} \gets \gamma \cdot \mathcal{L}^{(T)} + \mathcal{L}^{(T-1)}$ \Comment{$\gamma > 1$ emphasizes terminal step loss}
    \State \Return $\mathcal{L}_{\text{EGPRO}}$
  \end{algorithmic}
\end{algorithm}

As shown in Algorithm~\ref{alg:egpro_alg}, EGPRO is an extension of GRPO designed for episodic tasks tailored to sequential navigation through documents to answer queries. The algorithm initializes $G$ groups at the first page with empty note buffers and tracks visitation states across $P$ pages. During each episode step (up to $\mathit{maxSteps}$), EGPRO generates $\tilde{G}$ candidate outputs from the current policy $\pi_{\theta_{\text{old}}}$, evaluates them with reward function $\mathcal{R}$, and performs ordered uniform sampling to select $G$ diverse candidates. Each candidate output is parsed to extract notes, scroll positions, and potential answers, with rewards stored in a trajectory buffer. The algorithm then randomly selects one action from the top-$N$ highest-rewarded candidates and broadcasts this choice to all groups. The transition function updates the prompts, image inputs, page positions, and visitation matrix based on the selected action. After the training policy returns the answer, EGPRO computes a two-step loss: for the terminal step $T$, the loss is the conventional GRPO loss for the generated answer; for the penultimate step $T-1$, it estimates future returns by projecting terminal states from each candidate without sampling and incorporates these projected rewards into the advantage calculation.
\newpage

\section{Chain of Scroll trajectory examples of SCoPE VLM}\label{sec:full_main}
\begin{figure*}[!htbp] 
\begin{center}
\resizebox{1.0\textwidth}{!}{%
\includegraphics[width=\textwidth]{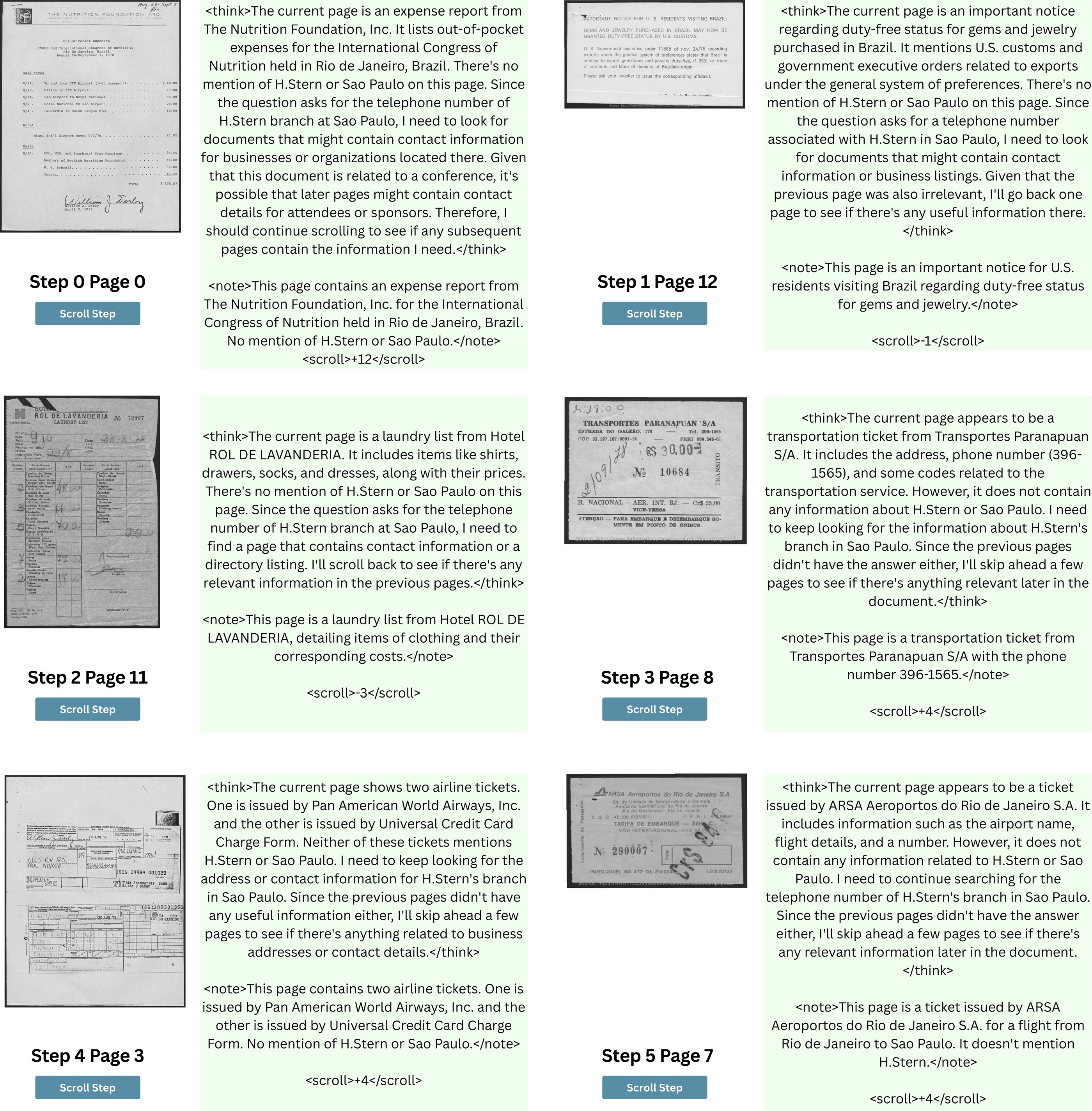}
}
\caption{Model responses from Step 0 to Step 5 of SCoPE VLM SFT to the question: What is the telephone number of H.Stern branch at Sao Paulo?}
\label{fig:appendix-sft-answer1}
\end{center}
\end{figure*}

\begin{figure*}[!htbp]
\begin{center}
\resizebox{1.0\textwidth}{!}{%
\includegraphics[width=\textwidth]{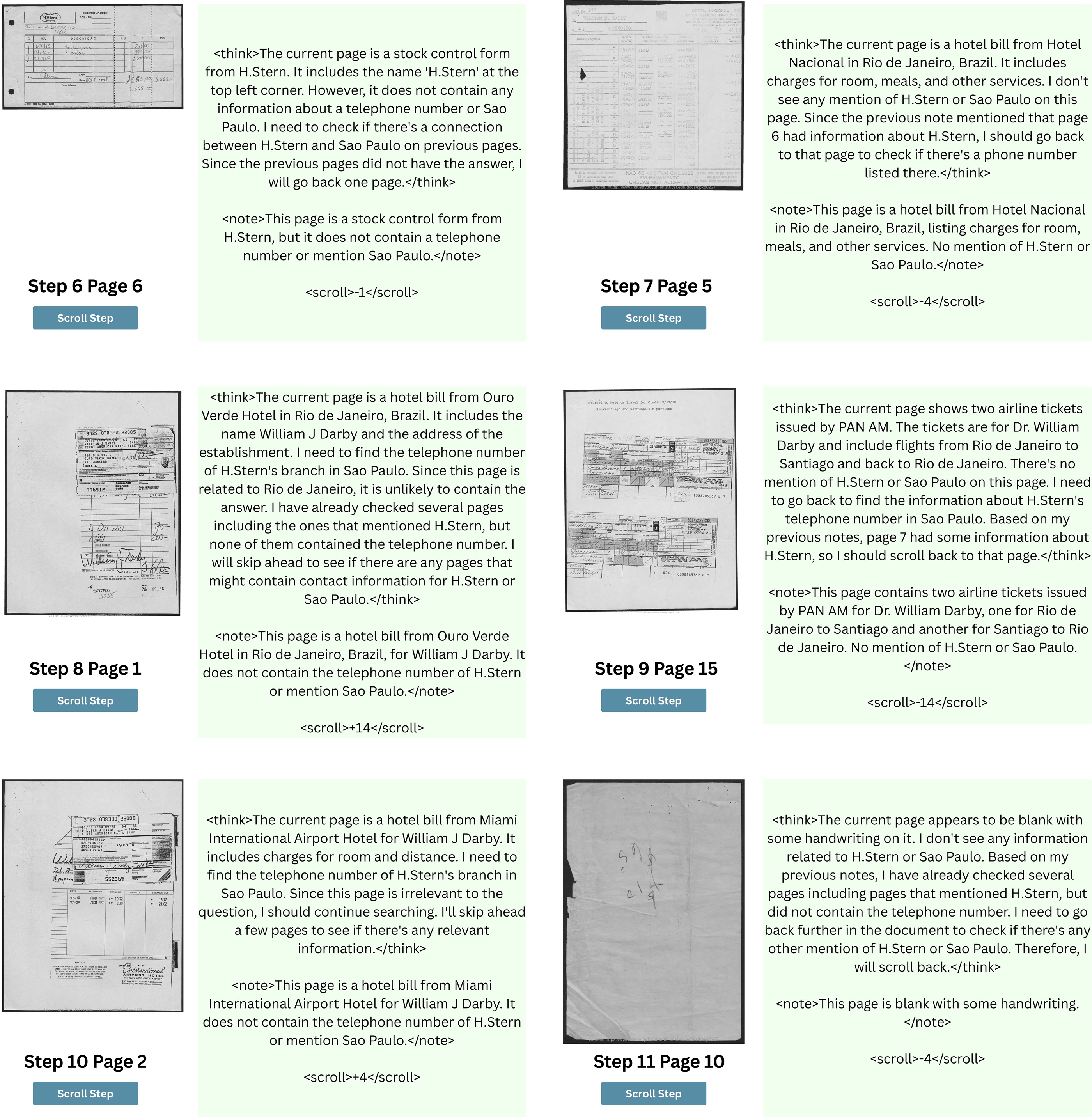}
}
\caption{Model responses from Step 6 to Step 11 of SCoPE VLM SFT to the question: What is the telephone number of H.Stern branch at Sao Paulo?}
\label{fig:appendix-sft-answer2}
\end{center}
\end{figure*}

\begin{figure*}[!htbp]
\begin{center}
\resizebox{1.0\textwidth}{!}{%
\includegraphics[width=\textwidth]{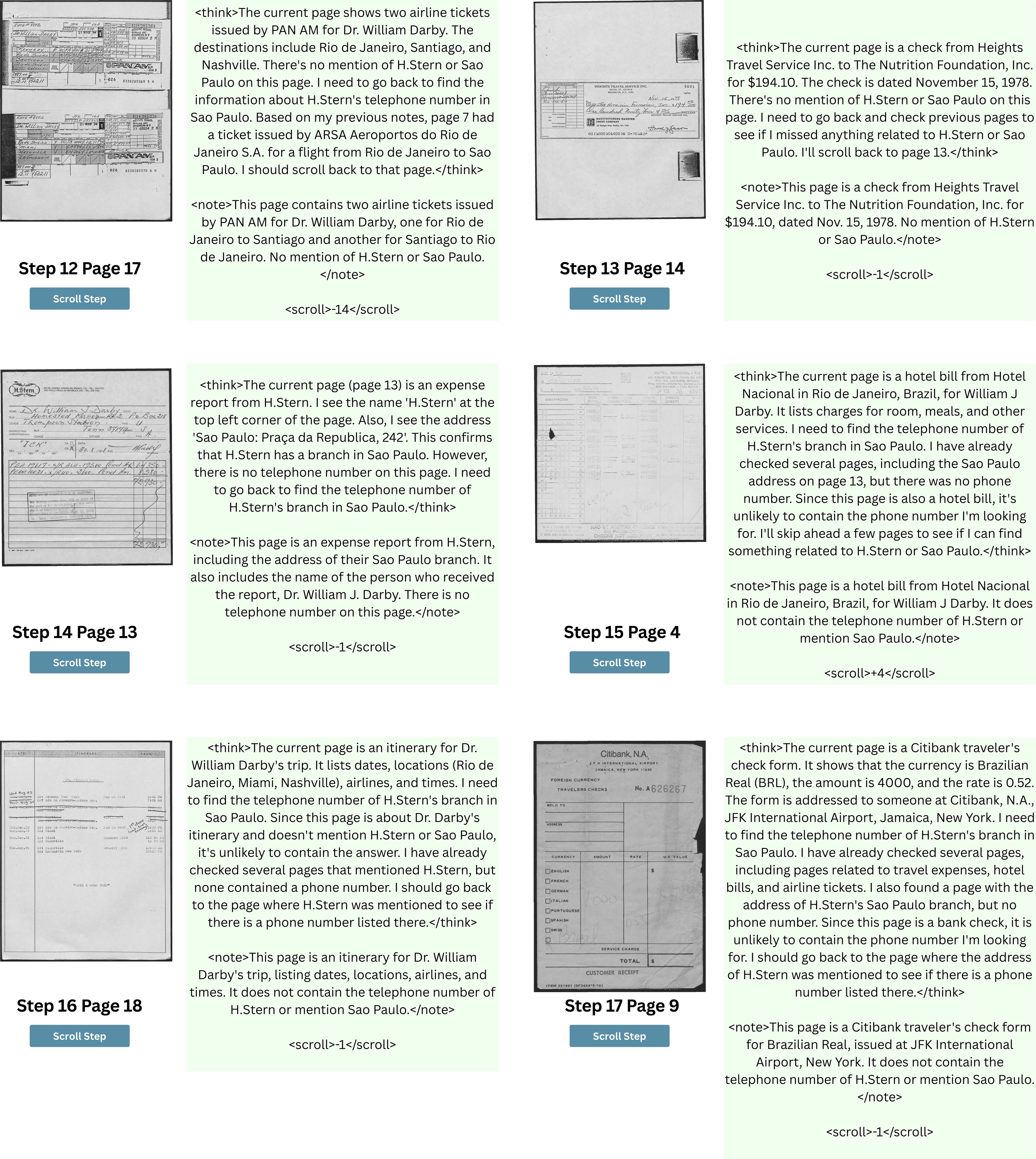}
}
\caption{Model responses from Step 12 to Step 17 of SCoPE VLM SFT to the question: What is the telephone number of H.Stern branch at Sao Paulo?}
\label{fig:appendix-sft-answer3}
\end{center}
\end{figure*}

\begin{figure*}[!t]
\begin{center}
\resizebox{0.5\textwidth}{!}{%
\includegraphics[width=\textwidth]{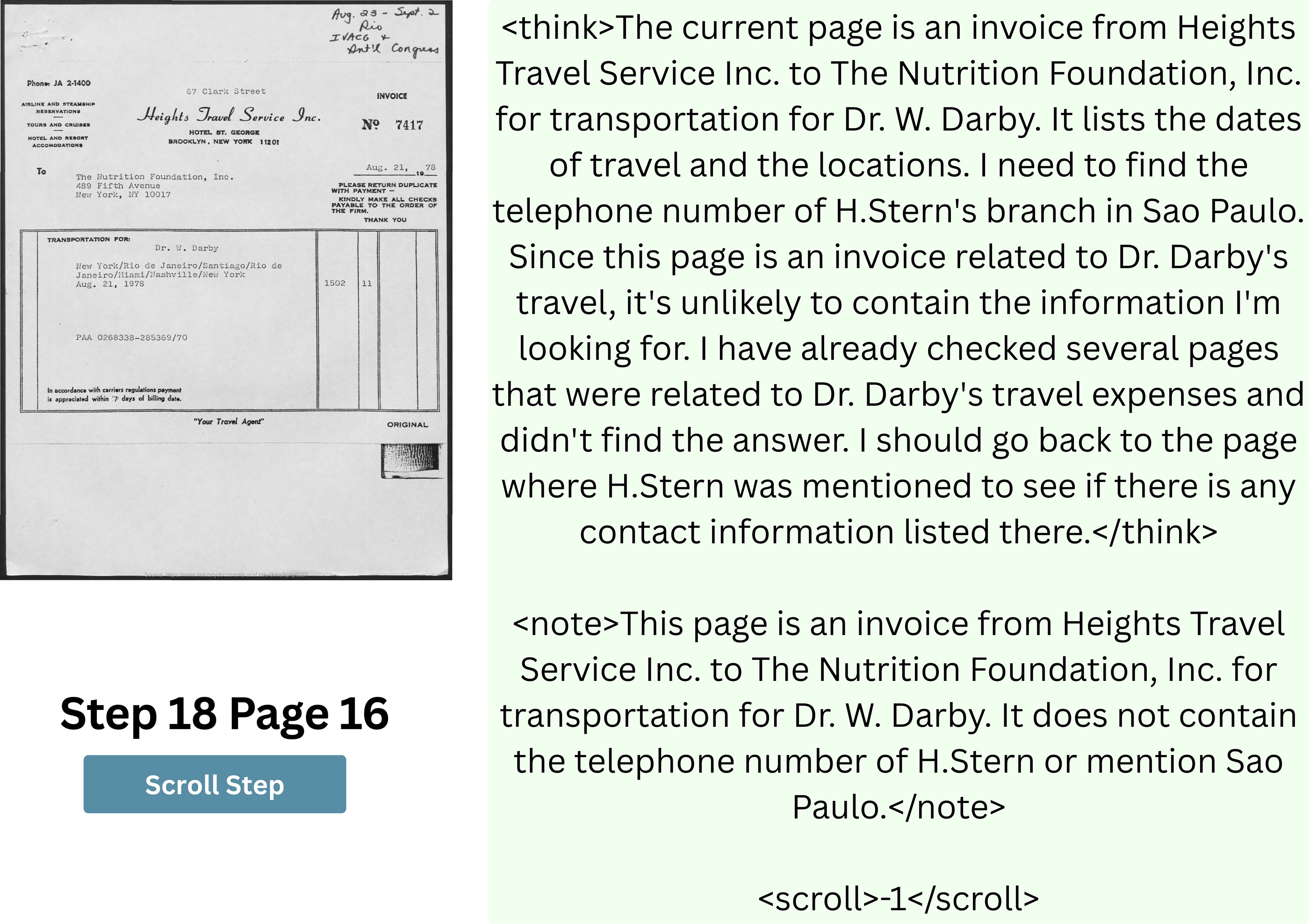}
}
\caption{Model responses at Step 18 of SCoPE VLM SFT to the question: What is the telephone number of H.Stern branch at Sao Paulo? Ultimately, the SFT model returns the scroll action, failing to return the correct answer.}
\label{fig:appendix-sft-answer4}
\end{center}
\end{figure*}

\begin{figure*}[!htbp]
\begin{center}
\resizebox{1.0\textwidth}{!}{%
\includegraphics[width=\textwidth]{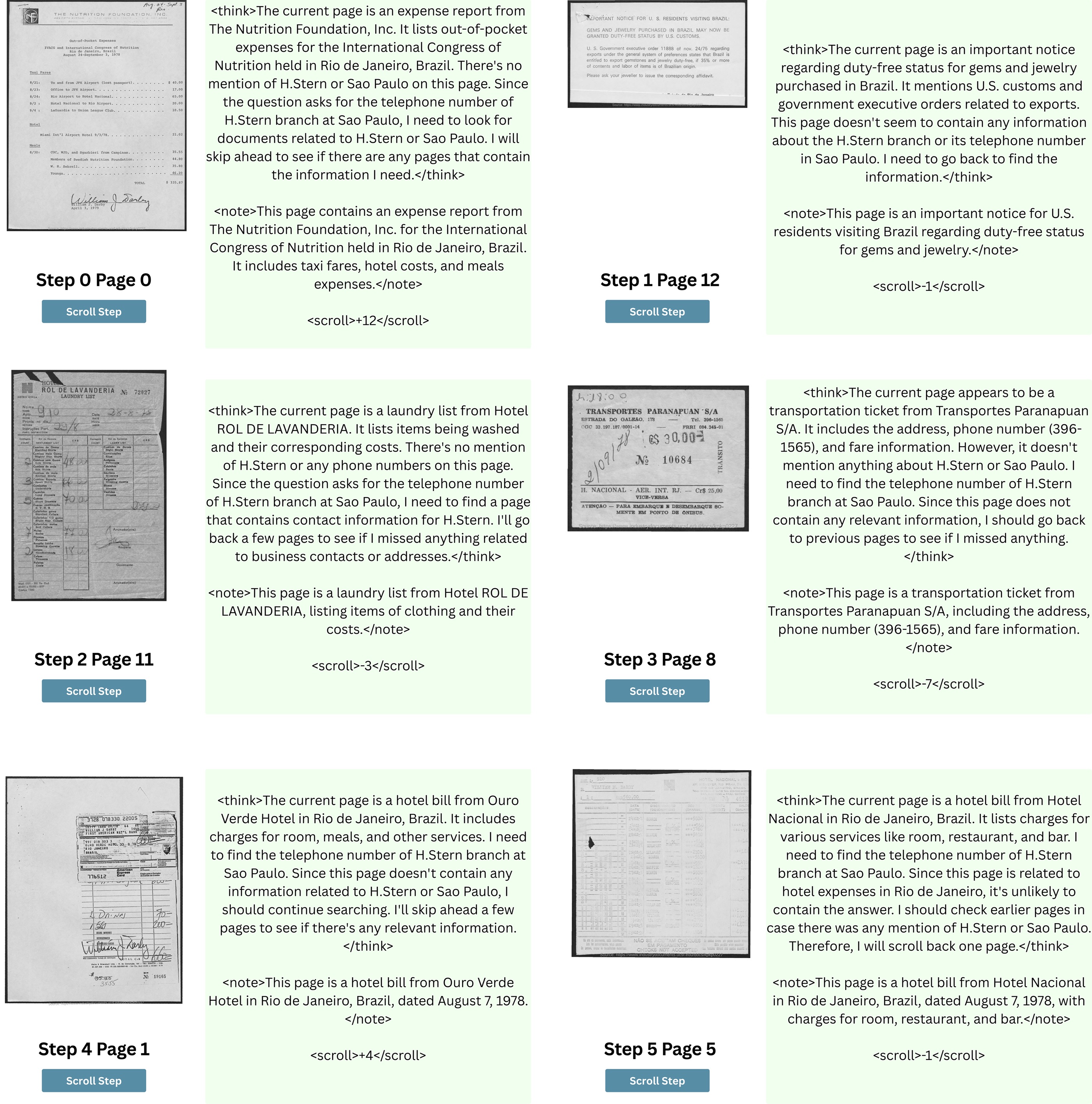}
}
\caption{Model responses from Step 0 to Step 5 of SCoPE VLM EGRPO to the question: What is the telephone number of H.Stern branch at Sao Paulo?}
\label{fig:appendix-egrpo-answer1}
\end{center}
\end{figure*}

\begin{figure*}[!htbp]
\begin{center}
\resizebox{1.0\textwidth}{!}{%
\includegraphics[width=\textwidth]{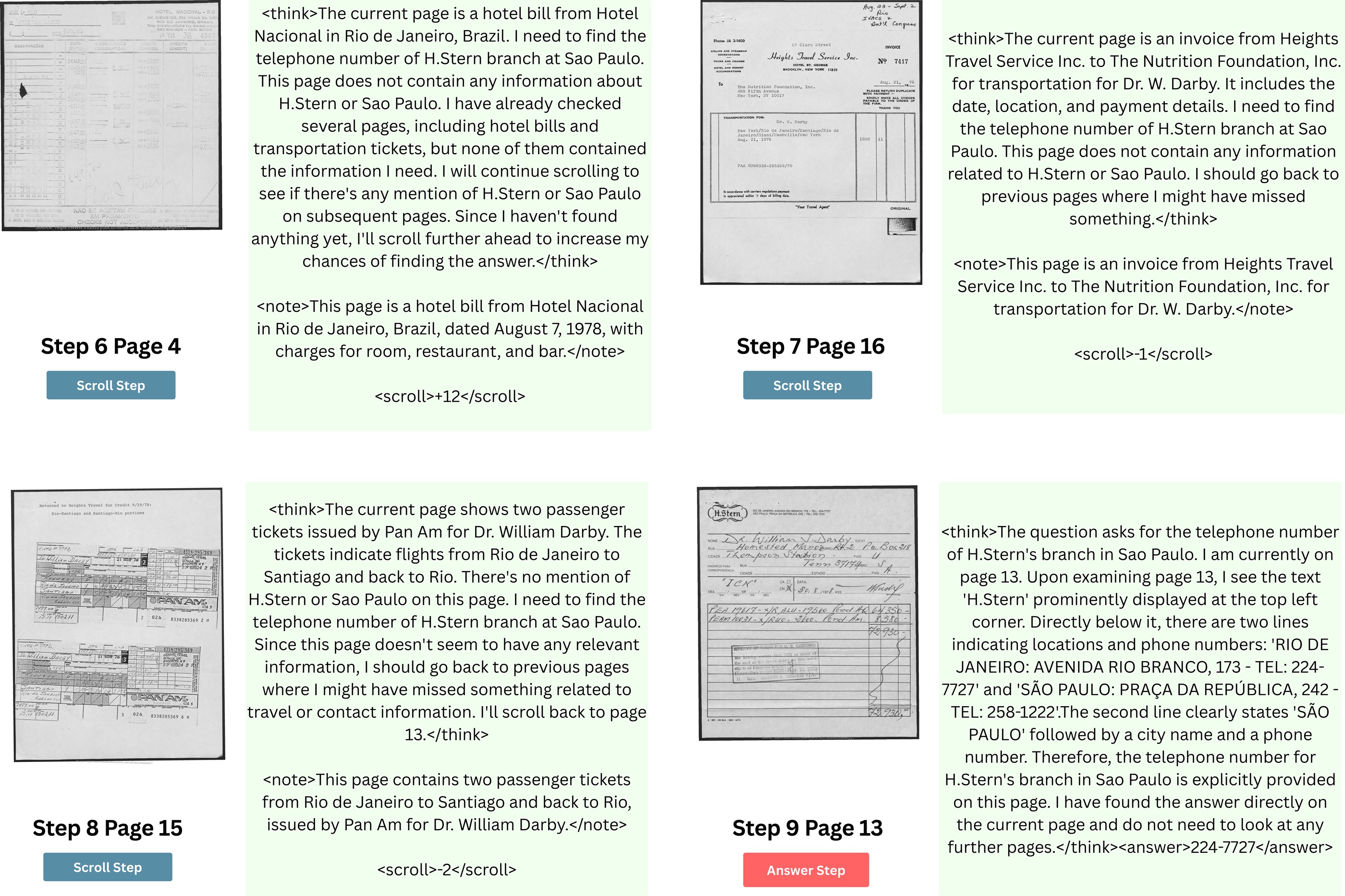}
}
\caption{Model responses from Step 6 to Step 9 of SCoPE VLM EGRPO to the question: What is the telephone number of H.Stern branch in Sao Paulo? In Step 9, SCoPE VLM EGRPO correctly answers the question, while the SFT model fails.}
\label{fig:appendix-egrpo-answer2}
\end{center}
\end{figure*}

As shown in Figure~\ref{fig:appendix-sft-answer1}, ~\ref{fig:appendix-sft-answer2}, ~\ref{fig:appendix-sft-answer3}, ~\ref{fig:appendix-sft-answer4}, ~\ref{fig:appendix-egrpo-answer1}, ~\ref{fig:appendix-egrpo-answer2}, we compare the Chain of Scroll trajectories of SCoPE VLM SFT and SCoPE VLM EGRPO for the question: "What is the telephone number of H.Stern branch at Sao Paulo?" (MP-docVQA Val-split QuestionID: 56580). SCoPE VLM SFT fails to find the answer, whereas SCoPE VLM EGRPO successfully finds the correct answer,  "224-7727".

\subsection{SCoPE VLM SFT}
\FloatBarrier 

\FloatBarrier 
\clearpage

\subsection{SCoPE VLM EGRPO}

\FloatBarrier

\clearpage

\section{Responses of SCoPE VLM SFT When No Answer Exists in the Context}
\label{sec:no_answer}
SCoPE-VLM is trained to answer "The answer cannot be found." when it concludes that the given document does not have the relevant information to answer the question.

\begin{figure*}[!htbp]
\begin{center}
\resizebox{0.85\textwidth}{!}{%
\includegraphics[width=\textwidth]{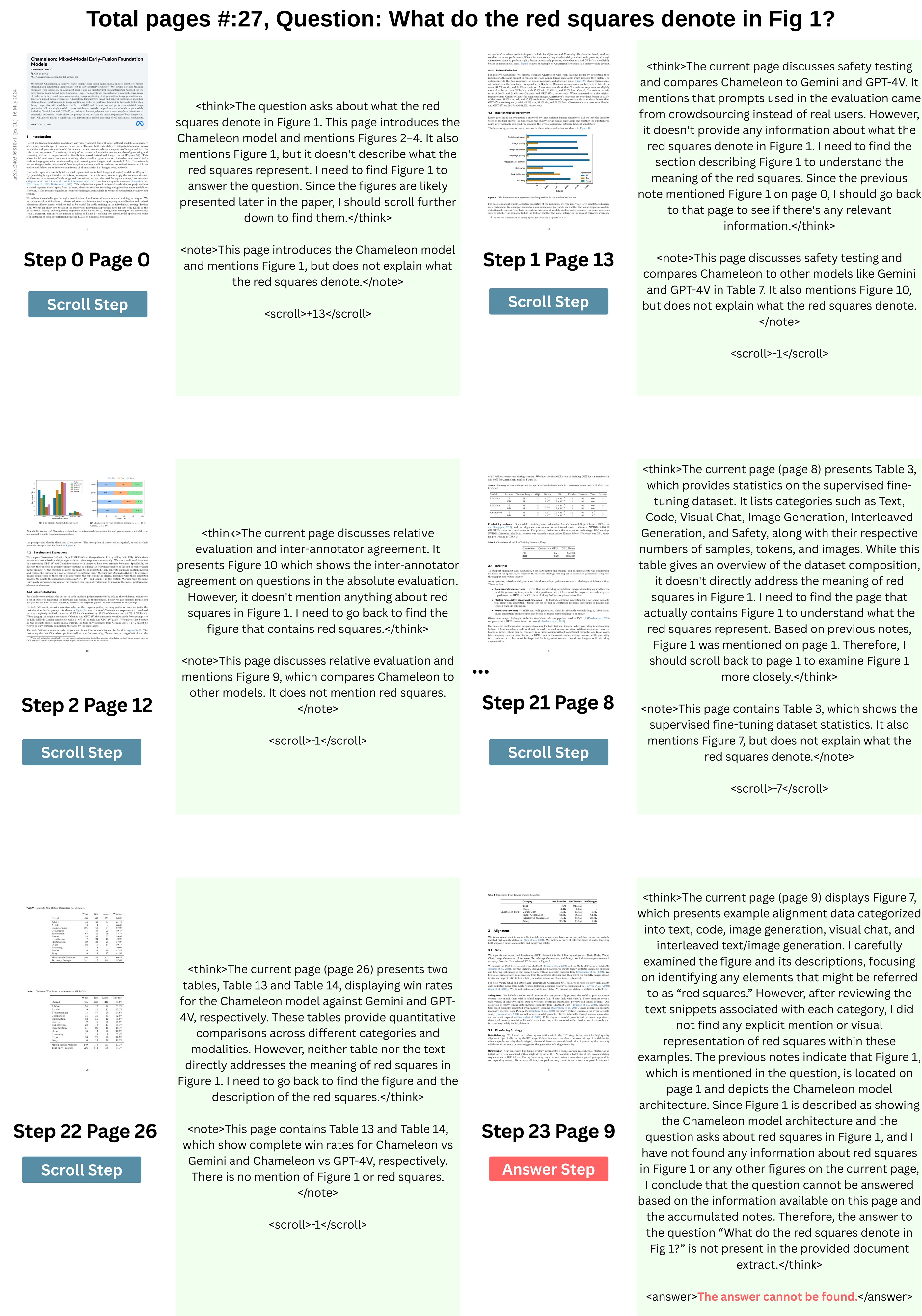}
}
\caption{SCoPE VLM SFT's answer to MMLongBench Valid 773 Question: What do the \textcolor{red}{red} squares denote in Fig 1?}
\label{fig:appendix-noanswer}
\end{center}
\end{figure*}
\newpage

\newpage
\section{Details of training hyperparameters}
\label{sec:hyperparameters}

\subsection{SCoPE VLM SFT}

\begin{table}[h]
\centering
\begin{tabular}{c|c}
\hline
\textbf{Hyperparameter} & \textbf{Value} \\
\hline
model & Qwen2.5 VL 3B Instruct \\
learning rate & 2e-5 \\
batch size & 8 \\
per device train batch size & 2 \\
gradient accumulation steps & 4 \\
max length & 8192 \\
max pixels & 1003520 \\
warmup ratio & 0.03 \\
num train epochs & 1 \\
optimizer & AdamW \\
data type & bf16 \\
gradient checkpointing & enabled \\
deepspeed & zero3 \\
\hline
\end{tabular}
\caption{Training Hyperparameters for SFT of SCoPE VLM SFT}
\label{tab:sft_hyper}
\end{table}

During the SFT stage, training is performed on 8*A100 GPU for one epoch. The full SCoPE dataset is fine-tuned with the max pixel of 1,003,520 pixels, 1280 tokens per image.
Further EGRPO training hyperparameters are shown in Table~\ref{tab:sft_hyper}.

\newpage
\subsection{SCoPE VLM EGRPO}
 
\begin{table}[h]
\centering
\begin{tabular}{c|c}
\hline
\textbf{Hyperparameter} & \textbf{Value} \\
\hline
\multicolumn{2}{c}{\textbf{General}} \\
\hline
model & SCOPE VLM 3B SFT \\
learning rate & 1e-4 \\
batch size & 1 \\
per device train batch size & 1 \\
gradient accumulation steps & 1 \\
max prompt length & 8096 \\
max completion length & 4096 \\
max pixels & 1003520 \\
warmup ratio & 0.01 \\
num train epochs & 1 (early terminated at 2500 step) \\
num iterations & 1 \\
optimizer & AdamW \\
data type & bf16 \\
gradient checkpointing & enabled \\
lr scheduler & cosine with min lr \\
min learning rate & 1e-9 \\
\hline
\multicolumn{2}{c}{\textbf{LoRA}} \\
\hline
LoRA rank & 128 \\
LoRA alpha & 16 \\
LoRA dropout & 0.05 \\
\hline
\multicolumn{2}{c}{\textbf{EGRPO}} \\
\hline
reward functions & accuracy, format \\
num generations & 8 \\
num samples per group & 4 \\
max steps & 24 \\
max window length & 2 \\
$\beta$ (KL penalty) & 0 \\
$\epsilon$ 
 (clipping) & 0.2 \\
$\gamma$ (weighing coefficient) & 3 \\
temperature & 0.9 \\
top p & 0.9 \\
top k & 50 \\
\hline
\end{tabular}
\caption{Training hyperparameters for EGRPO of SCoPE VLM EGRPO}
\label{tab:egrpo_hyper}
\end{table}

 For RL stage, we deploy EGRPO to enhance CoS performance through improved reasoning and memorization. We train our model on two H100 GPUs with LoRA configuration of r=128, alpha=16, dropout=0.05, and with the same max pixels of SFT stage. For training dataset, we set up training with SlideVQA and 50\% of MP-DocVQA but early terminated at 2500 steps with the maximum CoS step of 24. For ablation, the training step is further reduced to 1000 step. Also, we restrict the page visit to once in the transition function to limit extensive exploration for training. Further EGRPO training hyperparameters are shown in Table~\ref{tab:egrpo_hyper}.

 \newpage
\subsection{GRPO for ablation study}
 
\begin{table}[h]
\centering
\begin{tabular}{c|c}
\hline
\textbf{Hyperparameter} & \textbf{Value} \\
\hline
\multicolumn{2}{c}{\textbf{General}} \\
\hline
model & SCOPE VLM 3B SFT \\
learning rate & 1e-4 \\
batch size & 1 \\
per device train batch size & 1 \\
gradient accumulation steps & 1 \\
max prompt length & 8096 \\
max completion length & 4096 \\
max pixels & 1003520 \\
warmup ratio & 0.01 \\
num train epochs & 1 (early terminated at 1000 step) \\
num iterations & 1 \\
optimizer & AdamW \\
data type & bf16 \\
gradient checkpointing & enabled \\
lr scheduler & cosine with min lr \\
min learning rate & 1e-9 \\
\hline
\multicolumn{2}{c}{\textbf{LoRA}} \\
\hline
LoRA rank & 128 \\
LoRA alpha & 16 \\
LoRA dropout & 0.05 \\
\hline
\multicolumn{2}{c}{\textbf{EGRPO}} \\
\hline
reward functions & accuracy, format \\
num generations & 4 \\
max steps & 24 \\
max window length & 1 or 2 \\
$\beta$ (KL penalty) & 0 \\
$\epsilon$ 
 (clipping) & 0.2 \\
$\gamma$ (weighing coefficient) & 3 \\
temperature & 0.9 \\
top p & 0.9 \\
top k & 50 \\
\hline
\end{tabular}
\caption{Training hyperparameters for GRPO}
\label{tab:grpo_hyper}
\end{table}

 For the ablation study, we train SCoPE VLM SFT with GRPO objectives. We train the model in a similar manner as SCoPE VLM EGRPO for 1,000 steps with a maximum CoS step of 24. Further GRPO training hyperparameters are shown in Table~\ref{tab:grpo_hyper}.

\newpage

\section{Details of inference hyperparameters}
\label{sec:inference-hyperparameters}

\subsection{Multi page document question answering experiments: Table ~\ref{tab:main}, ~\ref{tab:vram}, ~\ref{tab:inference_strategy_comparison}, ~\ref{tab:training_method_comparison}}

\begin{table}[h]
\centering
\begin{tabular}{c|c}
\hline
\textbf{Hyperparameter} & \textbf{Value} \\
\hline
temperature & 0 \\
top\_p & 1.0(default) \\
num\_beams & 1 \\
max\_new\_tokens & 1280 \\
Table ~\ref{tab:main},  ~\ref{tab:inference_strategy_comparison}, ~\ref{tab:training_method_comparison} max\_visit\_count & 2 \\
Table ~\ref{tab:vram} max\_visit\_count & 1 \\
\hline
\end{tabular}
\caption{Hyperparameters for multi page document question answering experiments: Table ~\ref{tab:main}, ~\ref{tab:vram}, ~\ref{tab:inference_strategy_comparison}, ~\ref{tab:training_method_comparison} }
\label{tab:document_inference}
\end{table}

\subsection{GUI control experiments: Table ~\ref{tab:AitZ}}

\begin{table}[h]
\centering
\begin{tabular}{c|c}
\hline
\textbf{Hyperparameter} & \textbf{Value} \\
\hline
temperature & 0.1 \\
top\_p & 0.9 \\
do\_sample & True \\
max\_new\_tokens & 1024 \\
repetition\_penalty & 1.1 \\
seed & 2020 \\
\hline
\end{tabular}
\caption{Hyperparameters for GUI control experiments}
\label{tab:GUI_inference}
\end{table}

\section{Details of the Evaluation Setup in Section~\ref{sec:Experiments} and Examples}
\label{sec:evaluation_setup}

\subsection{Experimental Setup for Table~\ref{tab:main} and~\ref{tab:training_method_comparison}}
\label{table2_setup}

To ensure fair comparison under identical hardware constraints, we standardized VRAM usage across all evaluated models. In this experiment, we applied a fixed image token budget of 2,560 tokens per inference step (corresponding to a maximum of 2,007,040 pixels). The two inference methods in Table~\ref{tab:main} implement VRAM standardization as follows:

\textbf{Chain of Scroll (CoS):} Processes one image per inference step through sequential inference to reach the final conclusion. Therefore, each image in CoS inference receives the full allocation of 2,560 tokens.

\textbf{Multi-Image (MI):} Processes all images simultaneously in a single inference pass. The 2,560 token budget is distributed across all provided images. Consequently, the total token count for the entire image set cannot exceed this limit, resulting in lower per-image token allocation compared to CoS.

\noindent In addition, for all methods in this table, models were allowed to visit each image up to twice.

\subsection{Experimental Setup for Table~\ref{tab:vram}}
\label{table3_setup}

In contrast to the setup in Table~\ref{tab:main}, we did not enforce uniform VRAM usage in Table~\ref{tab:vram}. Instead, our primary objective was to evaluate the trade-off between model performance and VRAM consumption. To achieve this, we imposed a limit of 1,280 tokens per individual image. In addition, the maximum visit is restricted to once for efficiency and the LoRA for SCoPE VLM EGRPO has been merged in the inference time. 

For multi-image models, each image within a single concurrent input can be allocated up to 1,280 tokens. For CoS models, input images are similarly limited to a maximum of 1,280 tokens. This setup was designed to analyze how efficiently each model utilizes a fixed per-image token budget and its impact on VRAM usage and task performance.

\subsection{Experimental Setup for Table~\ref{tab:AitZ}}

\begin{table}[h]
\centering
\begin{tabular}{c|c}
\hline
\textbf{Hyperparameter} & \textbf{Value} \\
\hline
learning rate & 1e-5 \\
batch size & 8 \\
per device train batch size & 2 \\
gradient accumulation steps & 4 \\
max length & 8192 \\
max pixels & 1003520 \\
warmup ratio & 0.03 \\
num train epochs & 1 \\
optimizer & AdamW \\
data type & bf16 \\
gradient checkpointing & enabled \\
deepspeed & zero3 \\
\hline
\end{tabular}
\caption{Hyperparameters for SFT Training of SCoPE VLM and Qwen 2.5 VL for Table~\ref{tab:AitZ} in Section~\ref{sec:Experiments}}
\label{aitzhyper}
\end{table}

This experiment evaluates model performance on practical GUI navigation tasks using the Android in the Zoo (AitZ) benchmark. The evaluation protocol follows standard AitZ procedures, measuring task completion capability by comparing model-generated actions against ground truth for achieving target goals on given UI screens.

SCoPE VLM SFT, SCoPE VLM EGRPO, and Qwen2.5-3B-VL are fine-tuned with the hyperparameters shown in the Table ~\ref{aitzhyper}. Also, the max image token is set to 1,280 tokens per image. The LoRA for SCoPE VLM EGRPO has been merged in the inference time as well. 

\subsection{Experimental Setup for Table~\ref{tab:inference_strategy_comparison}: Performance Comparison of Serial, Random, and CoS Methods}

The ablation study is designed to isolate the effects of different information-seeking strategies. We standardized input conditions by allocating 2,560 tokens per image and limiting the maximum page visits to 2.

For this evaluation, three different inference strategies were compared:

\begin{itemize}
    \item \textbf{Serial:} Pages are presented to the model sequentially in natural order, starting from page 0.
    \item \textbf{Random:} After initial presentation of page 0, subsequent pages are provided in random order.
    \item \textbf{CoS:} Following initial input of page 0, the model autonomously determines which page to explore next based on its reasoning.
\end{itemize}

\section{Detailed Examples of Per-Image Token Limits}

To manipulate the maximum image token, we set the maximum pixel limit which is used to resize images to limit the number of image tokens. The image processor for this work follows the default processor provided by the baseline Qwen 2.5 VL series~\cite{qwen25vl}.

When an image is input, we first calculate its initial pixel count using the original height and width. If the calculated pixels exceed the maximum pixel limit, we compute a scaling factor $\beta$ as follows:

$$\beta = \sqrt{\frac{\text{height} \times \text{width}}{\text{max\_pixels}}}$$

We then divide both height and width by $\beta$, round down to the nearest multiple of 28 (the patch size), and multiply by 28 to obtain the new dimensions. The product of new height and new width becomes the final pixel count.

To illustrate this process, we provide examples for three cases under both experimental setups:

\begin{enumerate}
    \item Single image input with dimensions 5120 × 2880
    \item Ten images input, each with dimensions 1980 × 1080
    \item Ten images input, each with dimensions 720 × 144
\end{enumerate}

\subsection{Pixel Resizing for Table~\ref{tab:main},~\ref{tab:inference_strategy_comparison}, and~\ref{tab:training_method_comparison}: limiting image token usage per single inference step}

With max\_pixel = 2,007,040:

\textbf{Case 1: Single 5120 × 2880 image}
(Multi-image and CoS apply the same method for single images)
\begin{align*}
    \text{Initial Pixels} &= 5120 \times 2880 = 14,745,600 \\
    \beta &= \sqrt{\frac{14,745,600}{2,007,040}} \approx 2.71 \\
    \text{New Height} &= \left\lfloor \frac{2880 / 2.71}{28} \right\rfloor \times 28 = 1036 \\
    \text{New Width} &= \left\lfloor \frac{5120 / 2.71}{28} \right\rfloor \times 28 = 1876 \\
    \text{Final Pixels} &= 1036 \times 1876 = 1,943,536 \\
    \text{Final Tokens} &= \frac{1,943,536}{784} = 2479
\end{align*}

\textbf{Case 2: Ten 1980 × 1080 images}

For Multi-Image inference, the total pixels across all 10 images cannot exceed 2,007,040. Thus, the per-image pixel limit is 200,704:
\begin{align*}
    \text{Initial Pixels} &= 1980 \times 1080 = 2,138,400 \\
    \text{Max Pixels per Image} &= \frac{2,007,040}{10} = 200,704 \\
    \beta &= \sqrt{\frac{2,138,400}{200,704}} \approx 3.26 \\
    \text{New Height} &= \left\lfloor \frac{1080 / 3.26}{28} \right\rfloor \times 28 = 308 \\
    \text{New Width} &= \left\lfloor \frac{1980 / 3.26}{28} \right\rfloor \times 28 = 588 \\
    \text{Final Pixels} &= 308 \times 588 = 181,104 \\
    \text{Final Tokens} &= \frac{181,104}{784} = 231
\end{align*}

For CoS inference (processing one image at a time), each image has a 2,007,040 pixel limit:
\begin{align*}
    \text{Initial Pixels} &= 1980 \times 1080 = 2,138,400 \\
    \beta &= \sqrt{\frac{2,138,400}{2,007,040}} \approx 1.032 \\
    \text{New Height} &= \left\lfloor \frac{1080 / 1.032}{28} \right\rfloor \times 28 = 1036 \\
    \text{New Width} &= \left\lfloor \frac{1980 / 1.032}{28} \right\rfloor \times 28 = 1904 \\
    \text{Final Pixels} &= 1036 \times 1904 = 1,972,544 \\
    \text{Final Tokens} &= \frac{1,972,544}{784} = 2516
\end{align*}

\textbf{Case 3: Ten 720 × 144 images}

For Multi-Image inference, the per-image pixel limit is 200,704. Since each image's initial pixel count is below this limit, no downscaling is applied:
\begin{gather*}
    \text{Image Pixels} = 720 \times 144 = 103,680 \\
    \text{Per-Image Limit} = 200,704 \\
    103,680 < 200,704 \quad (\text{Below the limit}) \\
    \therefore \text{ No downscaling required}
\end{gather*}
Therefore, the original image dimensions are used for tokenization:
\begin{gather*}
    \text{Final Pixels per image} = 103,680 \\
    \text{Final Tokens per image} = \frac{103,680}{784} \approx 132
\end{gather*}

For CoS inference, the per-image limit is 2,007,040. The initial pixel count is also well below this threshold, so no downscaling is required:
\begin{gather*}
    \text{Image Pixels} = 720 \times 144 = 103,680 \\
    \text{Limit} = 2,007,040 \\
    103,680 < 2,007,040 \quad (\text{Below the limit}) \\
    \therefore \text{ No downscaling required}
\end{gather*}
The final token count is therefore identical to the Multi-Image case:
\begin{gather*}
    \text{Final Pixels per image} = 103,680 \\
    \text{Final Tokens per image} = \frac{103,680}{784} \approx 132
\end{gather*}

\subsection{Pixel Resizing for Table~\ref{tab:vram}: limiting image token usage per image}

In Table~\ref{tab:vram}, tokens are limited per image. All cases apply a 1,004,520 pixel limit per image:

\textbf{Case 1: Single 5120 × 2880 image}
\begin{align*}
    \text{Initial Pixels} &= 5120 \times 2880 = 14,745,600 \\
    \beta &= \sqrt{\frac{14,745,600}{1,004,520}} \approx 3.83 \\
    \text{New Height} &= \left\lfloor \frac{2880 / 3.83}{28} \right\rfloor \times 28 = 728 \\
    \text{New Width} &= \left\lfloor \frac{5120 / 3.83}{28} \right\rfloor \times 28 = 1316 \\
    \text{Final Pixels} &= 728 \times 1316 = 957,808 \\
    \text{Final Tokens} &= \frac{957,808}{784} = 1222
\end{align*}

\textbf{Case 2: Ten 1980 × 1080 images}

All inference methods apply the same resizing:
\begin{align*}
    \text{Initial Pixels} &= 1980 \times 1080 = 2,138,400 \\
    \beta &= \sqrt{\frac{2,138,400}{1,004,520}} \approx 1.459 \\
    \text{New Height} &= \left\lfloor \frac{1080 / 1.459}{28} \right\rfloor \times 28 = 728 \\
    \text{New Width} &= \left\lfloor \frac{1980 / 1.459}{28} \right\rfloor \times 28 = 1344 \\
    \text{Final Pixels} &= 728 \times 1344 = 978,432 \\
    \text{Final Tokens} &= \frac{978,432}{784} = 1248
\end{align*}

\textbf{Case 3: Ten 720 × 144 images}

Since each image's pixels are below the limit, no resizing is applied:

\begin{align*}
    \text{Pixels} &= 720 \times 144 = 103,680 \\
    \text{Limit} &= 1,004,520 \\
    103,680 &< 1,004,520 \quad (\text{below the limit}) \\
    &\therefore \text{No resizing required}
\end{align*}

Each image maintains its original dimensions:
\begin{align*}
\text{Final Pixels per image} &= 720 \times 144 = 103,680 \
\text{Final Tokens per image} &= \frac{103,680}{784} = 132.24 \approx 132
\end{align*}

\newpage

\section{Efficiency metrics of baseline models and SCoPE VLMs in Table~\ref{tab:main}}
\label{sec:main-performance-efficiency-appendix}

\begin{table*}[!h]
  \centering
  \resizebox{1.0\textwidth}{!}{%
  \begin{tabular}{@{}l ccccc|ccccc@{}}
    \toprule
    & \multicolumn{5}{c|}{\textbf{ANLS (\%)}} & \multicolumn{5}{c}{\textbf{Visit Ratio (\%)}} \\
    \cmidrule(lr){2-6} \cmidrule(lr){7-11}
    \textbf{Model} &
    \textbf{MP-DocVQA} & \textbf{SlideVQA} & \textbf{M3DocVQA} & \textbf{DUDE} & \textbf{MMLong} &
    \textbf{MP-DocVQA} & \textbf{SlideVQA} & \textbf{M3DocVQA} & \textbf{DUDE} & \textbf{MMLong} \\
    \midrule
    Qwen2.5 VL 3B   & 47.78 & 23.23 & 19.54 & 31.73 & 9.08  & 59.17 & 30.19 & 56.81 & 64.52 & 23.41 \\
    Qwen2.5 VL 72B  & 80.83 & 66.73 & 35.83 & 48.43 & --    & 67.77 & 50.28 & 57.47 & 84.32 & --    \\
    SCoPE VLM SFT             & 74.49 & 59.88 & 46.13 & 42.82 & 16.89 & 75.59 & 55.79 & 53.40 & 82.40 & 122.68 \\
    SCoPE VLM SFT EGRPO      & 73.07 & 57.31 & 48.27 & 42.29 & 17.90 & 68.98 & 81.89 & 89.33 & 92.86 & 108.57 \\
    \midrule
    & \multicolumn{5}{c|}{\textbf{Action Success Ratio (\%)}} & \multicolumn{5}{c}{\textbf{No Answer Ratio (\%)}} \\
    \cmidrule(lr){2-6} \cmidrule(lr){7-11}
    \textbf{Model} &
    \textbf{MP-DocVQA} & \textbf{SlideVQA} & \textbf{M3DocVQA} & \textbf{DUDE} & \textbf{MMLong} &
    \textbf{MP-DocVQA} & \textbf{SlideVQA} & \textbf{M3DocVQA} & \textbf{DUDE} & \textbf{MMLong} \\
    \midrule
    Qwen2.5 VL 3B   & 89.59 & 65.73 & 74.39 & 85.05 & 67.09 & 2.27  & 1.76  & 10.72 & 7.40  & 1.28  \\
    Qwen2.5 VL 72B  & 99.33 & 99.54 & 96.85 & 95.31 & --    & 0.98  & 3.09  & 12.70 & 9.52  & --    \\
    SCoPE VLM SFT   & 94.42 & 78.49 & 87.39 & 90.90 & 62.52 & 1.33  & 0.79  & 3.50  & 3.71  & 31.44 \\
    SCoPE VLM EGRPO & 96.03 & 81.89 & 89.33 & 92.86 & 66.16 & 0.42  & 0.61  & 1.75  & 3.40  & 23.37 \\
    \bottomrule
  \end{tabular}}
  \caption{Performance-efficiency comparison with baseline Qwen2.5 VL models and SCoPE VLMs on document understanding benchmarks.}
  \label{tab:appendix_efficiency_comparison}
\end{table*}

Table~\ref{tab:appendix_efficiency_comparison} presents the performance-efficiency comparison between baseline Qwen2.5 VL models and SCoPE VLMs. The baseline models demonstrate limited performance in the Chain of Scroll framework, with even Qwen2.5 VL 72B struggling despite its larger capacity. While SCoPE VLM SFT successfully learns the navigation task and achieves competitive ANLS scores, it fails to optimize efficiency metrics and action validity. In contrast, SCoPE VLM EGRPO successfully optimizes across all dimensions, achieving improved action success ratios and substantially reduced no-answer rates while maintaining comparable task performance to SCoPE VLM SFT.

\section{GUI control experiment in Table~\ref{tab:AitZ}: Input prompt and overall results.}
\label{GUI-result-appendix}

\subsection{GUI experiments using the AitZ dataset: Full evaluation results for the General and Web-shopping test set splits.}

Table~\ref{tab:detail_aitz} presents extended experimental results from Table~\ref{tab:AitZ} for the General and Web Shopping domains of the AitZ benchmark. Performance across five actions is evaluated using two metrics: Type Accuracy (Type Acc.) and Exact Match. Type Accuracy measures whether the predicted action type matches the ground truth, while Exact Match requires both the action type and its parameters to match completely. For instance, if the ground truth is ``Scroll Up'' but the model predicts ``Scroll Down,'' Type Accuracy is satisfied (both are scroll actions), but Exact Match is not. Goal Progress, an episode-level metric, represents the average progress toward the goal across all episodes.
\begin{table*}[!h]
\centering
\label{tab:appendix_AitZ_detailed}
\resizebox{1.0\textwidth}{!}{%
\begin{tabular}{@{}l|l|ccccc|c|ccccc|c@{}}
\toprule
\multicolumn{1}{c|}{\multirow{2}{*}{\textbf{Model}}} & \multicolumn{1}{c|}{\multirow{2}{*}{\textbf{Accuracy Type}}} & \multicolumn{6}{c|}{\textbf{General}} & \multicolumn{6}{c}{\textbf{Web Shopping}} \\
\cmidrule(lr){3-8} \cmidrule(lr){9-14}
\multicolumn{2}{c|}{} & \multicolumn{5}{c|}{\textbf{Action}} & \textbf{Episode} & \multicolumn{5}{c|}{\textbf{Action}} & \textbf{Episode} \\
\midrule
 & & \textbf{Click} & \textbf{Scroll} & \textbf{Type} & \textbf{Press} & \textbf{Stop} & \textbf{Goal Progress} & \textbf{Click} & \textbf{Scroll} & \textbf{Type} & \textbf{Press} & \textbf{Stop} & \textbf{Goal Progress} \\
\midrule
\multirow{2}{*}{Qwen 2.5‑VL‑3B} & Type Acc. & 73.77 & 9.49 & 83.95 & 44.07 & 66.67 & \multirow{2}{*}{35.46} & 81.67 & 43.05 & 80.30 & 42.55 & 83.57 & \multirow{2}{*}{41.17} \\
 & Exact Match & 30.21 & 6.57 & 61.73 & 0.00 & 66.67 & & 37.76 & 39.01 & 64.65 & 0.00 & 83.57 & \\
\midrule
\multirow{2}{*}{SCoPE VLM 3B SFT (Ours)} & Type Acc. & 70.11 & 10.22 & 79.01 & 32.20 & 73.08 & \multirow{2}{*}{36.10} & 82.03 & 39.91 & 73.74 & 35.46 & 87.14 & \multirow{2}{*}{41.56} \\
 & Exact Match & 31.00 & 8.76 & 56.17 & 0.00 & 73.08 & & 37.95 & 36.77 & 57.58 & 0.00 & 87.14 & \\
\midrule
\multirow{2}{*}{SCoPE VLM 3B EGRPO (Ours)} & Type Acc. & 72.66 & 8.03 & 82.10 & 35.59 & 76.28 & \multirow{2}{*}{37.51} & 82.31 & 46.64 & 80.81 & 40.43 & 84.29 & \multirow{2}{*}{42.32} \\
 & Exact Match & 33.55 & 8.03 & 54.32 & 0.00 & 76.28 & & 38.31 & 45.29 & 61.62 & 0.00 & 84.29 & \\
\bottomrule
\end{tabular}}
\caption{Detailed performance breakdown on the AitZ benchmark. This table reports Type Accuracy and Exact Match scores for each action type, along with the episode-level Goal Progress, across the General and Web Shopping test splits.}
\label{tab:detail_aitz}
\end{table*}
\newpage

\subsection{Example of input prompt}
Figure~\ref{fig:gui_agent_prompt} shows a prompt converted by merging the system prompts from the AitZ dataset and CoAT.

\begin{figure*}[!h]
  \centering
  \resizebox{1.0\textwidth}{!}{%
    \begin{minipage}{\textwidth}
      \footnotesize
      \begin{tcolorbox}[
        colback=white,
        colframe=black,
        boxsep=4pt,
        arc=0pt,
        outer arc=0pt
      ]
        \textbf{Prompt for GUI Agent Action Prediction}\\[4pt]
        You are a smart and helpful visual assistant that is well trained to manipulate mobile phones.
        Your task is to navigate on the current screen to complete the user request.
        You are provided with:
        \begin{itemize}
          \item Two screenshots of the current mobile phone (one raw, one with UI element annotations).
          \item A brief summarization of the screen content.
          \item A history of actions attempting to accomplish the user request.
          \item The result of the previous action that led to the current screen.
        \end{itemize}
        You are required to decide on the next single-step valid action to be conducted on the current screen so as to fulfill the user request.

        \hrulefill\\[2pt]
        \textbf{Valid Action Spaces}\\[2pt]
        \begin{description}
            \item[\texttt{CLICK\_ELEMENT(idx)}] Clicks on a visible UI element on the screen.
            \item[\texttt{SCROLL(direction)}] Scrolls the content in the given direction ('up', 'down', 'left', 'right').
            \item[\texttt{INPUT(text)}] Inputs the given text into an editable area.
            \item[\texttt{PRESS\_ENTER()}] Confirms or submits an input.
            \item[\texttt{PRESS\_HOME()}] Returns to the home screen.
            \item[\texttt{PRESS\_BACK()}] Returns to the previously visited screen.
            \item[\texttt{STOP(task\_status)}] Stops the task and sets its status ('success' or 'failure').
        \end{description}

        \hrulefill\\[2pt]
        \textbf{Output Format}\\[2pt]
        Your response must be strictly structured in an XML-like format with the following tags:
        \begin{description}
            \item[\texttt{<think>}] Your reasoning and analysis behind the chosen action.
            \item[\texttt{<next>}] A brief, human-readable description of the next action.
            \item[\texttt{<action>}] The precise API function call for the action.
        \end{description}

        \hrulefill\\[4pt]
        \textbf{User Request:} \textcolor{blue}{[User Request]}\\
        \textbf{History Actions:} \textcolor{blue}{[History Actions]}\\
        \textbf{Previous Action Result:} \textcolor{blue}{[Previous Action Result]}\\
        \textbf{Screen Content:} \textcolor{blue}{[Screen Content Summary]}\\
        \textbf{UI Elements:} \textcolor{blue}{[List of UI Elements on the current screen]}\\[4pt]

        You should return your thoughts, a brief description of the output action, and the final action.
        Return the thinking process in \texttt{<think>...</think>}, a brief description in \texttt{<next>...</next>}, and the action in \texttt{<action>...</action>} tags.

      \end{tcolorbox}
    \end{minipage}%
  }
  \caption{The modified AitZ prompt used for training and testing comprises four components: (1) general system instructions, (2) definitions of valid actions, (3) output format requirements, and (4) context-specific information for the current step, including user request, screen summary, and UI elements.}
  \label{fig:gui_agent_prompt}
\end{figure*}

\newpage
\end{document}


\maketitle

\bibliography{custom}

\appendix

\section{Example Appendix}

\subsection{Data recipe experiment results}\label{1-1}

\begin{table}[ht]  
\begin{center}
\resizebox{0.51\textwidth}{!}{%
\begin{tabular}{ll}  
\toprule  
Type & Dataset\\  
\midrule  
General & ShareGPT-4V, SVIT, laion-gpt4v\\  
\midrule  
Document & DocVQA, SynDog-EN, ChartQA, DVQA, AI2D, ref3rec, ref3reg, LLAVAR\\  
\bottomrule  
\end{tabular}  }
\caption{Dataset list}  
\end{center}  
\end{table}  

For our experiments, we utilized a combination of publicly available and proprietary datasets specifically curated to test the efficacy of vision-language models. These included image-text paired datasets, such as COCO and Visual Genome, which are standard in the field for evaluating the performance of VLMs. Our dataset is based on the visual instruction tuning dataset proposed in LLaVA~\cite{liu2023llava}, which has 640k of dataset (removed TextCaps). Although, the dataset has shown outstanding performance in visual instruction tuning, it lacks document oriented dataset which are a critical domain in evaluating VLMs. Thus, in this paper, we explore the relationship between the general performance and domain specific performance in the field of document understanding by supplementing additional document oriented, captioning, and VQA dataset. Although there have been great works to explore the relationship between general and domain specific performance, depending on the types and amount of instruction tuning dataset, the relationship needs to be further explored to optimize the performance of the model for intended use case. Thus, in this paper, we propose a novel curated dataset, achieving optimal general and document specific performance without drastically increasing the dataset size with a single epoch of fine-tuning stage. Furthermore, the proposed dataset only contains the publicly opened dataset without common crawling the web data. For the comparison, our model has been trained both with the 665k dataset and our proposed 760k dataset in this paper. However, for the comparison with the 665k dataset, the dataset ablation is conducted with the model proposed in LLaVA 7B.    

It is widely accepted in the community that Transformers need extensive scale of training dataset in order to reach its full performance. Thus, there are numerous research which leverage the advantages of a large scale dataset. Although it may be a simple straightforward method to increase the model performance, LLaVA demonstrates that meticulously curated dataset can perform better than others. Thus, our team initially attempts to leverage both scale and data curation by increasing dataset scale by the same ratio of dataset types. The dataset classification has been adopted from the proposed classification in InternVL~\cite{chen2023internvl} by captioning, VQA, OCR, grounding, and conversation. The size of the dataset is increased from 760k to 1.5M by about 200k increments, while maintaining the predefined ratio for each categories. For ablation, our team additionally experimented three different 760k dataset based on the 665k from LLAVA with supplementary dataset solely on a single category: VQA, OCR, and grounding which are critical to both general and domain specific performance on document understanding. 

\begin{table}[h]  
\centering  
\resizebox{0.48\textwidth}{!}{%
\begin{tabular}{|l|lllll|}  
\hline  
 & \multicolumn{1}{l|}{\textbf{LLAVA}} & \multicolumn{1}{l|}{\textbf{ocr +100K}} & \multicolumn{1}{l|}{\textbf{grounding +100K}} & \multicolumn{1}{l|}{\textbf{document vqa +100K}} \\ \hline  
MME-P & \multicolumn{1}{l|}{\underline{1510.75}}                & \multicolumn{1}{l|}{1467.97}                 & \multicolumn{1}{l|}{1502}                      & \multicolumn{1}{l|}{1496.26}             \\  
SEED Image      & \multicolumn{1}{l|}{65.41}                 & \multicolumn{1}{l|}{65.82}                  & \multicolumn{1}{l|}{55.66}                     & \multicolumn{1}{l|}{\underline{66.24}}              \\  
TEXTVQA         & \multicolumn{1}{l|}{46.07}                 & \multicolumn{1}{l|}{47.32}                  & \multicolumn{1}{l|}{\underline{47.40}}                     & \multicolumn{1}{l|}{46.68}              \\  
AI2D            & \multicolumn{1}{l|}{54.79}                 & \multicolumn{1}{l|}{55.44}                  & \multicolumn{1}{l|}{55.66}                     & \multicolumn{1}{l|}{\underline{69.20}}            \\  
CHARTQA        & \multicolumn{1}{l|}{18.24}                 & \multicolumn{1}{l|}{18.00}                  & \multicolumn{1}{l|}{16.04}                     & \multicolumn{1}{l|}{\underline{28.32}}              \\  
DOCVQA          & \multicolumn{1}{l|}{28.08}                 & \multicolumn{1}{l|}{30.24}                  & \multicolumn{1}{l|}{29.03}                     & \multicolumn{1}{l|}{\underline{41.39}} \\ \hline              
\end{tabular}%
}  
    \caption{Data curation by data type }  
    \label{tab:by-data-type}  
\end{table}  

The preliminary experiment results from the table ~\ref{tab:by-data-type} demonstrates that the simple un-curated data supplement generally undermines the general performance, while it certainly improves performance in the specific dataset domain. In addition, OCR and grounding is not only one of the general capabilities of VLMs but also critical aspects in document understanding. This is clearly demonstrates in the results from ocr +100k and grounding +100k results. Although the there are evident drops in the general domain benchmarks, there are clear improvements in the document understanding benchmarks. In addition, the document vqa +100K shows that the domain specific dataset can improve the general performance as well, as it still requires general image understanding for diagrams, tables, and insert images.  These results from table 2 further supports the necessities to find a sophiscated data recipes in order to fully leverages the advantages of various types of dataset, while optimizing the general performance of VLMs. 

\begin{table}[h]    
\centering    
\resizebox{0.48\textwidth}{!}{%
\begin{tabular}{|l|lllll|}    
\hline    
& \multicolumn{1}{l|}{\textbf{760k}} & \multicolumn{1}{l|}{\textbf{960k}} & \multicolumn{1}{l|}{\textbf{1.16M}} & \multicolumn{1}{l|}{\textbf{1.36M}} \\ \hline    
MME-P & \multicolumn{1}{l|}{1486.89} & \multicolumn{1}{l|}{\underline{1512.55}} & \multicolumn{1}{l|}{1475.02} & \multicolumn{1}{l|}{1467.30} \\    
SEED Image & \multicolumn{1}{l|}{65.97} & \multicolumn{1}{l|}{66.43} & \multicolumn{1}{l|}{67.21} & \multicolumn{1}{l|}{\underline{67.41}} \\    
TEXTVQA & \multicolumn{1}{l|}{47.9} & \multicolumn{1}{l|}{48.02} & \multicolumn{1}{l|}{48.88} & \multicolumn{1}{l|}{\underline{48.90}} \\    
AI2D & \multicolumn{1}{l|}{67.32} & \multicolumn{1}{l|}{68.1} & \multicolumn{1}{l|}{68.65} & \multicolumn{1}{l|}{67.78} \\    
CHARTQA & \multicolumn{1}{l|}{28.12} & \multicolumn{1}{l|}{30.64} & \multicolumn{1}{l|}{31.32} & \multicolumn{1}{l|}{\underline{32.84}} \\    
DOCVQA & \multicolumn{1}{l|}{41.9} & \multicolumn{1}{l|}{42.19} & \multicolumn{1}{l|}{43.87} & \multicolumn{1}{l|}{43.59} \\ \hline    
\end{tabular}%
}    
\caption{Data curation by scale }    
\label{tab:byscale}    
\end{table}   

The benchmark results of the different scale of dataset from table ~\ref{tab:byscale} indicates that simply increasing dataset scale may effective to increase domain specific performance but it is highly likely to undermines the overall performance or the performance gain converges quickly. From the results, the general domain performance converges around 1M but the domain specific performance still demonstrates a notable performance gain with the higher scale. Thus, it can be concluded that under the current 2 stage training scheme of VLMs, the general performance are susceptible to the dataset recipe and qualities, whereas the domain specific performance can be effectively improved by increasing the scale of training data. 

 \begin{table}[h]
\centering  
\resizebox{0.48\textwidth}{!}{%
\begin{tabular}{|l|l|l|l|l|l|l|l|l|l|}
\hline
760k       & {\ul v1} & \underline{v2}      & v3      & v4      & v5      & v6      & v7      & v8      & v9      \\ \hline
MME-P      & 1486.89  & 1547.67 & 1505.39 & 1515.90 & 1516.71 & 1502.60 & 1485.20 & 1503.23 & 1519.41 \\ \hline
SEED Image & 65.97    & 65.93   & 66.68   & 66.44   & 66.22   & 66.85   & 66.69   & 67.15   & 66.27   \\ \hline
TEXTVQA    & 47.9     & 47.56   & 47.36   & 47.7    & 47.51   & 47.14   & 48.07   & 47.34   & 47.24   \\ \hline
AI2D       & 67.32    & 66.9    & 68.29   & 67.87   & 67.39   & 67.38   & 67.64   & 67.74   & 66.93   \\ \hline
CHARTQA    & 28.12    & 27.72   & 27.48   & 27.8    & 28.04   & 26.6    & 26.48   & 25.92   & 25.72   \\ \hline
DOCVQA     & 41.9     & 40.05   & 36.94   & 38.34   & 36.51   & 37.48       & 37.31   & 36.86   & 36.62       \\ \hline
\end{tabular}
}  
\caption{Finetuning dataset recipe for each dataset}  
\label{tab:my-table}  
\end{table}  

Although table \ref{tab:byscale} shows that supplementary dataset can increase the general performance and the domain specific performance, it also highlights the importance of balance between curated high quality data and its scale. As the lower end 760k shows notable improvement from the 665K from LLAVA instruction tuning dataset, our team decided to find the optimal balance between the general and domain specific dataset. We aims to maximizing the general performance especially MME benchmark score while maintaining the dataset size to 760k and the document understanding performance. For experiments, our team used the final tuned version of 760Kv2 in order to minimize the training time by suppressing the dataset scale, while adequately leverages performance gain both from the general and domain specific dataset.

\begin{table}[h]
\centering  
\resizebox{0.48\textwidth}{!}{%
\begin{tabular}{|l|l|l|l|l|l|l|l|l|l|l|l|}
\hline
Dataset   & DocVQA      & SynDog-EN   & ChartQA    & DVQA        & AI2D       & LAION-GPT-V & ShareGPT-4V & ref3rec     & ref3reg     & SVIT       & LLAVAR      \\ \hline
760K & 30K & 10K & 7K & 26K & 4K & 8K & 30K & 0K & 0K & 0K & 5K \\ \hline
760Kv2 & 20K & 10K & 7K & 10K & 4K & 8K & 31K & 10K & 10K & 0K & 20K \\ \hline
960K & 39K & 20K & 7K & 104K & 4K & 8K & 45K & 0K & 0K & 40K & 53K \\ \hline
1.16K & 39K & 30K & 7K & 149K & 4K & 8K & 60K & 0K & 0K & 80K & 143K \\ \hline
1.36K & 39K & 60K & 7K & 239K & 4K & 8K & 100K & 0K & 0K & 120K & 143K        \\ \hline
\end{tabular}
}  
\caption{Dataset recipe in addition to 640K form LLAVA(without textcap dataset)}  
\label{tab:my-table}  
\end{table}